\documentclass[twoside]{article}
\usepackage[accepted]{aistats2018}
\usepackage[utf8]{inputenc} % allow utf-8 input
\usepackage[T1]{fontenc}    % use 8-bit T1 fonts
\usepackage{url}            % simple URL typesetting
\usepackage{amsfonts}       % blackboard math symbols
\usepackage{nicefrac}       % compact symbols for 1/2, etc.
\usepackage{microtype}      % microtypography
\usepackage{amssymb}
\usepackage{amsmath}
\usepackage{amsthm}
\usepackage{array}
\usepackage{algorithm,algorithmic}
\usepackage[tight,footnotesize]{subfigure}
\usepackage{multirow}
\usepackage{mdwmath}
\usepackage{mdwtab}
\usepackage{epsfig}

\newtheorem{definition}{Definition}
\newtheorem{theorem}{Theorem}
\newtheorem{lemma}{Lemma}
\newtheorem{corollary}{Corollary}
\newtheorem{property}{Property}
\newtheorem{assumption}{Assumption}

\begin{document}

\twocolumn[

\aistatstitle{Guaranteed Sufficient Decrease for Stochastic Variance Reduced Gradient Optimization}

\aistatsauthor{ Fanhua Shang,\,\quad Yuanyuan Liu,\,\quad Kaiwen Zhou,\,\quad James Cheng,\,\quad Kelvin K.W. Ng}
\aistatsaddress{ Department of Computer Science and Engineering, The Chinese University of Hong Kong\footnotetext{Corresponding author.}}
\vspace{-3mm}
\aistatsauthor{ Yuichi Yoshida}
\aistatsaddress{ National Institute of Informatics, Tokyo, Japan} ]

\begin{abstract}
In this paper, we propose a novel sufficient decrease technique for stochastic variance reduced gradient descent methods such as SVRG and SAGA. In order to make sufficient decrease for stochastic optimization, we design a new sufficient decrease criterion, which yields sufficient decrease versions of stochastic variance reduction algorithms such as SVRG-SD and SAGA-SD as a byproduct. We introduce a coefficient to scale current iterate and to satisfy the sufficient decrease property, which takes the decisions to shrink, expand or even move in the opposite direction, and then give two specific update rules of the coefficient for Lasso and ridge regression. Moreover, we analyze the convergence properties of our algorithms for strongly convex problems, which show that our algorithms attain linear convergence rates. We also provide the convergence guarantees of our algorithms for non-strongly convex problems. Our experimental results further verify that our algorithms achieve significantly better performance than their counterparts.
\end{abstract}

\section{Introduction}
Stochastic gradient descent (SGD) has been successfully applied to many large-scale machine learning problems~\cite{krizhevsky:deep,zhang:sgd}, due to its low per-iteration cost, $O(d)$. However, standard SGD estimates the gradient from only one or a few samples, and thus the variance of the stochastic gradient estimator may be large~\cite{johnson:svrg,zhao:prox-smd}, which leads to slow convergence and poor performance. In particular, even under the strongly convex (SC) condition, the convergence rate of standard SGD is only sub-linear~\cite{rakhlin:sgd,shamir:sgd2}. Recently, the convergence rate of the standard SGD has been improved by various stochastic variance reduction gradient methods, such as SAG~\cite{roux:sag}, SDCA~\cite{shalev-Shwartz:sdca}, SVRG~\cite{johnson:svrg}, SAGA~\cite{defazio:saga}, Finito~\cite{defazio:Finito}, MISO~\cite{mairal:miso}, and their proximal variants, such as~\cite{schmidt:sag}, \cite{shalev-Shwartz:prox-sdca} and \cite{xiao:prox-svrg}. Under the SC condition, these stochastic variance reduction algorithms achieve linear convergence rates.

Very recently, some acceleration techniques were proposed to further speed up the stochastic variance reduction methods mentioned above. These techniques include importance sampling~\cite{zhao:prox-smd}, exploiting neighborhood structure in the training data to share and re-use information about past stochastic gradients~\cite{hofmann:vrsg}, incorporating Nesterov's acceleration techniques~\cite{lin:vrsg,nitanda:svrg} or momentum acceleration tricks~\cite{zhu:Katyusha}, reducing the number of gradient computations in the early iterations~\cite{zhu:univr,babanezhad:vrsg,zhang:svrg,shang:fsvrg}, and the projection-free property of the conditional gradient method~\cite{hazan:svrf}. In particular, Katyusha~\cite{zhu:Katyusha} can attain the upper complexity bounds for both SC and non-strongly convex (non-SC) composite problems, respectively, as discussed in \cite{woodworth:bound}.

\textbf{Challenges and Main Contributions}. So far the two most popular stochastic gradient estimators are the SVRG estimator independently introduced by~\cite{johnson:svrg,zhang:svrg} and the SAGA estimator~\cite{defazio:saga}. All the estimators may be very different from their full gradient counterparts, thus moving in the direction may not decrease the objective function anymore, as stated in~\cite{zhu:Katyusha}. To address this problem, inspired by the success of sufficient decrease methods for deterministic optimization such as~\cite{li:apg,wolfe:sdg}, we propose a novel sufficient decrease technique for a class of stochastic variance reduction methods, including the widely-used SVRG and SAGA methods. Notably, our method with partial sufficient decrease achieves average time complexity per-iteration as low as the original SVRG and SAGA methods. We summarize our main contributions below.
\vspace{-2mm}
\begin{itemize}
  \item For making sufficient decrease for stochastic optimization, we design a sufficient decrease strategy to further reduce the cost function, in which we also introduce a coefficient to take the decision to shrink, expand or move in the opposite direction.
  \vspace{-5mm}
  \item We incorporate our sufficient decrease technique, together with momentum acceleration, into two representative SVRG and SAGA algorithms, which lead to SVRG-SD and SAGA-SD. Moreover, we give two specific update rules of the coefficient for Lasso and ridge regression problems as notable examples.
  \vspace{-1mm}
  \item  Moreover, we analyze the convergence property of SVRG-SD, which shows that SVRG-SD converges linearly for SC composite minimization problems. Unlike most of the stochastic variance reduction methods such as SVRG, we also provide the convergence guarantee of SVRG-SD for non-SC composite minimization problems.
  \vspace{-1mm}
  \item Finally, we show by experiments that SVRG-SD and SAGA-SD achieve significantly better performance than SVRG~\cite{johnson:svrg} and SAGA~\cite{defazio:saga}. Compared with the well-known  accelerated method, Katyusha~\cite{zhu:Katyusha}, our algorithms also have much better performance in most cases.
\end{itemize}

\section{Preliminary and Related Work}

\subsection{Notations}
Throughout this paper, we use $\|\!\cdot\!\|$ to denote the $\ell_{2}$-norm (also known as the standard Euclidean norm), and $\|\!\cdot\!\|_{1}$ is the $\ell_{1}$-norm, i.e., $\|x\|_{1}\!=\!\sum^{d}_{i=1}\!|x_{i}|$. $\nabla\!f(\cdot)$ denotes the full gradient of the function $f(\cdot)$ if it is differentiable, or $\partial\!f(\cdot)$ the subgradient if $f(\cdot)$ is only Lipschitz continuous.

In this paper, we consider the following unconstrained composite convex optimization problem:
\vspace{-1mm}
\begin{equation}\label{equ1}
\min_{x\in\mathbb{R}^{d}} F(x)\stackrel{\rm{def}}{=}f(x)+r(x)=\frac{1}{n}\!\sum\nolimits_{i=1}^{n}\!f_{i}(x)+r(x),
\vspace{-1mm}
\end{equation}
where $f_{i}(x)\!:\!\mathbb{R}^{d}\!\rightarrow\!\mathbb{R},\,i\!=\!1,\ldots,n$ are the smooth convex component functions, and $r(x)$ is a relatively simple convex (but possibly non-differentiable) function.

We consider the problem \eqref{equ1} under the following standard assumptions.

\begin{assumption}
\label{assum1}
Each convex function $f_{i}(\cdot)$ is $L$-smooth, iff there exists a constant $L\!>\!0$ such that for any $x,y\!\in\! \mathbb{R}^{d}$, $\|\nabla\! f_{i}(x)-\nabla\! f_{i}(y)\|\leq L\|x-y\|$.
\end{assumption}

\begin{assumption}
\label{assum2}
$F(\cdot)$ is $\mu$-strongly convex, iff there exists a constant $\mu\!>\!0$ such that for any $x,y\!\in\!\mathbb{R}^{d}$,
\vspace{-1mm}
\begin{equation}\label{equ15}
F(y)\geq F(x)\!+\!\vartheta^{T}\!(y-x)\!+\!\frac{\mu}{2}\|y-x\|^{2},\;\,\forall\vartheta\!\in\!\partial F(x),
\vspace{-1mm}
\end{equation}
where $\partial F(x)$ is the subdifferential of $F(\cdot)$ at $x$. If $F(\cdot)$ is smooth, we can revise the inequality~\eqref{equ15} by simply replacing the sub-gradient $\vartheta\in\partial F(x)$ with $\nabla\! F(x)$.
\end{assumption}

\begin{algorithm}[t]
\caption{SVRG}
\label{alg3}
\renewcommand{\algorithmicrequire}{\textbf{Input:}}
\renewcommand{\algorithmicensure}{\textbf{Initialize:}}
\renewcommand{\algorithmicoutput}{\textbf{Output:}}
\begin{algorithmic}[1]
\REQUIRE The number of epochs $S$, the number of iterations $m$ per epoch, and the step size $\eta$.\\
\ENSURE $\widetilde{x}^{0}$.\\
\FOR{$s=1,2,\ldots,S$}
\STATE {$\widetilde{\mu}^{s-\!1}=\frac{1}{n}\!\sum^{n}_{i=1}\!\nabla\!f_{i}(\widetilde{x}^{s-1})$;}
\STATE {$x^{s}_{0}=\widetilde{x}^{s-1}$;}
\FOR{$k=1,2,\ldots,m$}
\STATE {Pick $i^{s}_{k}$ uniformly at random from $[n]$;}
\STATE {$\widetilde{\nabla}\! f_{i^{s}_{k}}(x^{s}_{k-1})=\nabla\! f_{i^{s}_{k}}(x^{s}_{k-1})-\nabla\! f_{i^{s}_{k}}(\widetilde{x}^{s-\!1})+\widetilde{\mu}^{s-\!1}$;}
\STATE {$x^{s}_{k}=x^{s}_{k-1}-\eta\widetilde{\nabla}\!f_{i^{s}_{k}}(x_{k-1})$;}
\ENDFOR
\STATE {$\widetilde{x}^{s}=x^{s}_{m}$;}
\ENDFOR
\OUTPUT {$\widetilde{x}^{S}$}
\end{algorithmic}
\end{algorithm}

\subsection{SVRG and SAGA}
Recently, many stochastic variance reduced methods~\cite{johnson:svrg,roux:sag,xiao:prox-svrg,zhang:svrg} have been proposed for some special cases of Problem~\eqref{equ1}. Under smoothness and SC assumptions, and $r(x)\equiv0$, SAG~\cite{roux:sag} achieves a linear convergence rate. A recent line of work, such as~\cite{johnson:svrg,xiao:prox-svrg}, has been proposed with similar convergence rates to SAG but without the memory requirements for all gradients. SVRG~\cite{johnson:svrg} begins with an initial estimate $\widetilde{x}^{0}$, sets $x^{1}_{0}=\widetilde{x}^{0}$ and then generates a sequence of $x^{s}_{k}$ at the $s$-th epoch ($k\!=\!1,2,\ldots,m$, where $m$ is usually set to $2n$, as suggested in~\cite{johnson:svrg,xiao:prox-svrg}) using
\begin{equation}\label{equ2}
x^{s}_{k}=x^{s}_{k-\!1}\!-\eta\!\left[\nabla\! f_{i^{s}_{k}}\!(x^{s}_{k-\!1})-\nabla\! f_{i^{s}_{k}}\!(\widetilde{x}^{s-1})+\widetilde{\mu}^{s-1}\right]\!,
\end{equation}
where $\eta\!>\!0$ is the step size, $\widetilde{\mu}^{s-\!1}\!=\!\frac{1}{n}\!\sum^{n}_{i=1}\!\nabla\! f_{i}(\widetilde{x}^{s-1})$ is the full gradient of $f(\cdot)$ at the snapshot $\widetilde{x}^{s-1}$, and $i^{s}_{k}$ is chosen uniformly at random from $\{1,2,\ldots,n\}$. After every $m$ stochastic iterations, we set $\widetilde{x}^{s}\!=\!x^{s}_{m}$, and reset $k\!=\!1$ and $x^{s+1}_{0}\!=\widetilde{x}^{s}$ (see Algorithm~\ref{alg3} for details).

In particular, the stochastic gradient estimator in Line 6 of Algorithm~\ref{alg3}, i.e., the SVRG estimator independently introduced in~\cite{johnson:svrg,zhang:svrg}, is a popular choice for stochastic gradient estimators. Unfortunately, most of the stochastic variance reduction methods~\cite{shalev-Shwartz:sdca,defazio:Finito,xiao:prox-svrg}, including SVRG, only have convergence guarantees for smooth and SC problems. However, $F(\cdot)$ may be non-SC in many machine learning applications (e.g., Lasso) or non-smooth and SC problems, e.g., elastic-net. For solving the non-smooth problem~\eqref{equ1}, a proximal variant of SVRG, i.e., Prox-SVRG~\cite{xiao:prox-svrg}, has been proposed and relies on the use of the proximal operator, $\textrm{prox}^{r}_{\eta}(\cdot)$, which is defined as:
\begin{equation*}
\textrm{prox}^{r}_{\eta}(y)=\arg\min_{x}\Big\{({1}/{2\eta})\!\cdot\!\|x\!-\!y\|^{2}+r(x)\Big\}.
\end{equation*}

\cite{defazio:saga} proposed SAGA, a fast incremental gradient method in the spirit of SAG and SVRG, which works for both SC and non-SC objective functions, as well as in the proximal setting. Unlike SVRG and Prox-SVRG, both of which are two-stage algorithms, SAGA is a single-stage incremental gradient algorithm. Its main update rules are formulated as follows:
\vspace{-1mm}
\begin{align}
&\widetilde{\nabla}\! f_{i_{k}}\!(x_{k-\!1})=g^{k}_{i_{k}}\!-\!g^{k-\!1}_{i_{k}}\!+\!\frac{1}{n}\!\sum^{n}_{j=1}g^{k-\!1}_{j}\tag{\theequation a},\label{equ31}\\
&x_{k}=\textrm{prox}^{r}_{\eta}\!\left(x_{k-\!1}\!-\eta \widetilde{\nabla}\! f_{i_{k}}\!(x_{k-\!1})\right)\tag{\theequation b},\label{equ32}
\vspace{-1mm}
\end{align}
where $g^{k}_{j}$ is updated for all $j\!=\!1,\ldots,n$ as follows: $g^{k}_{j}\!=\!\nabla\!f_{i_{k}}\!(x^{k-\!1})$ if $i_{k}\!=\!j$, and $g^{k}_{j}\!=\!g^{k-\!1}_{j}$ otherwise (see \cite{defazio:saga,defazio:sagab} for details). Here, the stochastic gradient estimator in~\eqref{equ31} is called as the SAGA estimator.

\subsection{Sufficient Decrease for Deterministic Optimization}
The technique of sufficient decrease (e.g., the well-known line search technique~\cite{more:ls}) has been studied for deterministic optimization~\cite{li:apg,wolfe:sdg}. For example, \cite{li:apg} proposed the following sufficient decrease condition:
\begin{equation}\label{equ4}
F(x_{k})\leq F(x_{k-1})-\delta\|y_{k}-x_{k-1}\|^{2},
\end{equation}
where $\delta\!>\!0$ is a small constant, and $y_{k}\!=\!\textrm{prox}^{r}_{\eta_{k}}\!(x_{k-\!1}\!-\!\eta_{k}\nabla\! f(x_{k-\!1}))$.
Inspired by the strategy for deterministic optimization, in this paper we design a novel sufficient decrease technique for stochastic optimization, which is used to further reduce the cost function and speed up its convergence.

\section{SVRG with Sufficient Decrease}
In this section, we propose a novel sufficient decrease technique for stochastic variance reduction optimization methods, which include the widely-used SVRG method.\ To make sufficient decrease for stochastic optimization, we design an effective sufficient decrease strategy that can further reduce the cost function. Then a coefficient $\theta$ is introduced to satisfy the sufficient decrease condition, and takes the decisions to shrink, expand or move in the opposite direction. Moreover, we present a variant of SVRG with sufficient decrease and momentum acceleration (called SVRG-SD). We also give two specific schemes to compute $\theta$ for Lasso and ridge regression.

\begin{algorithm}[t]
\caption{SVRG-SD}
\label{alg1}
\renewcommand{\algorithmicrequire}{\textbf{Input:}}
\renewcommand{\algorithmicensure}{\textbf{Initialize:}}
\renewcommand{\algorithmicoutput}{\textbf{Output:}}
\begin{algorithmic}[1]
\REQUIRE the number of epochs $S$, the number of iterations $m$ per epoch, and the step size $\eta$.\\
\ENSURE $\widetilde{x}^{0}$ for Case of SC, or $\widetilde{x}^{0}\!=\widetilde{y}^{0}$ for Case of non-SC.
\FOR{$s=1,2,\ldots, S$}
\STATE {Case of SC: $\,x^{s}_{0}=\widehat{x}^{s}_{0}=\widetilde{x}^{s-1}$;}
\STATE {Case of non-SC: $\,x^{s}_{0}=\widehat{x}^{s}_{0}=\widetilde{y}^{s-1}$;}
\STATE {$\widetilde{\mu}^{s-\!1}=\frac{1}{n}\!\sum^{n}_{i=1}\!\nabla\!f_{i}(\widetilde{x}^{s-\!1})$;}
\FOR{$k=1,2,\ldots,m$}
\STATE {Pick $i^{s}_{k}$ uniformly at random from $[n]$;}
\STATE {$\widetilde{\nabla}\! f_{i^{s}_{k}}(x^{s}_{k-\!1})=\nabla\! f_{i^{s}_{k}}(x^{s}_{k-\!1})-\!\nabla\! f_{i^{s}_{k}}(\widetilde{x}^{s-\!1})+\widetilde{\mu}^{s-\!1}$;}
\STATE {Update $y^{s}_{k}$, $\theta_{k}$, and $x^{s}_{k}$ by \eqref{equ8}, \eqref{equ5}, and \eqref{equ7};}
\ENDFOR
\STATE {$\widetilde{x}^{s}=\frac{1}{m}\!\sum^{m}_{k=1}\widehat{x}^{s}_{k}$;}
\STATE {Case of non-SC: $\widetilde{y}^{s}=[x^{s}_{m}-(1\!-\!\sigma)\widehat{x}^{s}_{m}]/\sigma$;}
\ENDFOR
\OUTPUT $\overline{x}\!=\!\widetilde{x}^{S}$ (SC), \,or \,$\overline{x}\!=\!\widetilde{x}^{S}$ if $F(\frac{1}{S}\!\sum^{S}_{s=1}\!\widetilde{x}^{s})\!\geq\! F(\widetilde{x}^{S})$ and $\overline{x}\!=\!\frac{1}{S}\!\sum^{S}_{s=1}\!\widetilde{x}^{s}$ otherwise (non-SC)
\end{algorithmic}
\end{algorithm}

\subsection{Our Sufficient Decrease Technique}
\label{sec31}
Suppose $x^{s}_{k}\!=\!\textrm{prox}^{r}_{\eta}(x^{s}_{k-\!1}\!-\!\eta[\nabla\! f_{i^{s}_{k}}\!(x^{s}_{k-\!1})\!-\!\nabla\! f_{i^{s}_{k}}\!(\widetilde{x}^{s-\!1})\!+\!\widetilde{\mu}^{s-\!1}])$ for the $s$-th outer-iteration and the $k$-th inner-iteration. Unlike the full gradient method, the stochastic gradient estimator is somewhat inaccurate (i.e., it may be very different from $\nabla\! f(x^{s}_{k-\!1})$), then further moving in the updating direction may not decrease the objective value anymore~\cite{zhu:Katyusha}. That is, $F(x^{s}_{k})$ may be larger than $F(x^{s}_{k-\!1})$ even for very small step length $\eta\!>\!0$. Motivated by this observation, we design a factor $\theta$ to scale the current iterate $x^{s}_{k-\!1}$ for the decrease of the objective function. For SVRG-SD, the cost function with respect to $\theta$ is formulated as follows:
\vspace{-1mm}
\begin{equation}\label{equ5}
\min_{\theta\in\mathbb{R}} F(\theta x^{s}_{k-\!1})\!+\!\frac{\zeta(1\!-\!\theta)^2}{2}\!\|\nabla\!f_{i^{s}_{k}}\!(x^{s}_{k-\!1})\!-\!\nabla\! f_{i^{s}_{k}}\!(\widetilde{x}^{s-\!1})\|^{2}\!,
\vspace{-1mm}
\end{equation}
where $\zeta\!=\!\frac{\delta\eta}{1-L\eta}$ is a trade-off parameter between the two terms, $\delta$ is a small constant and set to $0.1$. The second term in~\eqref{equ5} involves the norm of the residual of stochastic gradients, and plays the same role as the second term of the right-hand side of~\eqref{equ4}. Different from existing sufficient decrease techniques including \eqref{equ4}, a varying factor $\theta$ instead of a constant is introduced to scale $x^{s}_{k-\!1}$ and the coefficient of the second term of~\eqref{equ5}, and $\theta$ plays a similar role as the step-size parameter optimized via a line-search for deterministic optimization. However, line search techniques have a high computational cost in general, which limits their applicability to stochastic optimization~\cite{mahsereci:sgd}.

Note that $\theta$ is a scalar and takes the decisions to shrink, expand $x^{s}_{k-\!1}$ or move in the opposite direction of $x^{s}_{k-\!1}$. The detailed schemes to calculate $\theta$ for Lasso and ridge regression are given in Section~\ref{subsec33}. We first present the following sufficient decrease condition in the statistical sense for stochastic optimization.

\begin{property}
\label{prop11}
For given $x^{s}_{k-\!1}$ and the solution $\theta_{k}$ of Problem \eqref{equ5}, then the following inequality holds
\vspace{-1mm}
\begin{equation}\label{equ6}
F(\theta_{k}x^{s}_{k-\!1})\leq F(x^{s}_{k-\!1})-\frac{\zeta(1\!-\!\theta_{k})^2}{2}\|\widetilde{p}_{i^{s}_{k}}\|^{2},
\end{equation}
where $\widetilde{p}_{i^{s}_{k}}\!=\!\nabla\!f_{i^{s}_{k}}\!(x^{s}_{k-\!1})\!-\!\nabla\! f_{i^{s}_{k}}\!(\widetilde{x}^{s-\!1})$ for SVRG-SD.
\end{property}

It is not hard to verify that $F(\cdot)$ can be further decreased via our sufficient decrease technique, when the current iterate $x^{s}_{k-\!1}$ is scaled by the coefficient $\theta_{k}$. Indeed, for the special case when $\theta_{k}\!=\!1$ for some $k$, the inequality in (\ref{equ6}) can be still satisfied. Unlike the sufficient decrease condition for deterministic optimization~\cite{li:apg,wolfe:sdg}, $\theta_{k}$ may be a negative number, which means to move in the opposite direction of $x^{s}_{k-\!1}$.

\subsection{Momentum Acceleration}
\label{subsec32}
In this part, we first design the update rule for the key variable $x^{s}_{k}$ with the coefficient $\theta_{k}$ as follows:
\begin{equation}\label{equ7}
x^{s}_{k}=y^{s}_{k}+(1-\sigma)(\widehat{x}^{s}_{k}-\widehat{x}^{s}_{k-1}),
\end{equation}
where $\widehat{x}^{s}_{k}\!=\!\theta_{k}x^{s}_{k-\!1}$, $\sigma\!\in\![0,1]$ is a constant and can be set to $\sigma\!=\!1/2$ which also works well in practice. In fact, the second term of the right-hand side of~\eqref{equ7} plays a momentum acceleration role as in stochastic and deterministic optimization~\cite{nitanda:svrg,zhu:Katyusha,liu:sadmm,nesterov:fast,nesterov:co,beck:fista,su:nag}. That is, by introducing this term, we can utilize the previous information of gradients to update $x^{s}_{k}$. Then the update rule of $y^{s}_{k}$ is given by
\vspace{-1mm}
\begin{equation}\label{equ8}
y^{s}_{k}=\textrm{prox}^{r}_{\eta}\!\left(x^{s}_{k-1}-\eta\widetilde{\nabla} f_{i^{s}_{k}}(x^{s}_{k-1})\right),
\end{equation}
where $\eta\!=\!1/(L\alpha)$, $L\!>\!0$ is a Lipschitz constant (see Assumption~\ref{assum1}), and $\alpha\!\geq\! 1$ denotes a constant. From the update rule in~\eqref{equ8}, it is clear that SVRG-SD can tackle the non-smooth minimization problem~\eqref{equ1} directly as Prox-SVRG~\cite{xiao:prox-svrg} and SAGA~\cite{defazio:saga}. In addition, $\widetilde{\nabla}\! f_{i^{s}_{k}}\!(x^{s}_{k-\!1})$ can be the two most popular choices for stochastic gradient estimators: the SVRG estimator~\cite{johnson:svrg,zhang:svrg} and the SAGA estimator~\cite{defazio:saga}. For SVRG-SD, $\widetilde{\nabla}\! f_{i^{s}_{k}}\!(x^{s}_{k-\!1})$ is defined as follows:
\begin{eqnarray}\label{equ91}
\widetilde{\nabla}\! f_{i^{s}_{k}}\!(x^{s}_{k-1})=\nabla\! f_{i^{s}_{k}}\!(x^{s}_{k-1})\!-\!\nabla\! f_{i^{s}_{k}}\!(\widetilde{x}^{s-1})\!+\!\widetilde{\mu}^{s-1}.
\end{eqnarray}

In summary, we propose a novel variant of SVRG with sufficient decrease (called SVRG-SD) to solve both SC and non-SC problems, as outlined in \textbf{Algorithm} \ref{alg1}. For the case of SC, $x^{s}_{0}\!=\!\widehat{x}^{s}_{0}\!=\!\widetilde{x}^{s-\!1}$, while $x^{s}_{0}\!=\!\widehat{x}^{s}_{0}\!=\!\widetilde{y}^{s-\!1}$ and $\widetilde{y}^{s}\!=\![x^{s}_{m}\!-\!(1-\sigma)\widehat{x}^{s}_{m}]/\sigma$ are set for the case of non-SC. Note that when $\theta_{k}\!\equiv\!1$ and $\sigma\!=\!1$, the proposed SVRG-SD degenerates to the original SVRG or its proximal variant, Prox-SVRG~\cite{xiao:prox-svrg}. In this sense, SVRG and Prox-SVRG can be seen as the special cases of the proposed SVRG-SD algorithm.

\subsection{Computing Coefficients for Lasso and Ridge Regression}
\label{subsec33}
In this part, we give the closed-form solutions of the coefficient $\theta$ for Lasso and ridge regression.

For Lasso problems and given $x^{s}_{k-\!1}$, we have
\vspace{-1mm}
\begin{equation*}
F(\theta x^{s}_{k-\!1})=\frac{1}{2n}\!\sum^{n}_{i=1}\!(\theta a^{T}_{i}\!x^{s}_{k-\!1}\!-\!b_{i})^{2}\!+\!\lambda\|\theta x^{s}_{k-\!1}\|_{1}.
\vspace{-1mm}
\end{equation*}
The closed-form solution of Problem~\eqref{equ5} for SVRG-SD can be obtained as follows:
\vspace{-1mm}
\begin{equation}\label{equ10}
\theta_{k}=\mathcal{S}_{\tau}\!\left(\frac{\frac{1}{n}b^{T}\!Ax^{s}_{k-\!1}+\zeta\|\widetilde{p}_{i^{s}_{k}}\|^{2}}{\|Ax^{s}_{k-\!1}\|^{2}/n+\zeta\|\widetilde{p}_{i^{s}_{k}}\|^{2}}\right),
\vspace{-1mm}
\end{equation}
where $A\!=\![a_{1},\ldots,a_{n}]^{T}\!$ is the data matrix containing $n$ data samples, $b\!=\![b_{1},\ldots,b_{n}]^{T}\!$, and $\mathcal{S}_{\tau}$ is the so-called soft thresholding operator~\cite{donoho:st} with the following threshold,
\vspace{-1mm}
\begin{equation*}
\tau=\frac{\lambda\|x^{s}_{k-\!1}\|_{1}}{\|Ax^{s}_{k-\!1}\|^{2}/n+\zeta\|\widetilde{p}_{i^{s}_{k}}\|^{2}}.
\vspace{-1mm}
\end{equation*}

For ridge regression problems, we have
\vspace{-1mm}
\begin{equation*}
F(\theta x^{s}_{k-\!1})=\frac{1}{2n}\!\sum^{n}_{i=1}\!(\theta a^{T}_{i}\!x^{s}_{k-\!1}\!-\!b_{i})^{2}\!+\!\frac{\lambda}{2}\|\theta x^{s}_{k-\!1}\|^{2}.
\vspace{-1mm}
\end{equation*}
The closed-form solution of Problem~\eqref{equ5} for SVRG-SD is given by
\vspace{-1mm}
\begin{equation}\label{equ11}
\theta_{k}=\frac{\frac{1}{n}b^{T}\!Ax^{s}_{k-\!1}+\zeta\|\widetilde{p}_{i^{s}_{k}}\|^{2}}{\|Ax^{s}_{k-\!1}\|^{2}/n+\zeta\|\widetilde{p}_{i^{s}_{k}}\|^{2}+\lambda\|x^{s}_{k-\!1}\|^{2}}.
\vspace{-1mm}
\end{equation}
We can compute the coefficient for other loss functions (e.g., logistic regression) using line search (e.g., Armijo's line search~\cite{Armijo:line}) or their approximations~\cite{bach:sgd}.

\section{Convergence Analysis}
In this section, we provide the convergence analysis of SVRG-SD for solving both SC and non-SC minimization problems.

\subsection{Convergence Property for SC Problems}
In this part, we analyze the convergence property of SVRG-SD for solving the SC problem \eqref{equ1}. The first main result is the following theorem, which provides the convergence rate of SVRG-SD.

\begin{theorem}
\label{theo1}
Suppose Assumption~\ref{assum1} holds. Let $x^{*}$ be the optimal solution of Problem \eqref{equ1}, and $\{(x^{s}_{k},y^{s}_{k},\theta_{k})\}$ be the sequence produced by SVRG-SD, $\eta\!=\!1/(L\alpha)$, and $\frac{2}{\alpha-1}\!<\!\sigma$, then
\begin{equation*}
\begin{split}
&\mathbb{E}\!\left[F(\widetilde{x}^{s})\!-\!F(x^{*})\right]\!\leq\left(\!\frac{\frac{1-\sigma}{m}\!+\!\frac{2}{\alpha-1}}{\sigma\!-\!\frac{2}{\alpha-1}\!+\!\widehat{\beta}}\right)\!\mathbb{E}\!\left[F(\widetilde{x}^{s-\!1})\!-\!F(x^{*})\right]\\
&\quad+\frac{L\alpha\sigma^{2}}{2m\!\left(\sigma\!-\!\frac{2}{\alpha-1}\!+\!\widehat{\beta}\right)}\mathbb{E}\!\left[\|x^{*}\!-\!z^{s}_{0}\|^{2}\!-\!\|x^{*}\!-\!z^{s}_{m}\|^{2}\right]\!,
\end{split}
\end{equation*}
where $z^{s}_{0}\!=\!\left[x^{s}_{0}\!-\!(1\!-\!\sigma)\widehat{x}^{s}_{0}\right]/\sigma$, $z^{s}_{m}\!=\!\left[x^{s}_{m}\!-\!(1\!-\!\sigma)\widehat{x}^{s}_{m-\!1}\right]/\sigma$, $\widehat{\beta}\!=\!\min_{s=1,\ldots,S}\widehat{\beta}^{s}\!\geq\!0$, and $\widehat{\beta}^{s}\!=\!\mathbb{E}[\sum^{m}_{k=1}\!\!\frac{2c_{k}\beta_{k}}{\alpha-1}(F(\widehat{x}^{s}_{k})\!-\!F(x^{*}))]/\mathbb{E}[\sum^{m}_{k=1}\!(F(\widehat{x}^{s}_{k})\!-\!F(x^{*}))]$.
\end{theorem}

The proof of Theorem~\ref{theo1} and the definitions of $c_{k}$ and $\beta_{k}$ are given in the Supplementary Material. The linear convergence of SVRG-SD follows immediately.

\begin{corollary}[SC]
\label{coro1}
Suppose each $f_{i}(\cdot)$ is $L$-smooth, and $F(\cdot)$ is $\mu$-strongly convex. Setting $\alpha\!=\!19$, $\sigma\!=\!1/2$, and $m$ sufficiently large so that
\begin{equation*}
\rho=\frac{9}{(7\!+\!18\widehat{\beta})m}+\frac{2}{7\!+\!18\widehat{\beta}}+\frac{171L}{(14\!+\!36\widehat{\beta})m\mu}<1,
\end{equation*}
then SVRG-SD has the geometric convergence in expectation:
\begin{equation*}
\mathbb{E}\!\left[F(\overline{x})-F(x^{*})\right]\leq\rho^{S}\!\left[F(\widetilde{x}^{0})-F(x^{*})\right]\!.
\end{equation*}
\end{corollary}

The proof of Corollary~\ref{coro1} is given in the Supplementary Material. From Corollary~\ref{coro1}, we can see that SVRG-SD has a linear convergence rate for SC problems. As discussed in~\cite{xiao:prox-svrg}, $\rho\!\approx\!\frac{L/\mu}{\nu(1-4\nu)m}\!+\!\frac{4\nu}{1-4\nu}$ for the proximal variant of SVRG~\cite{xiao:prox-svrg}, where $\nu\!=\!1/\alpha$. For a reasonable comparison, we use the same parameter settings for SVRG and SVRG-SD, e.g., $\alpha\!=\!19$ and $m\!=\!57L/\mu$. Then we can see that $\rho_{\textrm{SVRG}}\!\approx\!31/45$ for SVRG and $\rho_{\textrm{SVRG-SD}}\!\approx\!{7}/{(14\!+\!36\widehat{\beta})}\!<\!{1}/2$ for SVRG-SD, that is, $\rho_{\textrm{SVRG-SD}}$ is smaller than $\rho_{\textrm{SVRG}}$. Note that setting $\alpha$ to $19$ is for the analysis only and not necessary in practice. That is, $\alpha$ can be set to a much smaller value, e.g., $\alpha\!=\!1$, which means that SVRG-SD can use a much larger step size, e.g., $\eta\!=\!1/L$ for SVRG-SD vs.\ $\eta\!=\!1/(10L)$ for SVRG. Thus, SVRG-SD can significantly improve the convergence rate of SVRG and Prox-SVRG in practice, which will be confirmed by our experiments in Section~\ref{sec6}.

\subsection{Convergence Property for Non-SC Problems}
Unlike most of the stochastic variance reduction methods~\cite{johnson:svrg,xiao:prox-svrg}, including SVRG and Prox-SVRG, which only have the convergence guarantees for some SC problems, the convergence result of SVRG-SD for the non-SC case is also provided, as shown below.

\begin{theorem}[Non-SC]
\label{coro2}
Suppose each $f_{i}(\cdot)$ is $L$-smooth. Setting $\alpha\!=\!19$, $\sigma\!=\!1/2$, and $m$ sufficiently large, then
\begin{equation*}
\begin{split}
&\mathbb{E}[F(\overline{x})-F(x^{*})]\leq\frac{171L}{(16\!+\!40\widehat{\beta})mS}\|x^{*}-\widetilde{x}^{0}\|^{2}\\
&\quad+\!\left(\frac{9}{(4\!+\!8\widehat{\beta})mS}\!+\!\frac{1}{(2\!+\!4\widehat{\beta})S}\right)\!\left[F(\widetilde{x}^{0})\!-\!F(x^{*})\right]\!.
\end{split}
\end{equation*}
\end{theorem}

The proof of Theorem \ref{coro2} is provided in the Supplementary Material. The constant $\widehat{\beta}\geq0$ is from the sufficient decrease strategy, which thus implies that the convergence bound in Theorem~\ref{coro2} can be further improved using our sufficient decrease strategy with an even larger $\widehat{\beta}$. Theorem~\ref{coro2} shows that SVRG-SD attains the convergence rate of $\mathcal{O}(1/S)$ for the non-strongly convex objective function \eqref{equ1}. Although SVRG-SD is guaranteed to have a slower theoretical convergence rate than the accelerated stochastic variance reduction method, Katyusha~\cite{zhu:Katyusha}, whose convergence rate is $\mathcal{O}(1/S^2)$, SVRG-SD usually converges much faster than Katyusha in practice (see Section~\ref{sec63} for details).

\begin{algorithm}[t]
\caption{SAGA-SD}
\label{alg12}
\renewcommand{\algorithmicrequire}{\textbf{Input:}}
\renewcommand{\algorithmicensure}{\textbf{Initialize:}}
\renewcommand{\algorithmicoutput}{\textbf{Output:}}
\begin{algorithmic}[1]
\REQUIRE the number of epochs $S$, the number of iterations $m$ per epoch, and the step size $\eta$.\\
\ENSURE $\widetilde{x}^{0}$.
\FOR{$s=1,2,\ldots S$}
\STATE {$\,x^{s}_{0}\!=\widehat{x}^{s}_{0}\!=\widetilde{x}^{s-\!1}$;}
\FOR{$k=1,\ldots,m$}
\STATE {Pick $i^{s}_{k}$ uniformly at random from $[n]$;}
\STATE {Take $\phi^{k}_{i^{s}_{k}}\!=x^{s}_{k-\!1}$ and store $\nabla\! f_{i^{s}_{k}}\!(\phi^{k}_{i^{s}_{k}})$ in the table;}
\STATE {$\!\!\!\!\!\!\!\!\!\widetilde{\nabla}\!f_{i^{s}_{k}}\!(x^{s}_{k\!-\!1})\!\!=\!\!\nabla\! f_{i^{s}_{k}}\!(x^{s}_{k\!-\!1})\!\!-\!\!\nabla\! f_{i^{s}_{k}}\!(\phi^{k\!-\!1}_{i^{s}_{k}})\!\!+\!\!\frac{1}{n}\!\sum_{j}\!\!\nabla\! f_{j}(\phi^{k\!-\!1}_{j})$;}
\STATE {$y^{s}_{k}=\textrm{prox}^{r}_{\eta}\!\left(x^{s}_{k-1}-\eta\widetilde{\nabla} f_{i^{s}_{k}}(x^{s}_{k-1})\right)$;}
\STATE {$\!\theta_{k}\!\!=\! \arg\min_{\theta\in\mathbb{R}}F(\theta x^{s}_{k-\!1})\!\!+\!\!\frac{\zeta(1\!-\!\theta)^2}{2}\!\|\nabla\!f_{i^{s}_{k}}\!(x^{s}_{k-\!1})\!\!-\!\!\nabla\! f_{i^{s}_{k}}\!(\phi^{k\!-\!1}_{i^{s}_{k}})\|^{2}$;}
\STATE {$x^{s}_{k}=y^{s}_{k}+(1\!-\!\sigma)(\widehat{x}^{s}_{k}-\widehat{x}^{s}_{k-1})$\: and\: $\widehat{x}^{s}_{k}=\theta_{k}x^{s}_{k-\!1}$;}
\ENDFOR
\STATE {$\widetilde{x}^{s}=\frac{1}{m}\!\sum^{m}_{k=1}\!\widehat{x}^{s}_{k}$;}
\ENDFOR
\OUTPUT $\overline{x}=\widetilde{x}^{S}$
\end{algorithmic}
\end{algorithm}

\section{Extensions and Implementations}
\label{sec5}

\subsection{SAGA-SD}
As mentioned above, we can use the proposed sufficient decrease technique to accelerate other stochastic variance reduction methods, e.g., SAGA~\cite{defazio:saga}, and present a novel variant of SAGA with sufficient decrease (called SAGA-SD), as shown in Algorithm~\ref{alg12}. Like SVRG-SD, SAGA-SD is also a two-stage algorithm, whereas the original SAGA is a single-stage algorithm. More specifically, for SAGA-SD, the cost function with respect to $\theta$ in \eqref{equ5} can be revised by simply replacing $\nabla\! f_{i^{s}_{k}}\!(\widetilde{x}^{s-\!1})$ with $\nabla\! f_{i^{s}_{k}}\!(\phi^{k-\!1}_{i^{s}_{k}})=g^{k-\!1}_{i_{k}}$ defined in~\eqref{equ31}. Moreover, Property~\ref{prop11} can be extended for SAGA-SD by setting $\widetilde{p}_{i^{s}_{k}}\!=\!\nabla\!f_{i^{s}_{k}}\!(x^{s}_{k-\!1})\!-\!\nabla\! f_{i^{s}_{k}}\!(\phi^{k-\!1}_{i^{s}_{k}})$, as well as for other stochastic variance reduction algorithms such as SAG. The main differences between SVRG-SD and SAGA-SD are the stochastic gradient estimators in \eqref{equ91} and \eqref{equ31}, and the update rule of the sufficient decrease coefficient in \eqref{equ5}. For the convergence analysis of SAGA-SD, we refer to~\cite{shang:svrgsd}.

Similar to Katyusha~\cite{zhu:Katyusha}, SVRG-SD and SAGA-SD trivially extend to the mini-batch setting. As suggested in~\cite{frostig:sgd} and \cite{lin:vrsg}, we can add a proximal term into a non-strongly convex objective function $F(x)$ as follows: $F_{\tau}(x,y)\!=\!f(x)\!+\!\frac{\tau}{2}\|x\!-\!y\|^{2}\!+\!r(x)$, where $\tau\!\geq\!0$ is a constant that can be determined as in~\cite{lin:vrsg,frostig:sgd}, and $y\!\in\! \mathbb{R}^{d}$ is a proximal point. Then the condition number of this proximal function $F_{\tau}(x,y)$ can be much smaller than that of the original function $F(x)$, if $\tau$ is sufficiently large. However, adding the proximal term may degrade the performance of the involved algorithms both in theory and in practice~\cite{zhu:univr}. Therefore, we directly use SVRG-SD and SAGA-SD to solve the non-strongly convex objective function \eqref{equ1}. If some component functions in \eqref{equ1} are non-smooth (e.g., support vector machine), we can use the proximal operator oracle \cite{zhu:box} or the Nesterov's smoothing \cite{nesterov:smooth} technique to smoothen them, and thereby obtain the smoothed approximations of the functions $f_{i}(x)$.

\subsection{Efficient Implementations}
Both \eqref{equ10} and \eqref{equ11} require the calculation of $b^{T}\!A$, thus we need to precompute and save $b^{T}\!A$ in the initial stage. To further reduce the computational complexity of $\|Ax^{s}_{k-\!1}\|^{2}$ in \eqref{equ10} and \eqref{equ11}, we use the fast partial singular value decomposition (SVD) to obtain the best rank-$r$ approximation $U_{r}S_{r}V^{T}_{r}$ to $A$ and save $S_{r}V^{T}_{r}$. Then $\|Ax^{s}_{k-\!1}\|\!\approx\!\|S_{r}V^{T}_{r}x^{s}_{k-\!1}\|$. In practice, e.g., in our experiments, $r$ can be set to a small number to capture 99.5\% of the spectral energy of the data matrix $A$, e.g., $r\!=\!10$ for the Covtype data set, similar to inexact line search methods~\cite{more:ls} for deterministic optimization.

For high-dimensional sparse data $A$ (e.g., Rcv1), we use the lazy computing trick for the calculation of $\|Ax^{s}_{k-\!1}\|$, as well as $b^{T}\!A$, instead of the partial SVD. That is, we calculate $\|Ax^{s}_{k-\!1}\|$ and $b^{T}\!A$ only for the non-zero dimensions of each sample, rather than all dimensions. In particular, we can introduce the lazy update tricks in~\cite{koneeny:mini} to our algorithms, and perform the update steps only for the non-zero dimensions of each example. That is, the average per-iteration complexity can be improved from $O(d)$ to $O(d')$ as stated below, where $d'\!\leq\!d$ is the sparsity of feature vectors.

The time complexity of each inter-iteration in SVRG-SD, as well as SAGA-SD, with full sufficient decrease is $O(rd)$, which is a little higher than that of SVRG. In fact, we can just randomly select only a small fraction (e.g., $1/10^3$) of stochastic gradient iterations in each epoch to update with sufficient decrease, while the remaining iterations are executed without sufficient decrease technique, i.e., $\widehat{x}^{s}_{k}\!=\!x^{s}_{k-\!1}$. Let $m_{1}$ be the number of iterations with our sufficient decrease technique in each epoch. By fixing $m_{1}\!=\!\lfloor m/10^3\rfloor$ and thus without increasing parameters tuning difficulties, SVRG-SD\footnote{Note that SVRG-SD with partial sufficient decrease possesses the similar convergence properties as SVRG-SD with full sufficient decrease because Property~\ref{prop11} still holds in the case when $\theta_{k}\!=\!1$.} can always converge much faster than its counterpart, SVRG, as shown in Figure~\ref{fig1} (similar results are also observed for SAGA-SD, as shown in the Supplementary Materials). It is easy to see that our algorithms are very robust with respect to the choice of $m_{1}$, and achieve average per-iteration time complexity as low as the original SVRG and SAGA. Thus, we mainly consider SVRG-SD and SAGA-SD with partial sufficient decrease for real-world machine learning applications.

\begin{figure}[t]
\centering
\subfigure[Gap vs.\ number of passes]{\includegraphics[width=0.492\columnwidth]{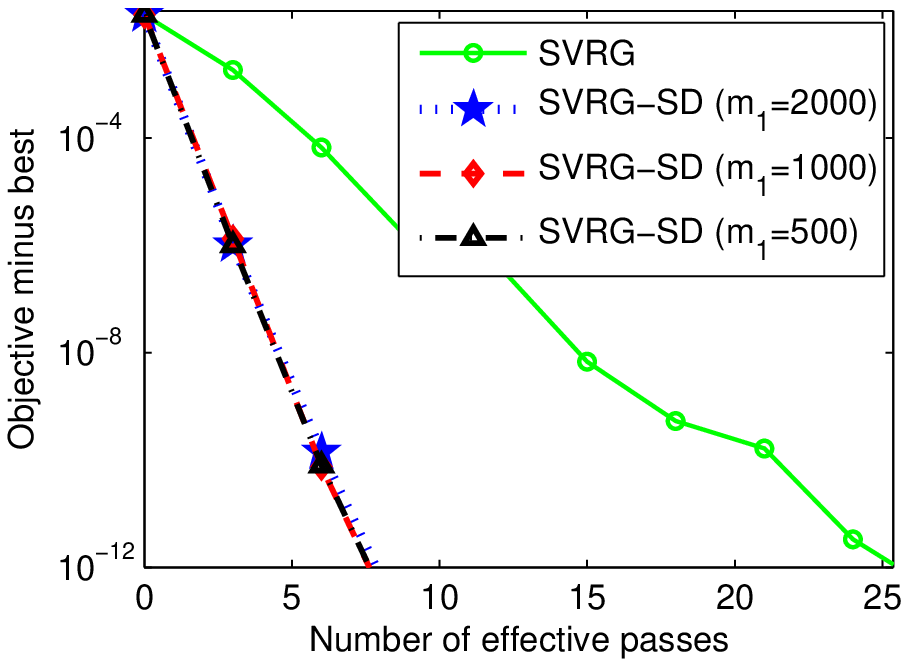}}
\subfigure[Gap vs.\ running time]{\includegraphics[width=0.492\columnwidth]{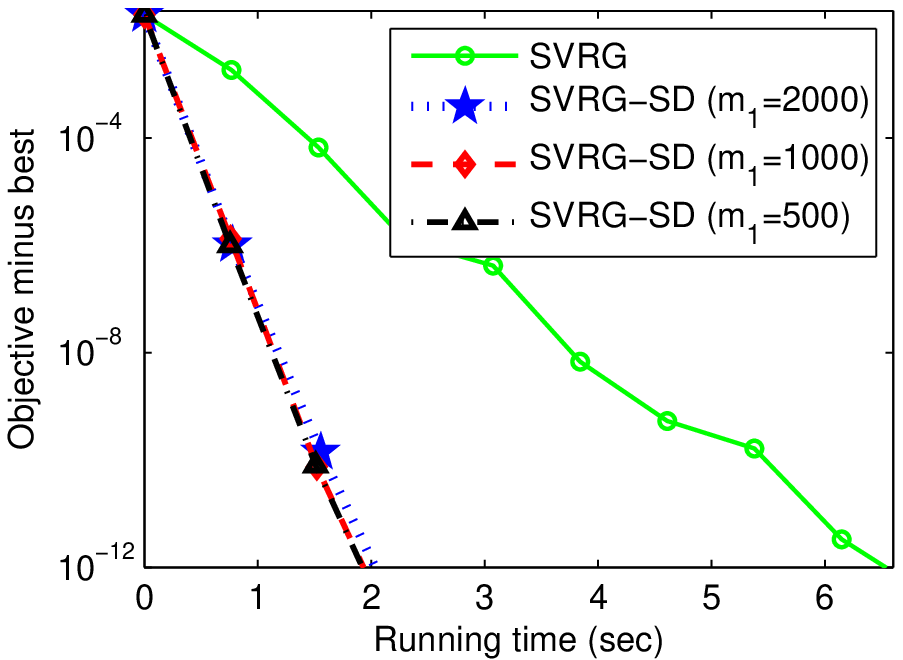}}
\caption{Comparison of SVRG and SVRG-SD with different values of $m_{1}$ for solving ridge regression problems on the Covtype dataset.}
\label{fig1}
\end{figure}

\begin{figure*}[t]
\centering
\includegraphics[width=0.509\columnwidth]{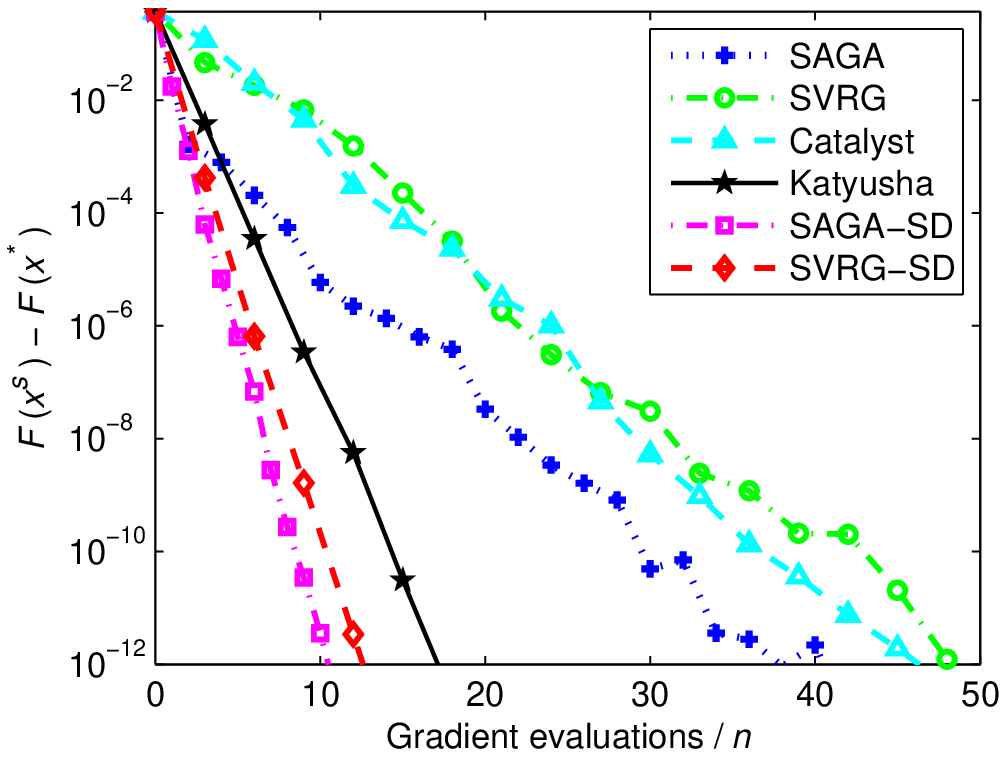}
\includegraphics[width=0.509\columnwidth]{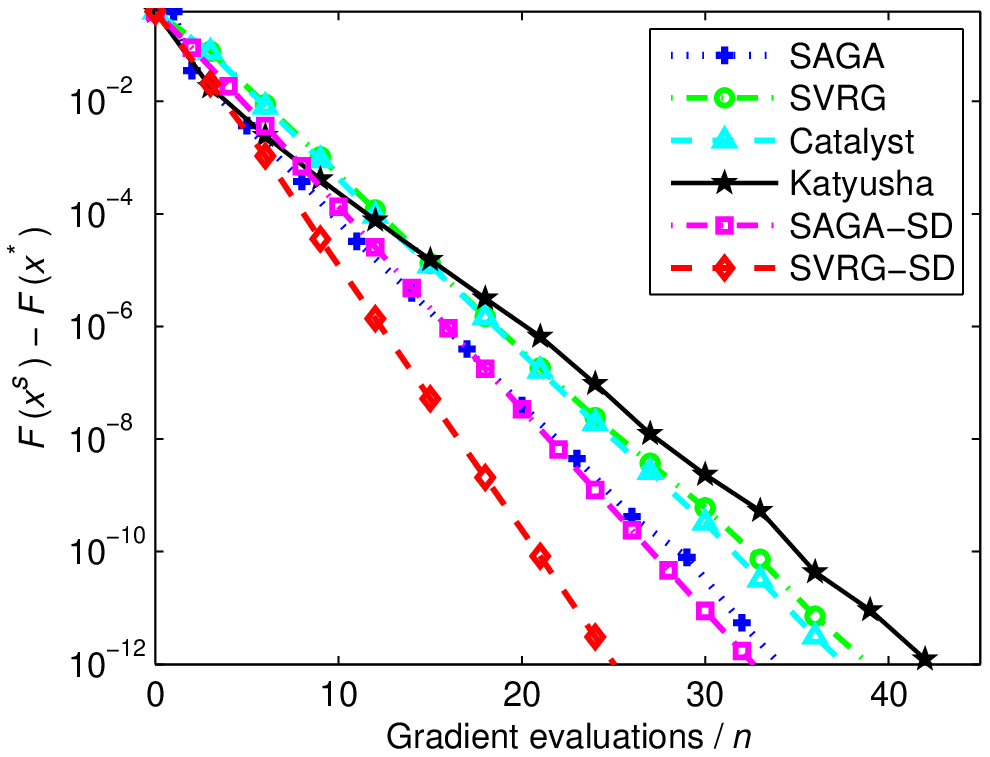}
\includegraphics[width=0.509\columnwidth]{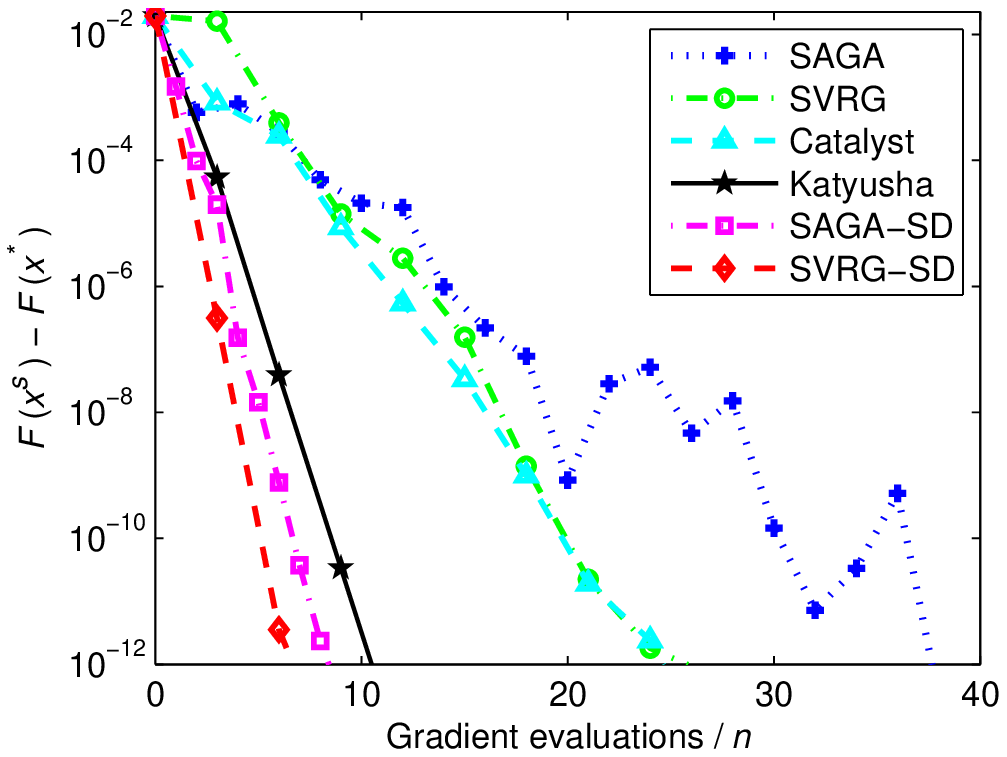}
\includegraphics[width=0.509\columnwidth]{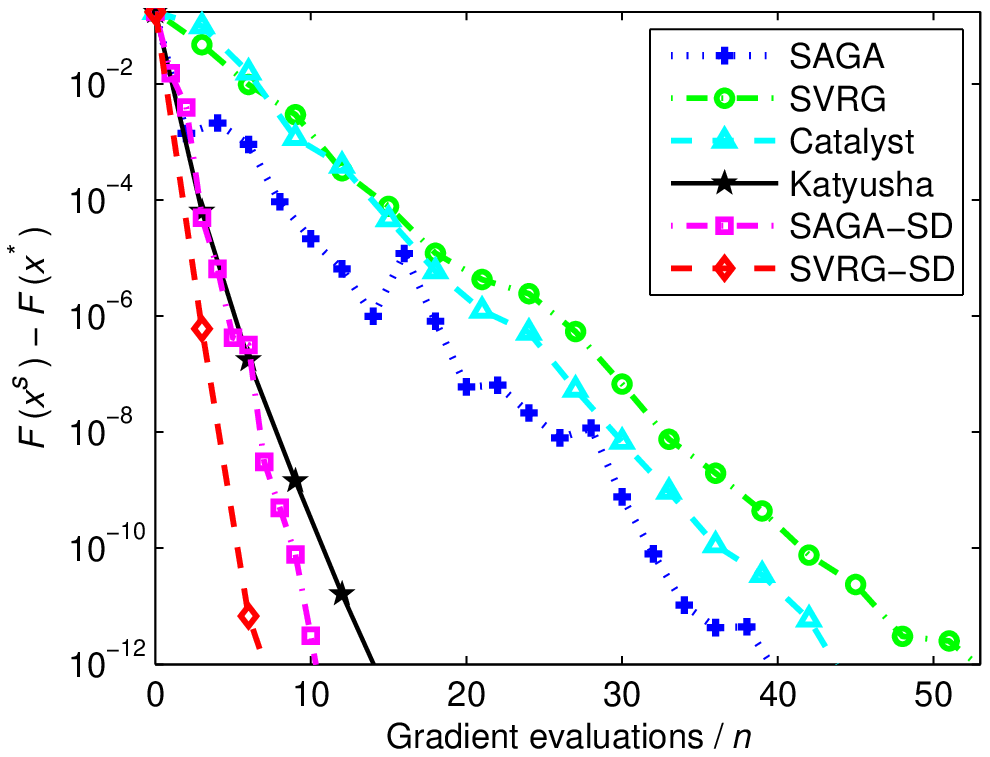}

\subfigure[Ijcnn1]{\includegraphics[width=0.509\columnwidth]{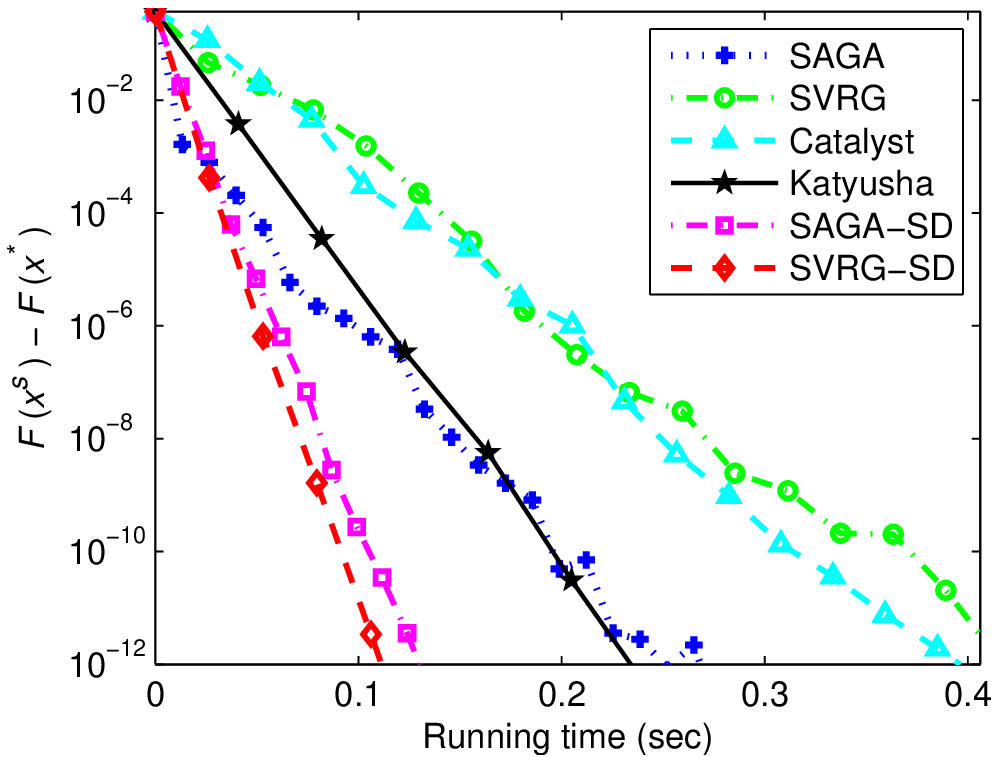}}
\subfigure[Rcv1]{\includegraphics[width=0.509\columnwidth]{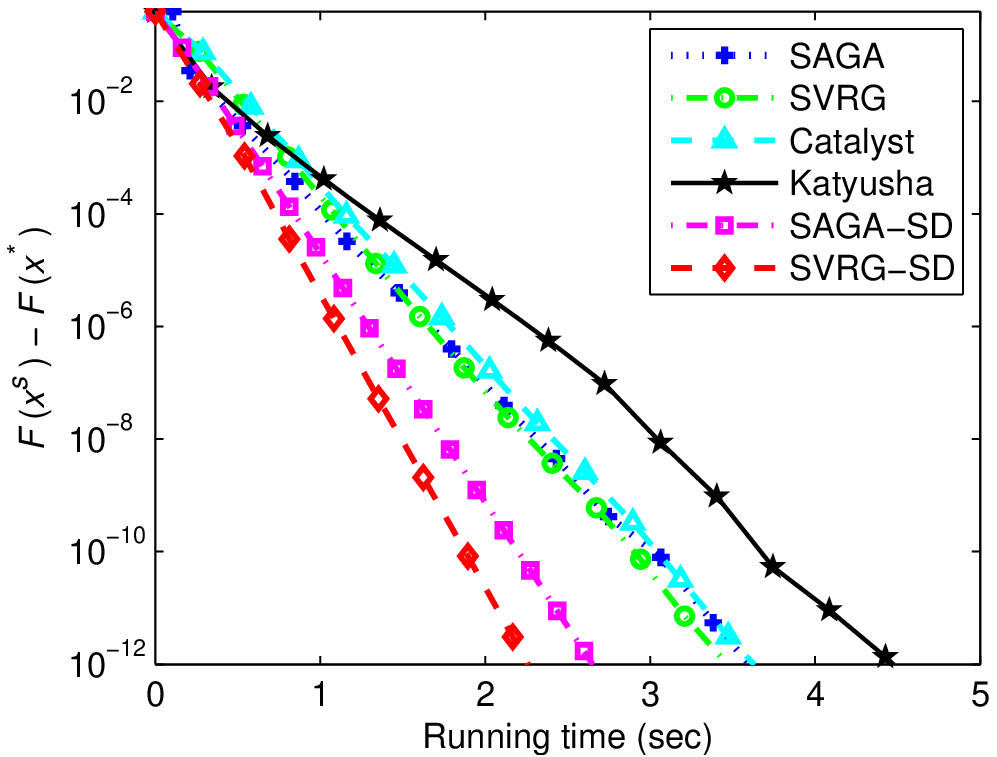}\label{fig2b}}
\subfigure[Covtype]{\includegraphics[width=0.509\columnwidth]{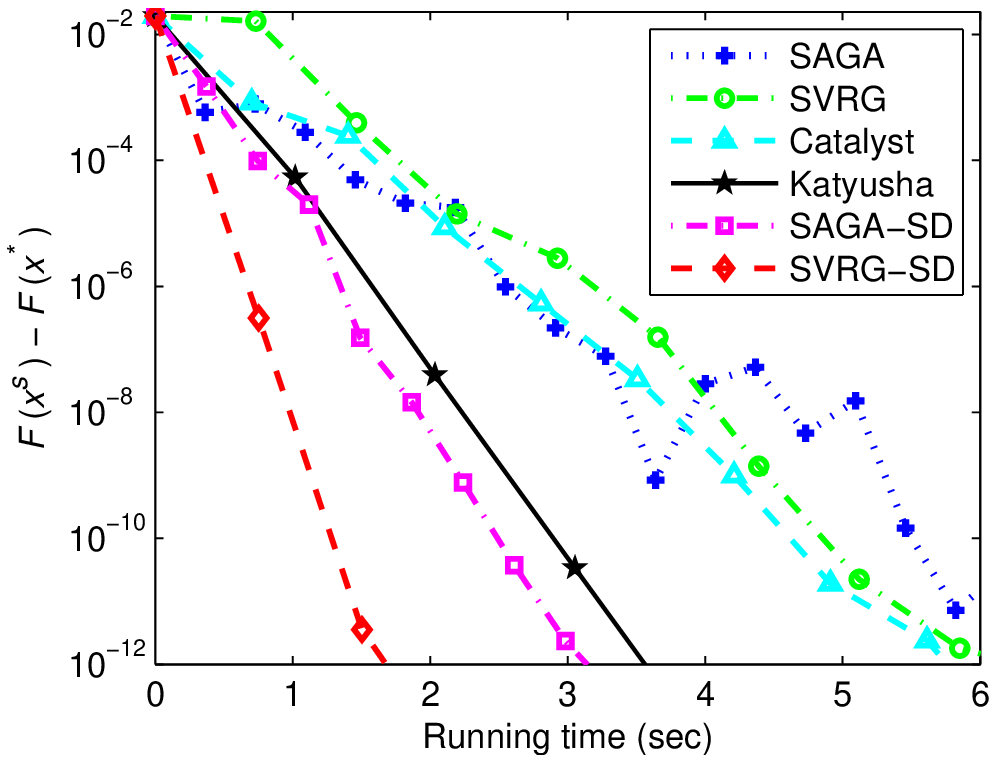}}
\subfigure[SUSY]{\includegraphics[width=0.509\columnwidth]{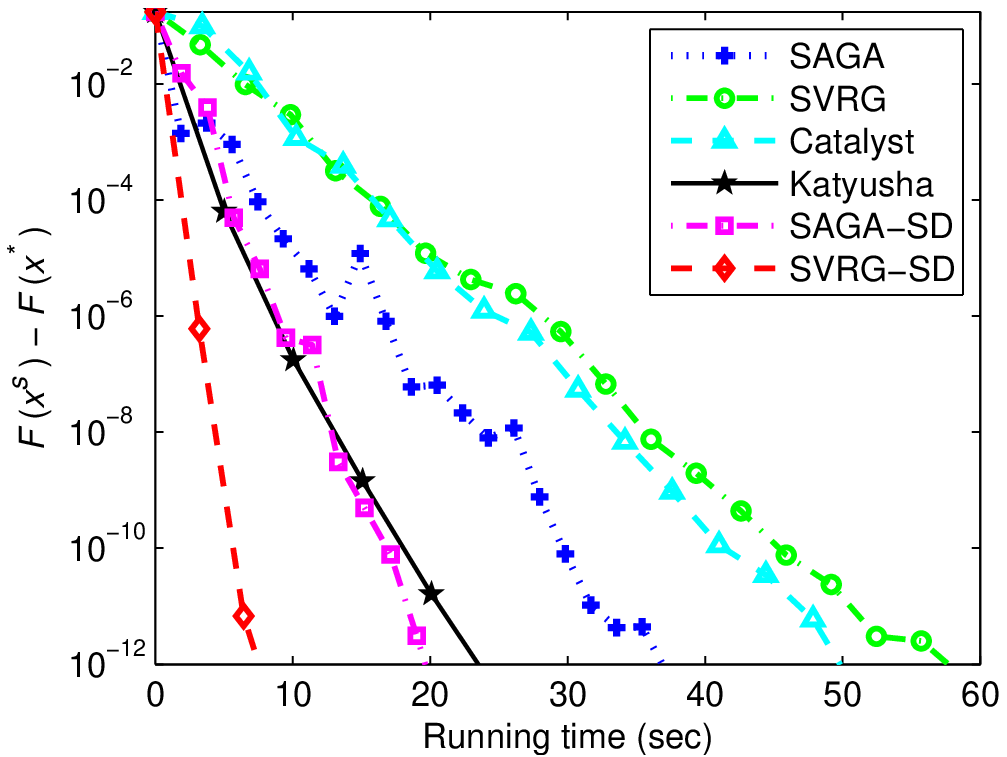}}
\caption{Comparison of the stochastic variance reduction methods for solving ridge regression problems ($\lambda_{1}\!=\!10^{-4}$) on Ijcnn1, Rcv1, Covtype, and SUSY. The vertical axis is the objective value minus the minimum, and the horizontal axis denotes the number of effective passes over the data (top) or running time (bottom).}
\label{fig2}
\end{figure*}

\section{Experiments}
\label{sec6}
In this section, we evaluate the performance of SVRG-SD and SAGA-SD, and compare their performance with their counterparts including SVRG~\cite{johnson:svrg}, its proximal variant (Prox-SVRG)~\cite{xiao:prox-svrg}, and SAGA~\cite{defazio:saga}. Moreover, we also report the performance of the well-known accelerated stochastic variance reduction methods, Catalyst~\cite{lin:vrsg} and Katyusha~\cite{zhu:Katyusha}.

\begin{table}[!th]
\centering
\caption{Summary of dense and sparse data sets.}
\label{tab_sim1}
\setlength{\tabcolsep}{6.9pt}
\linespread{1.36}
\begin{tabular}{lccc}
\hline
\ Data sets   & Sizes $n$    & Dimensions $d$  & Sparsity \\
\hline
\ Ijcnn1      & 49,990         & 22             & 59.09\% \\
\ Rcv1        & 20,242         & 47,236         & 0.16\% \\
\ Covtype     & 581,012        & 54             & 22.12\% \\
\ SUSY        & 5,000,000      & 18             & 98.82\% \\
\hline
\end{tabular}
\end{table}

\subsection{Experimental Setup}
In our experiments, we used the four publicly available data sets: Ijcnn1, Rcv1, Covtype, and SUSY, all of which can be downloaded from the LIBSVM Data website. Note that each sample of all the date sets was normalized so that they have unit length as in~\cite{xiao:prox-svrg}, which leads to the same upper bound on the Lipschitz constants $L_{i}$, i.e., $L\!=\!L_{i}$ for all $i\!=\!1,\ldots,n$. This step is for comparison only and not necessary in practice. As suggested in~\cite{johnson:svrg,xiao:prox-svrg,zhu:Katyusha}, the epoch length is set to $m\!=\!2n$ for the stochastic variance reduced methods, SVRG~\cite{johnson:svrg}, Prox-SVRG~\cite{xiao:prox-svrg}, Catalyst~\cite{lin:vrsg}, and Katyusha~\cite{zhu:Katyusha}, as well as SVRG-SD. In addition, unlike SAGA~\cite{defazio:saga}, we fixed $m\!=\!n$ for each epoch of SAGA-SD. Then the only parameter we have to tune by hand is the step size, $\eta$. More specifically, we select step sizes from $\{10^{j},2.5\times10^{j},5\times10^{j},7.5\times10^{j},10^{j+1}\}$ as in~\cite{zhu:Katyusha}, where $j\!\in\!\{-2,-1,0\}$. Each of these methods had its step size parameter chosen so as to give the fastest convergence. Since Katyusha has a much higher per-iteration complexity than SVRG, we compare their performance in terms of both the number of effective passes and running time (seconds), where computing a single full gradient or evaluating $n$ component gradients is considered as one effective pass over the data. For fair comparison, we implemented all the methods in C++ with a Matlab interface (all codes are made available, see link in the Supplementary Materials), and performed all the experiments on a PC with an Intel i5-2400 CPU and 16GB RAM.

\subsection{Ridge Regression}
In this part, we focus on the following ridge regression (i.e., $\lambda_{2}\!\equiv\!0$) problem as the SC example:
\vspace{-2mm}
\begin{equation}\label{equ151}
\min_{x\in\mathbb{R}^{d}}\;\frac{1}{2n}\sum_{i=1}^n (a_i^Tx - b_i)^2+\frac{\lambda_{1}}{2}\|x\|^{2}+\lambda_{2}\|x\|_{1},
\vspace{-1mm}
\end{equation}
where $\lambda_{1},\lambda_{2}\!\geq\!0$ are the regularization parameters.

\begin{figure}[t]
\centering
\includegraphics[width=0.4935\columnwidth]{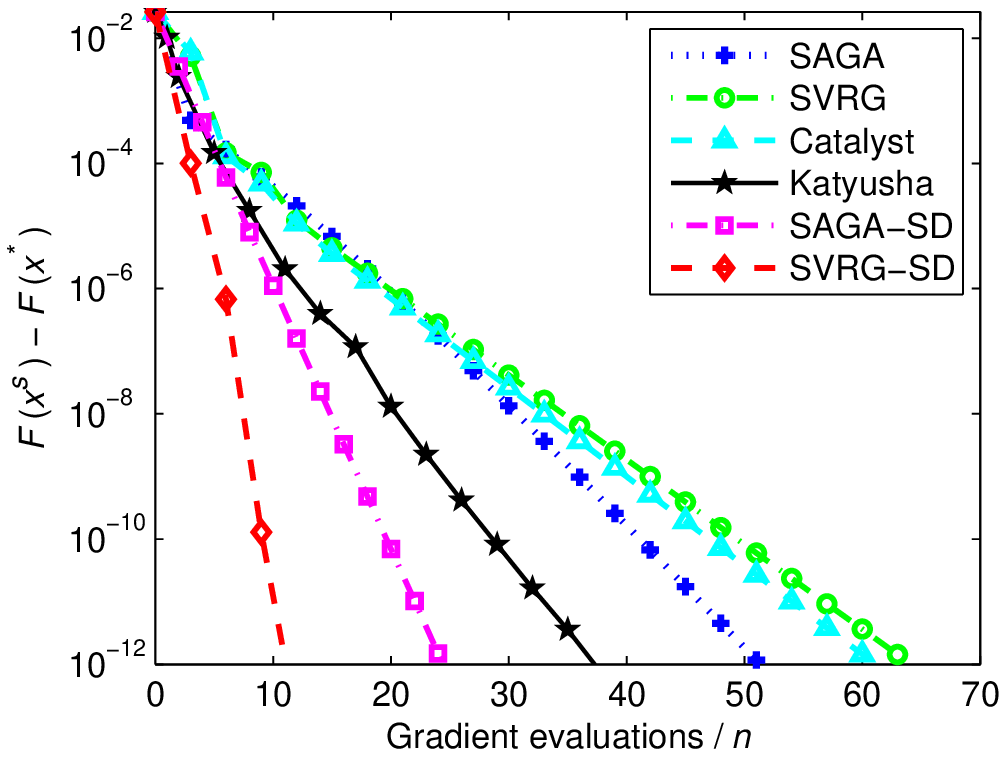}\includegraphics[width=0.4935\columnwidth]{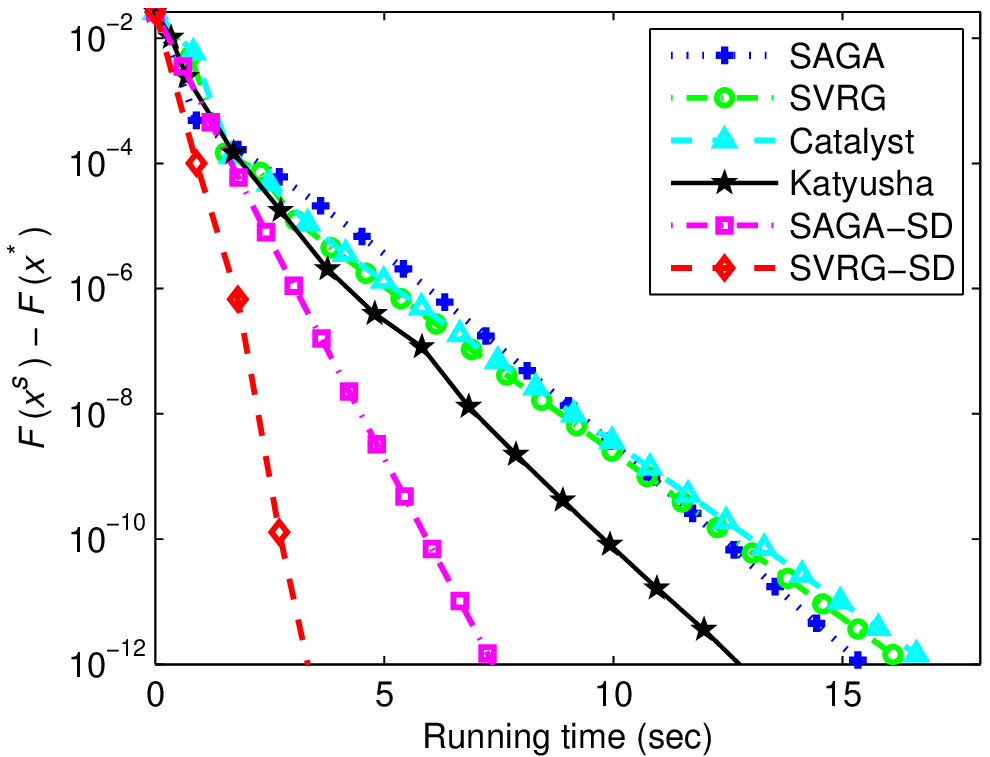}
\caption{Comparison of the stochastic variance reduction methods for solving ridge regression problems with $\lambda_{1}\!=\!10^{-6}$ on Covtype.}
\label{fig3}
\end{figure}

Figure~\ref{fig2} shows how the objective gap, i.e., $F(x^{s})\!-\!F(x^{*})$, of all these algorithms decreases for ridge regression problems with the regularization parameters $\lambda_{1}\!=\!10^{-4}$ and $\lambda_{2}\!\equiv\!0$ (more results are given in the Supplementary Material). For SVRG-SD and SAGA-SD, we set $\sigma\!=\!1/2$ on the four data sets. As seen in Figure~\ref{fig2}, SVRG-SD and SAGA-SD achieve consistent speedups for all the data sets, and significantly outperform their counterparts, SVRG and SAGA, in all the settings. This confirms that our stochastic sufficient decrease technique is able to accelerate SVRG and SAGA. Impressively, SVRG-SD and SAGA-SD usually converge much faster than the well-known accelerated methods, Catalyst and Katyusha, which further justifies the effectiveness of our sufficient decrease technique. As SVRG-SD and SAGA-SD have a much lower per-iteration complexity than Katyusha, they have more obvious advantage over Katyusha in terms of running time, especially for the case of sparse data (e.g., Rcv1), as shown in Fig.\ \ref{fig2b}.

\begin{figure*}[t]
\centering
\includegraphics[width=0.509\columnwidth]{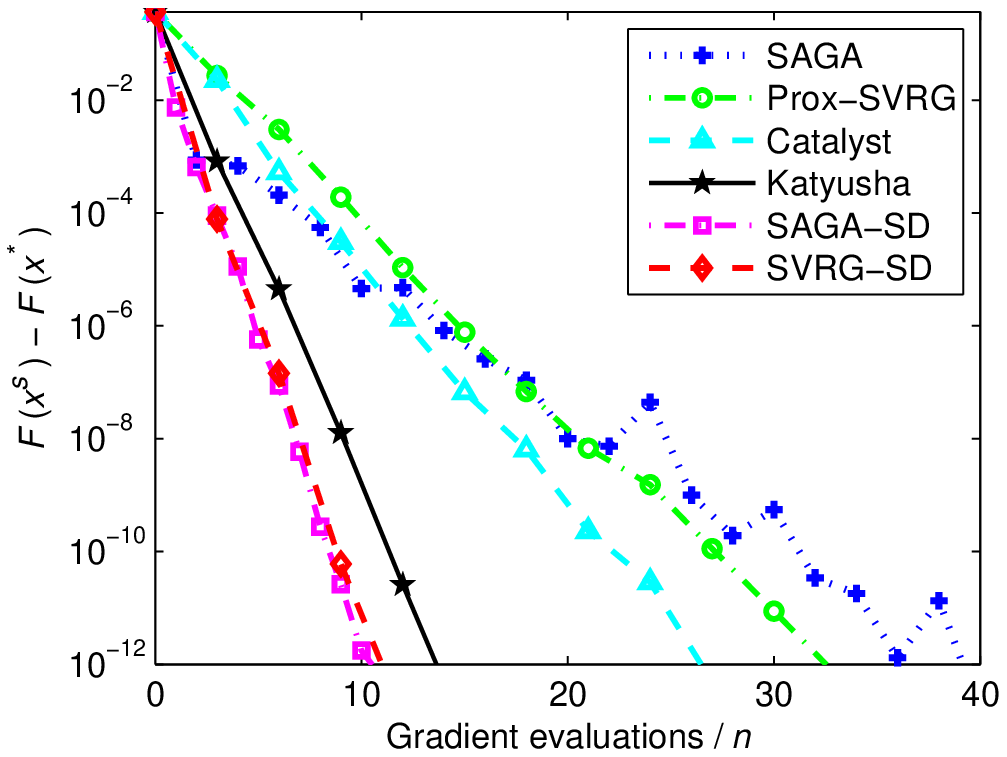}\includegraphics[width=0.509\columnwidth]{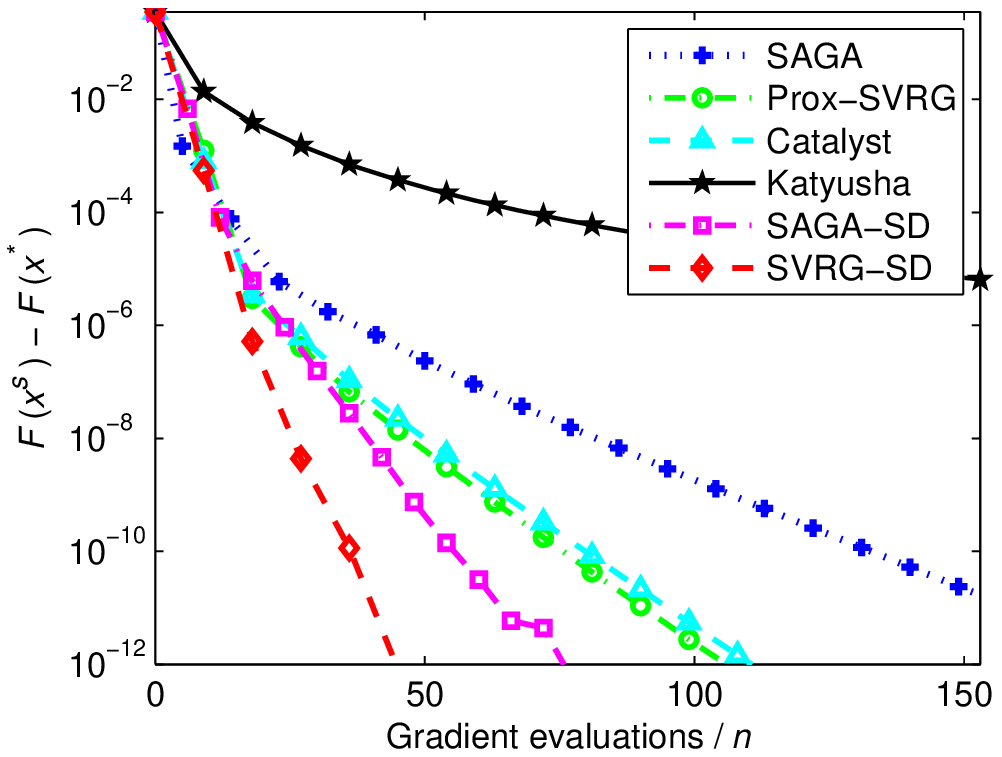}\includegraphics[width=0.509\columnwidth]{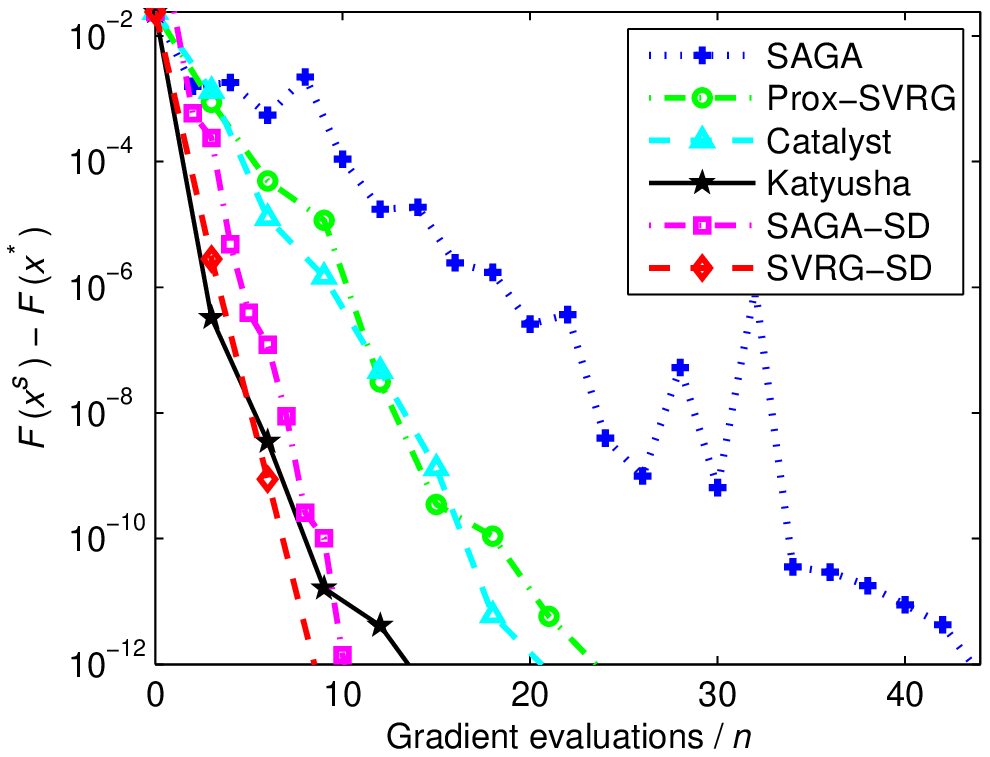}\includegraphics[width=0.509\columnwidth]{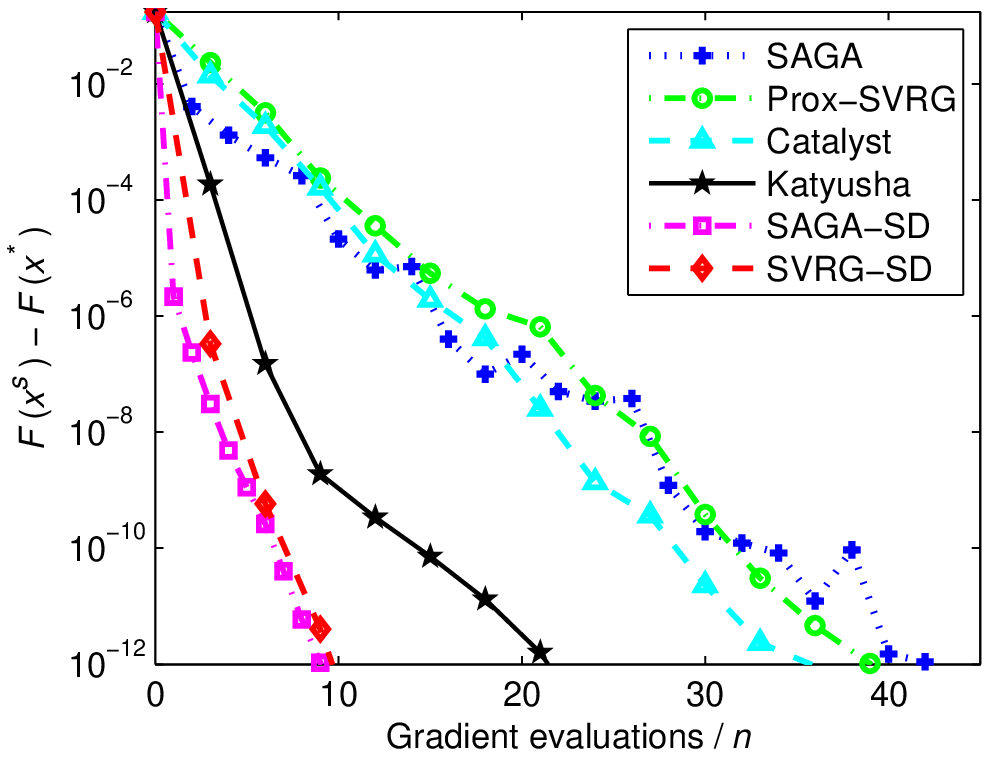}

\subfigure[Lasso, $\lambda_{2}\!=\!10^{-4}$]{\includegraphics[width=0.509\columnwidth]{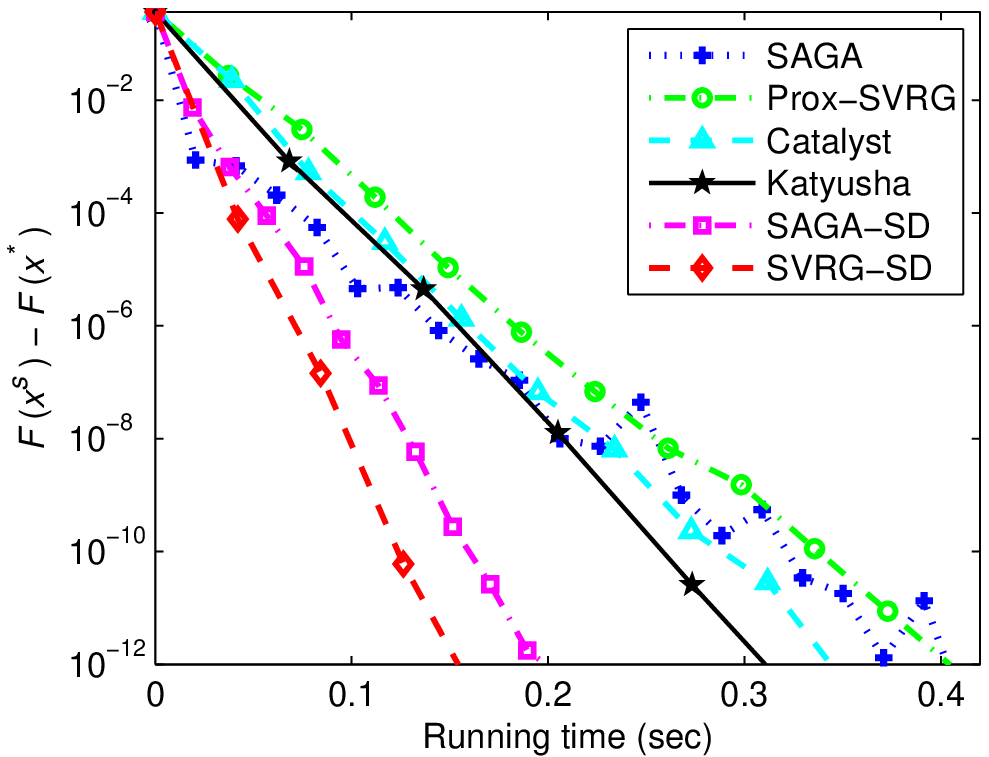}\includegraphics[width=0.509\columnwidth]{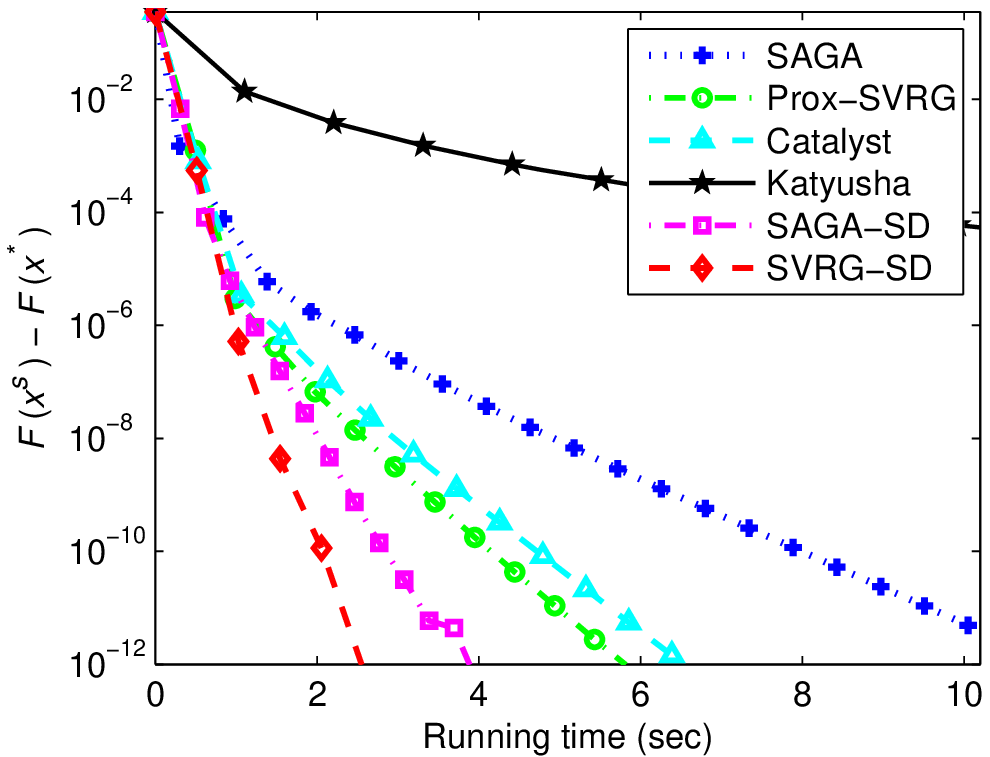}}
\subfigure[Elastic-net, $\lambda_{1}\!=\!\lambda_{2}\!=\!10^{-5}$]{\includegraphics[width=0.509\columnwidth]{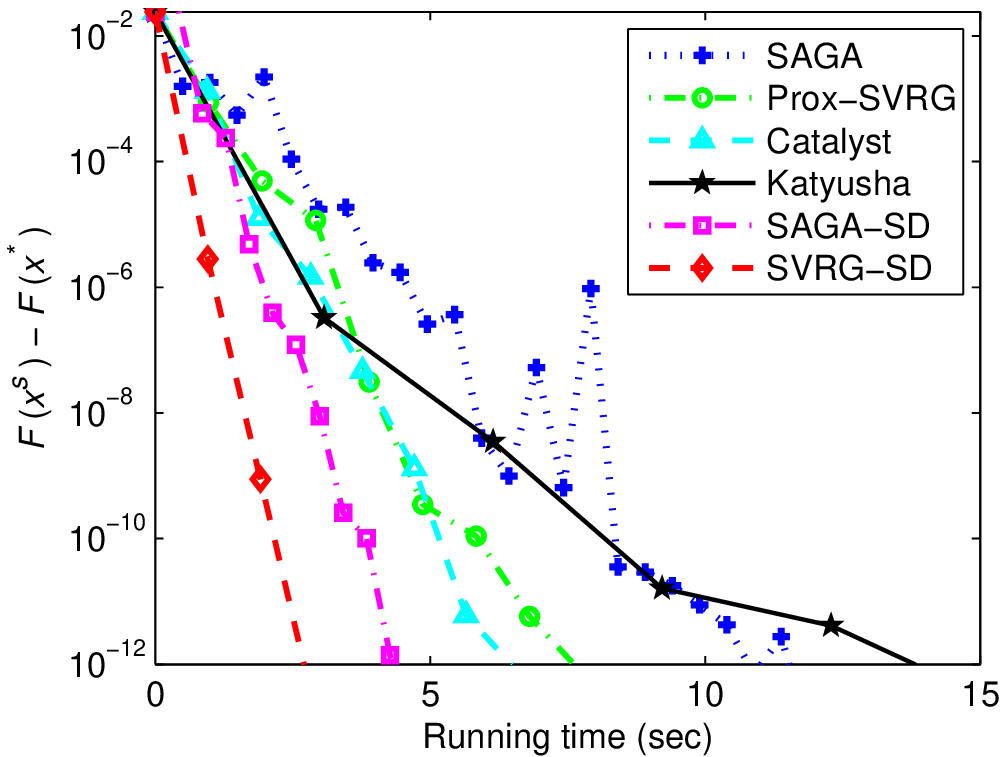}\includegraphics[width=0.509\columnwidth]{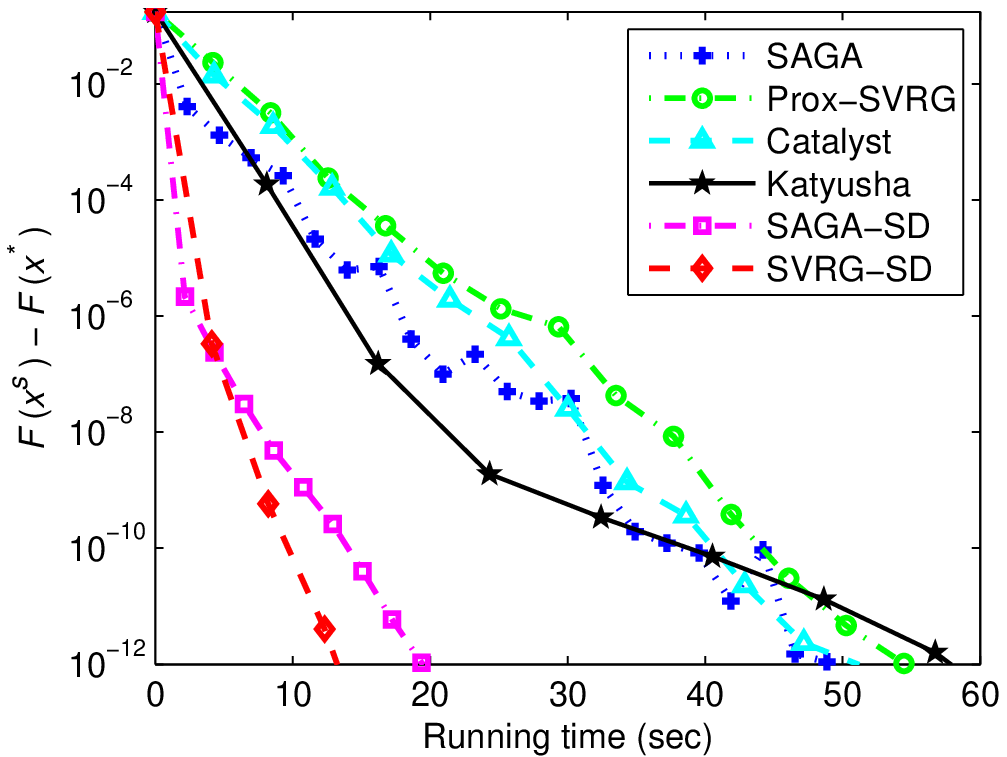}}
\caption{Comparison of all the stochastic methods for solving Lasso and elastic-net problems on Ijcnn1 (first column), Rcv1 (second column), Covtype (third column), and SUSY (last column).}
\label{fig4}
\end{figure*}

We also compared the performance of all these methods, when the regularization parameter is relatively small, as shown in Figure~\ref{fig3}. The result shows that Katyusha, SVRG-SD and SAGA-SD converge significantly faster than the other methods. In particular, SVRG-SD and SAGA-SD also outperform the well-known accelerated methods, Catalyst and Katyusha, which further verifies the importance of our sufficient decrease technique for stochastic optimization.

\subsection{Lasso and Elastic-Net}
\label{sec63}
Finally, we conducted some experiments for the generalized regression problem~\eqref{equ151}, including Lasso (i.e., $\lambda_{1}\!\equiv\!0$) and elastic-net (i.e., $\lambda_{1}\!\neq\!0$ and $\lambda_{2}\!\neq\!0$) problems as non-SC and non-smooth examples.

We plot some representative results in Figure~\ref{fig4} (see the Supplementary Material for more results), which show that SVRG-SD and SAGA-SD significantly outperform their counterparts (i.e., Prox-SVRG and SAGA) in all the settings, and are considerably better than the accelerated methods, Catalyst and Katyusha, in most cases. This empirically verifies that our sufficient decrease technique allows us to take much larger step sizes for SVRG-SD and SAGA-SD than the original SVRG/Prox-SVRG and SAGA (e.g., $\eta\!=\!1$ for SVRG-SD vs.\ $\eta\!=\!0.4$ for SVRG), and can accelerate them for solving both SC and non-SC objective functions.

\section{Conclusions \& Future Work}
To the best of our knowledge, this is the first work to design an efficient sufficient decrease technique for stochastic optimization. Moreover, we proposed two different schemes for Lasso and ridge regression to efficiently update the coefficient $\theta$, which takes the important decisions to shrink, expand or move in the opposite direction. This is very different from adaptive learning rate methods, e.g., \cite{kingma:sgd}, and line search methods, e.g., \cite{mahsereci:sgd}, all of which cannot address the issue in Section~\ref{sec31} whatever value the step size is. Unlike most stochastic variance reduction methods~\cite{johnson:svrg,shalev-Shwartz:sdca,xiao:prox-svrg}, which only have convergence guarantees for SC problems\footnote{\cite{hofmann:vrsg} proposed a simple SVRG version with convergence guarantees for both SC and non-SC problems as in SAGA.}, we provided the convergence guarantees of our algorithm for both SC and non-SC objective functions. Experimental results verified the effectiveness of our sufficient decrease technique for stochastic optimization, especially for the case when the regularization parameter is relatively small (or the condition number $\kappa\!:=\!L/\mu$ is relatively large). Naturally, it can also be used to further speed up accelerated methods such as~\cite{lin:vrsg,zhu:Katyusha,zhu:univr,shang:svrg}.

As each component function $f_{i}(\cdot)$ may have different degrees of smoothness, to select the random index $i^{s}_{k}$ from a non-uniform distribution is a much better choice than simple uniform random sampling~\cite{zhao:prox-smd}, as well as without-replacement sampling vs.\ with-replacement sampling~\cite{shamir:sgd}. On the practical side, both our algorithms tackle the non-SC and non-smooth problems directly, without using any quadratic regularizer as in~\cite{lin:vrsg,zhu:Katyusha}, as well as the proximal setting. Note that some asynchronous parallel and distributed variants~\cite{reddi:sgd,mania:svrg,pedregosa:saga,lee:dsgd} of stochastic variance reduction methods have also been proposed for such stochastic settings. We leave these variations out from our comparison and consider similar extensions to our stochastic sufficient decrease method as future work.

\subsubsection*{Acknowledgments}
We thank the reviewers for their valuable comments. This work was supported in part by Grants (CUHK 14206715 \& 14222816) from the Hong Kong RGC.

\bibliographystyle{unsrt}
\bibliography{nips2017}

\newpage

\twocolumn[

\aistatstitle{Supplementary Materials for ``Guaranteed Sufficient Decrease for Stochastic Variance Reduced Gradient Optimization''}

\aistatsauthor{ Fanhua Shang,\,\quad Yuanyuan Liu,\,\quad Kaiwen Zhou,\,\quad James Cheng,\,\quad Kelvin K.W. Ng}
\aistatsaddress{ Department of Computer Science and Engineering, The Chinese University of Hong Kong\footnotetext{Corresponding author.}}
\vspace{-3mm}
\aistatsauthor{ Yuichi Yoshida}
\aistatsaddress{ National Institute of Informatics, Tokyo, Japan} ]

\setcounter{lemma}{0}
\setcounter{algorithm}{3}
\setcounter{corollary}{1}

\linespread{1.39}
\vspace{23mm}
In this supplementary material, we give the detailed proofs for some lemmas, theorems and corollaries stated in the main paper. Moreover, we also report more experimental results for both of our algorithms on several dense and sparse data sets.
\vspace{2mm}

\section*{Notations}
Throughout this paper, $\|\!\cdot\!\|$ denotes the standard Euclidean norm, and $\|\!\cdot\!\|_{1}$ is the $\ell_{1}$-norm, i.e., $\|x\|_{1}\!=\!\sum^{d}_{i=1}\!|x_{i}|$. We denote by $\nabla\!f(x)$ the full gradient of $f(x)$ if it is differentiable, or $\partial f(x)$ the subdifferential of $f(\cdot)$ at $x$ if it is only Lipschitz continuous. Note that Assumption 2 is the general form for the two cases when $F(x)$ is smooth or non-smooth\footnote{Strictly speaking, when the function $F(\cdot)$ is non-smooth, $\vartheta\in \partial F(x)$; while $F(\cdot)$ is smooth, $\vartheta=\nabla F(x)$.}. That is, if $F(x)$ is smooth, the inequality in (12) in Assumption 2 becomes the following form:
\begin{displaymath}
F(y)\geq F(x)+\nabla F(x)(y-x)+\frac{\mu}{2}\|y-x\|^{2}.
\end{displaymath}

In the main paper, we assume that all component functions have the same smoothness parameter, $L$. In fact, we can extend the theoretical result for the case, when the gradients of all component functions have the same Lipschitz constant $L$, to the more general case, when some component functions $f_{i}(\cdot)$ have different degrees of smoothness.

\begin{definition}\label{def1}
The SVRG estimator in the mini-batch setting is defined as follows:
\begin{equation*}
\widetilde{\nabla}\! f_{I^{s}_{k}}(x^{s}_{k})=\frac{1}{b}\!\sum_{i\in I^{s}_{k}}\!\left[\nabla\!f_{i}(x^{s}_{k})\!-\!\nabla\! f_{i}(\widetilde{x}^{s-\!1})\right]\!+\!{\nabla}\! f(\widetilde{x}^{s-\!1})
\end{equation*}
where $I^{s}_{k}\!\subset\![n]$ is a mini-batch of size $b$.
\end{definition}
Using the above definition, our algorithms naturally generalize to the mini-batch setting.

\vspace{5mm}

\section*{Appendix A: Proof of Theorem 1}
Although the proposed SVRG-SD algorithm is a variant of SVRG, it is non-trivial to analyze its convergence property. Before proving Theorem 1, we first give the following lemma.

\begin{lemma}
\label{lemm1}
Let $x^{*}$ be the optimal solution of Problem (1), then the following inequality holds
\begin{equation*}
\begin{split}
&\mathbb{E}\!\left[\left\|\nabla\! f_{i^{s}_{k}}\!(x^{s}_{k-\!1})\!-\!\nabla\! f(x^{s}_{k-\!1})\!-\!\nabla\! f_{i^{s}_{k}}\!(\widetilde{x}^{s-\!1})\!+\!\nabla\! f(\widetilde{x}^{s-\!1})\right\|^{2}\right]\\
\leq&\, 4L\!\left[F(x^{s}_{k-\!1})-F(x^{*})+F(\widetilde{x}^{s-\!1})-F(x^{*})\right]\!.
\end{split}
\end{equation*}
\end{lemma}

Lemma~\ref{lemm1} provides the upper bound on the expected variance of the variance reduced gradient estimator in (9) (i.e., the SVRG estimator independently introduced in~\cite{johnson:svrg,zhang:svrg}), which satisfies $\mathbb{E}[\widetilde{\nabla}\! f_{i^{s}_{k}}(x^{s}_{k-\!1})]\!=\!\nabla\!f(x^{s}_{k-\!1})$. This lemma is essentially identical to Corollary 3.5 in~\cite{xiao:prox-svrg} and Lemma A.2 in~\cite{zhu:univr2}. In addition, the upper bound on the variance of $\widetilde{\nabla}\!f_{i^{s}_{k}}(x^{s}_{k})$ can be extended to the mini-batch setting as in~\cite{koneeny:mini}.

\vspace{2mm}
Using Lemma~\ref{lemm1}, we immediately get the following result, which is useful in our convergence analysis.
\begin{corollary}\label{coro4}
For any $\alpha\geq\beta>0$, the following inequality holds
\begin{equation*}
\begin{split}
&\alpha\mathbb{E}\!\left[\left\|\nabla\! f_{i^{s}_{k}}\!(x^{s}_{k-\!1})-\!\nabla\! f(x^{s}_{k-\!1})\!-\!\nabla\! f_{i^{s}_{k}}\!(\widetilde{x}^{s-\!1})\!+\!\nabla\! f(\widetilde{x}^{s-\!1})\right\|^{2}\right]\\
&-\beta\mathbb{E}\!\left[\left\|\nabla\! f_{i^{s}_{k}}\!(x^{s}_{k-\!1})\!-\!\nabla\! f_{i^{s}_{k}}\!(\widetilde{x}^{s-\!1})\right\|^{2}\right]\\
\leq\, &4L(\alpha\!-\!\beta)\!\left[F(x^{s}_{k-1})-F(x^{*})+F(\widetilde{x}^{s-1})-F(x^{*})\right]\!.
\end{split}
\end{equation*}
\end{corollary}

\onecolumn

\begin{proof}
\begin{equation*}
\begin{split}
&\alpha\mathbb{E}\!\left[\left\|\nabla\! f_{i^{s}_{k}}\!(x^{s}_{k-\!1})\!-\!\nabla\! f(x^{s}_{k-\!1})\!-\!\nabla\! f_{i^{s}_{k}}\!(\widetilde{x}^{s-\!1})\!+\!\nabla\! f(\widetilde{x}^{s-\!1})\right\|^{2}\right]-\beta\mathbb{E}\!\left[\left\|\nabla\! f_{i^{s}_{k}}\!(x^{s}_{k-\!1})\!-\!\nabla\! f_{i^{s}_{k}}\!(\widetilde{x}^{s-\!1})\right\|^{2}\right]\\
=\,&\alpha\mathbb{E}\!\!\left[\left\|[\nabla\! f_{i^{s}_{k}}\!(x^{s}_{k-\!1})\!-\!\nabla\! f_{i^{s}_{k}}\!(\widetilde{x}^{s-\!1})]\!-\![\nabla\! f(x^{s}_{k-1})\!-\!\nabla\! f(\widetilde{x}^{s-\!1})]\right\|^{2}\right]-\beta\mathbb{E}\!\!\left[\left\|\nabla\! f_{i^{s}_{k}}\!(x^{s}_{k-\!1})\!-\!\nabla\! f_{i^{s}_{k}}\!(\widetilde{x}^{s-\!1})\right\|^{2}\right]\\
=\,&\alpha\mathbb{E}\!\!\left[\left\|\nabla\! f_{i^{s}_{k}}\!(x^{s}_{k-\!1})\!-\!\nabla\! f_{i^{s}_{k}}\!(\widetilde{x}^{s-\!1})\right\|^{2}\right]\!-\!\alpha\!\left\|\nabla\! f(x^{s}_{k-\!1})\!-\!\nabla\! f(\widetilde{x}^{s-\!1})\right\|^{2}\!-\!\beta\mathbb{E}\!\!\left[\left\|\nabla\! f_{i^{s}_{k}}\!(x^{s}_{k-\!1})\!-\!\nabla\! f_{i^{s}_{k}}\!(\widetilde{x}^{s-\!1})\right\|^{2}\right]\\
\leq\,&\alpha\mathbb{E}\!\left[\left\|\nabla\! f_{i^{s}_{k}}\!(x^{s}_{k-1})-\nabla\! f_{i^{s}_{k}}\!(\widetilde{x}^{s-1})\right\|^{2}\right]\!-\!\beta\mathbb{E}\!\left[\left\|\nabla\! f_{i^{s}_{k}}\!(x^{s}_{k-\!1})\!-\!\nabla\! f_{i^{s}_{k}}\!(\widetilde{x}^{s-\!1})\right\|^{2}\right]\\
=\,&(\alpha\!-\!\beta)\mathbb{E}\!\left[\left\|\left[\nabla\! f_{i^{s}_{k}}\!(x^{s}_{k-1})-\nabla\! f_{i^{s}_{k}}\!(x^{*})\right]-[\nabla\! f_{i^{s}_{k}}\!(\widetilde{x}^{s-1})-\nabla\! f_{i^{s}_{k}}\!(x^{*})]\right\|^{2}\right]\\
\leq\,&2(\alpha\!-\!\beta)\left\{\mathbb{E}\!\left[\left\|\nabla\! f_{i^{s}_{k}}\!(x^{s}_{k-\!1})-\nabla\! f_{i^{s}_{k}}\!(x^{*})\right\|^{2}\right]+\mathbb{E}\!\left[\left\|\nabla\! f_{i^{s}_{k}}\!(\widetilde{x}^{s-\!1})-\nabla\! f_{i^{s}_{k}}\!(x^{*})\right\|^{2}\right]\right\}\\
\leq\, & 4L(\alpha\!-\!\beta)\!\left[F(x^{s}_{k-1})-F(x^{*})+F(\widetilde{x}^{s-1})-F(x^{*})\right]\!,
\end{split}
\end{equation*}
where the second equality holds due to the fact that $\mathbb{E}[\|x\!-\!\mathbb{E}x\|^{2}]\!=\!\mathbb{E}[\|x\|^{2}]\!-\!\|\mathbb{E}x\|^{2}$; the second inequality holds due to the fact that $\|a-b\|^{2}\leq2(\|a\|^{2}+\|b\|^{2})$; and the last inequality follows from Lemma 3.4 in~\cite{xiao:prox-svrg} (i.e., $\mathbb{E}[\left\|\nabla\! f_{i}(x)\!-\!\nabla\! f_{i}(x^{*})\right\|^{2}]\!\leq\! 2L\!\left[F(x)\!-\!F(x^{*})\right]$).
\end{proof}

Moreover, we also introduce the following lemmas~\cite{baldassarre:prox,lan:sgd}, which are useful in our convergence analysis.

\begin{lemma}
\label{prop1}
Let $\widetilde{F}(x,y)$ be the linear approximation of $F(\cdot)$ at $y$ with respect to $f$, i.e.,
\begin{displaymath}
\widetilde{F}(x,y)=f(y)+\left\langle \nabla f(y),\, x-y\right\rangle+ r(x).
\end{displaymath}
Then
\begin{displaymath}
F(x)\leq \widetilde{F}(x,y)+\frac{L}{2}\|x-y\|^{2}\leq F(x)+\frac{L}{2}\|x-y\|^{2}.
\end{displaymath}
\end{lemma}

\begin{lemma}
\label{prop2}
Assume that $\hat{x}$ is an optimal solution of the following problem,
\begin{displaymath}
\min_{x\in\mathbb{R}^{d}}\frac{\tau}{2}\|x-y\|^{2}+g(x),
\end{displaymath}
where $g(x)$ is a convex function (but possibly non-differentiable). Then the following inequality holds for all $x\!\in\!\mathbb{R}^{d}$:
\begin{equation*}
g(\hat{x})+\frac{\tau}{2}\|\hat{x}-y\|^{2}+\frac{\tau}{2}\|x-\hat{x}\|^{2}\leq g(x)+\frac{\tau}{2}\|x-y\|^{2}.
\end{equation*}
\end{lemma}
\vspace{1mm}

\textbf{Proof of Theorem 1:}
\begin{proof}
Let $\eta=\frac{1}{L\alpha}$ and $p_{i^{s}_{k}}\!=\!\widetilde{\nabla}\! f_{i^{s}_{k}}\!(x^{s}_{k-\!1})\!=\!\nabla\! f_{i^{s}_{k}}\!(x^{s}_{k-1})-\nabla\! f_{i^{s}_{k}}\!(\widetilde{x}^{s-1})+\nabla\! f(\widetilde{x}^{s-1})$. Using Lemma~\ref{prop1}, we have
\begin{equation}\label{equ71}
\begin{split}
F(y^{s}_{k})\leq\,& f(x^{s}_{k-1})+\left\langle\nabla\! f(x^{s}_{k-\!1}),\,y^{s}_{k}\!-\!x^{s}_{k-\!1}\right\rangle+\frac{L\alpha}{2}\!\left\|y^{s}_{k}\!-\!x^{s}_{k-\!1}\right\|^{2}\!-\!\frac{L(\alpha\!-\!1)}{2}\!\left\|y^{s}_{k}\!-\!x^{s}_{k-\!1}\right\|^{2}+r(y^{s}_{k})\\
=\,& f_{i^{s}_{k}}\!(x^{s}_{k-1})+\left\langle p_{i^{s}_{k}},\,y^{s}_{k}-x^{s}_{k-1}\right\rangle+r(y^{s}_{k})+\frac{L\alpha}{2}\!\|y^{s}_{k}-x^{s}_{k-1}\|^2\\
&+\left\langle\nabla\! f(x^{s}_{k-1})-p_{i^{s}_{k}},\,y^{s}_{k}-x^{s}_{k-1}\right\rangle-\frac{L(\alpha\!-\!1)}{2}\|y^{s}_{k}-x^{s}_{k-1}\|^{2}+f(x^{s}_{k-1})-f_{i^{s}_{k}}(x^{s}_{k-1}).
\end{split}
\end{equation}
Then
\begin{equation}\label{equ72}
\begin{split}
&\left\langle\nabla\! f(x^{s}_{k-1})-p_{i^{s}_{k}},\,y^{s}_{k}-x^{s}_{k-1}\right\rangle-\frac{L(\alpha\!-\!1)}{2}\|y^{s}_{k}-x^{s}_{k-1}\|^{2}\\
\leq\,& \frac{1}{2L(\alpha\!-\!1)}\|\nabla\!f(x^{s}_{k-1})-p_{i^{s}_{k}}\|^{2}+\frac{L(\alpha\!-\!1)}{2}\|y^{s}_{k}\!-\!x^{s}_{k-1}\|^{2}-\frac{L(\alpha\!-\!1)}{2}\|y^{s}_{k}\!-\!x^{s}_{k-1}\|^{2}\\
=\,&\frac{1}{2L(\alpha\!-\!1)}\|\nabla\!f(x^{s}_{k-1})-p_{i^{s}_{k}}\|^{2},
\end{split}
\end{equation}
where the inequality follows from the Young's inequality, i.e., $a^{T}b\leq\|a\|^{2}/(2\rho)+\rho\|b\|^{2}/2$ for any $\rho\!>\!0$. Substituting the inequality \eqref{equ72} into the inequality \eqref{equ71}, we have
\begin{equation}\label{equ73}
\begin{split}
F(y^{s}_{k})&\leq f_{i^{s}_{k}}\!(x^{s}_{k-1})+\left\langle p_{i^{s}_{k}},\,y^{s}_{k}-x^{s}_{k-1}\right\rangle+r(y^{s}_{k})+\frac{L\alpha}{2}\|y^{s}_{k}-x^{s}_{k-1}\|^2\\
&\quad+\frac{1}{2L(\alpha\!-\!1)}\|\nabla\!f(x^{s}_{k-1})-p_{i^{s}_{k}}\|^{2}+f(x^{s}_{k-1})-f_{i^{s}_{k}}(x^{s}_{k-1})\\
&\leq f_{i^{s}_{k}}\!(x^{s}_{k-\!1})+r(\widehat{w}^{s}_{k-\!1})+\frac{L\alpha}{2}\!\left(\|\widehat{w}^{s}_{k-\!1}\!-\!x^{s}_{k-1}\|^{2}\!-\!\|\widehat{w}^{s}_{k-\!1}\!-\!y^{s}_{k}\|^{2}\right)+\langle p_{i^{s}_{k}},\,\widehat{w}^{s}_{k-\!1}\!-\!x^{s}_{k-\!1}\rangle\\
&\quad+\frac{1}{2L(\alpha\!-\!1)}\|\nabla\!f(x^{s}_{k-1})-p_{i^{s}_{k}}\|^{2}+f(x^{s}_{k-\!1})-f_{i^{s}_{k}}(x^{s}_{k-\!1})\\
&\leq F_{i^{s}_{k}}\!(\widehat{w}^{s}_{k-1})+\frac{L\alpha}{2}\left(\|\widehat{w}^{s}_{k-1}-x^{s}_{k-1}\|^{2}-\|\widehat{w}^{s}_{k-1}-y^{s}_{k}\|^{2}\right)+f(x^{s}_{k-1})-f_{i^{s}_{k}}(x^{s}_{k-1})\\
&\quad+\frac{1}{2L(\alpha\!-\!1)}\|\nabla\!f(x^{s}_{k-1})-p_{i^{s}_{k}}\|^{2}+\left\langle-\nabla f_{i^{s}_{k}}(\widetilde{x}^{s-1})+\nabla f(\widetilde{x}^{s-1}),\,\widehat{w}^{s}_{k-1}-x^{s}_{k-1}\right\rangle\\
&\leq \sigma F_{i^{s}_{k}}\!(x^{*})+(1-\sigma)F_{i^{s}_{k}}\!(\widehat{x}^{s}_{k-1})+\frac{L\alpha\sigma^{2}}{2}\left(\|x^{*}-z^{s}_{k-1}\|^{2}-\|x^{*}-z^{s}_{k}\|^{2}\right)\\
&\quad+\frac{1}{2L(\alpha\!-\!1)}\|\nabla\!f(x^{s}_{k-1})-p_{i^{s}_{k}}\|^{2}+f(x^{s}_{k-1})-f_{i^{s}_{k}}(x^{s}_{k-1})\\
&\quad+\!\left\langle\nabla f(\widetilde{x}^{s-1})\!-\!\nabla f_{i^{s}_{k}}(\widetilde{x}^{s-1}),\,\widehat{w}^{s}_{k-1}\!-\!x^{s}_{k-1}\right\rangle,
\end{split}
\end{equation}
where $\widehat{w}^{s}_{k-1}\!=\!\sigma x^{*}+(1\!-\!\sigma)\widehat{x}^{s}_{k-1}$, and $\widehat{x}^{s}_{k-\!1}\!=\!\theta_{k-\!1}x^{s}_{k-2}$. The second inequality follows from Lemma \ref{prop2} with $g(x)\!:=\!\left\langle p_{i^{s}_{k}},\,x\!-\!x^{s}_{k-1}\right\rangle\!+\!r(x)$, $\tau\!=\!L\alpha$, $\hat{x}\!=\!y^{s}_{k}$, $x\!=\!\widehat{w}^{s}_{k-1}$ and $y\!=\!x^{s}_{k-1}$; the third inequality holds due to the convexity of the component function $f_{i^{s}_{k}}(x)$ (i.e., $f_{i^{s}_{k}}\!(x^{s}_{k-\!1})\!+\!\langle\nabla\! f_{i^{s}_{k}}\!(x^{s}_{k-\!1}),\widehat{w}^{s}_{k-\!1}\!-\!x^{s}_{k-\!1}\rangle\!\leq\! f_{i^{s}_{k}}\!(\widehat{w}^{s}_{k-\!1})$); and the last inequality holds due to the convexity of the function $F_{i^{s}_{k}}\!(x)\!:=\!f_{i^{s}_{k}}\!(x)\!+\!r(x)$, and
\begin{displaymath}
z^{s}_{k-1}=[x^{s}_{k-1}-(1\!-\!\sigma)\widehat{x}^{s}_{k-1}]/{\sigma},\;\,z^{s}_{k}=[y^{s}_{k}-(1\!-\!\sigma)\widehat{x}^{s}_{k-1}]/{\sigma},
\end{displaymath}
which mean that $\widehat{w}^{s}_{k-\!1}\!-x^{s}_{k-\!1}=\sigma(x^{*}-z^{s}_{k-\!1})$ and $\widehat{w}^{s}_{k-\!1}\!-y^{s}_{k}=\sigma(x^{*}-z^{s}_{k})$.

Using Property 1 with $\zeta=\frac{\delta\eta}{1-L\eta}$ and $\eta=1/L\alpha,$\footnote{Note that our fast versions of SVRG-SD (i.e., SVRG-SD with randomly partial sufficient decrease) have the similar convergence properties as SVRG-SD because Property 1 still holds in the case when $\theta_{k}\!=\!1$. That is, the main difference between their convergence properties is the different values of $\beta_{k}$, as shown below.} we obtain
\begin{equation}\label{equ74}
\begin{split}
F(\theta_{k}{x}^{s}_{k-1})=F(\widehat{x}^{s}_{k})&\leq F(x^{s}_{k-1})-\frac{(\theta_{k}\!-\!1)^{2}}{2L(\alpha-1)}\|\nabla\!f_{i^{s}_{k}}\!(x^{s}_{k-\!1})-\!\nabla\! f_{i^{s}_{k}}\!(\widetilde{x}^{s-1})\|^{2}\\
&\leq F(x^{s}_{k-1})-\frac{\beta_{k}}{2L(\alpha\!-\!1)}\|\nabla\!f_{i^{s}_{k}}\!(x^{s}_{k-\!1})-\!\nabla\! f_{i^{s}_{k}}\!(\widetilde{x}^{s-1})\|^{2},
\end{split}
\end{equation}
where $\beta_{k}=\min\!\left[1/\alpha_{k},\,(\theta_{k}\!-\!1)^{2}\right]$, and $\alpha_{k}$ is defined below. Then there exists $\overline{\beta}_{k}$ such that
\begin{equation}\label{equ83}
\mathbb{E}\!\left[\frac{\beta_{k}}{2L(\alpha\!-\!1)}\|\nabla\!f_{i^{s}_{k}}\!(x^{s}_{k-\!1})-\!\nabla\! f_{i^{s}_{k}}\!(\widetilde{x}^{s-1})\|^{2}\right]=\frac{\overline{\beta}_{k}}{2L(\alpha\!-\!1)}\mathbb{E}\!\left[\|\nabla\!f_{i^{s}_{k}}\!(x^{s}_{k-\!1})-\!\nabla\! f_{i^{s}_{k}}\!(\widetilde{x}^{s-1})\|^{2}\right],
\end{equation}
where $\overline{\beta}_{k}=\mathbb{E}[\beta_{k}\|\nabla\!f_{i^{s}_{k}}\!(x^{s}_{k-\!1})-\!\nabla\! f_{i^{s}_{k}}\!(\widetilde{x}^{s-1})\|^{2}]/\mathbb{E}[\|\nabla\!f_{i^{s}_{k}}\!(x^{s}_{k-\!1})-\!\nabla\! f_{i^{s}_{k}}\!(\widetilde{x}^{s-1})\|^{2}]$, and $\overline{\beta}_{k}<(\alpha\!-\!1)/{2}$. Using the inequality~\eqref{equ74}, then we have
\begin{equation}\label{equ101}
\begin{split}
\mathbb{E}\!\left[F(\widehat{x}^{s}_{k})-F(x^{*})\right]&\leq  \mathbb{E}\!\left[F(x^{s}_{k-1})-F(x^{*})-\frac{\beta_{k}}{2L(\alpha\!-\!1)}\|\nabla\!f_{i^{s}_{k}}\!(x^{s}_{k-\!1})-\!\nabla\! f_{i^{s}_{k}}\!(\widetilde{x}^{s-1})\|^{2}\right]\\
&= \mathbb{E}\!\left[F(x^{s}_{k-1})-F(x^{*})\right]-\frac{\overline{\beta}_{k}}{2L(\alpha\!-\!1)}\mathbb{E}\!\left[\|\nabla\!f_{i^{s}_{k}}\!(x^{s}_{k-\!1})-\!\nabla\! f_{i^{s}_{k}}\!(\widetilde{x}^{s-1})\|^{2}\right].
\end{split}
\end{equation}

There must exist a constant $\alpha_{k}\!>\!0$ such that $\mathbb{E}\!\left[F(y^{s}_{k})\!-\!F(x^{*})\right]\!=\!\alpha_{k}\mathbb{E}\!\left[F(x^{s}_{k-\!1})\!-\!F(x^{*})\right]$. Since
$\mathbb{E}\!\left[f(x^{s}_{k-\!1})\!-\!f_{i^{s}_{k}}\!(x^{s}_{k-\!1})\right]\!=\!0$, $\mathbb{E}\!\left[\nabla\! f_{i^{s}_{k}}(\widetilde{x}^{s-1})\right]\!=\!\nabla\! f(\widetilde{x}^{s-1})$, $\mathbb{E}\!\left[F_{i^{s}_{k}}\!(x^{*})\right]\!=\!F(x^{*})$, and $\mathbb{E}\!\left[F_{i^{s}_{k}}\!(x^{s}_{k-1})\right]\!=\!F(x^{s}_{k-1})$, and taking the expectation of both sides of (\ref{equ73}), we have
\begin{equation}\label{equ75}
\begin{split}
&\alpha_{k}\mathbb{E}\!\left[F(x^{s}_{k-1})-F(x^{*})\right]-\frac{c_{k}\overline{\beta}_{k}}{2L(\alpha\!-\!1)}\mathbb{E}\!\left[\|\nabla\!f_{i^{s}_{k}}\!(x^{s}_{k-\!1})-\!\nabla\! f_{i^{s}_{k}}\!(\widetilde{x}^{s-1})\|^{2}\right]\\
\leq\,&(1-\sigma)\mathbb{E}\!\left[F(\widehat{x}^{s}_{k-1})-F(x^{*})\right]+\frac{L\alpha\sigma^{2}}{2}\mathbb{E}\!\left[\|x^{*}-z^{s}_{k-1}\|^{2}-\|x^{*}-z^{s}_{k}\|^{2}\right]\\
&+\frac{1}{2L(\alpha\!-\!1)}\mathbb{E}\|\nabla\!f(x^{s}_{k-1})-p_{i^{s}_{k}}\|^{2}-\frac{c_{k}\overline{\beta}_{k}}{2L(\alpha\!-\!1)}\mathbb{E}\!\left[\|\nabla\!f_{i^{s}_{k}}\!(x^{s}_{k-\!1})-\!\nabla\! f_{i^{s}_{k}}\!(\widetilde{x}^{s-1})\|^{2}\right]\\
\leq&(1-\sigma)\mathbb{E}\!\left[F(\widehat{x}^{s}_{k-1})-F(x^{*})\right]+\frac{L\alpha\sigma^{2}}{2}\mathbb{E}\!\left[\|x^{*}-z^{s}_{k-1}\|^{2}-\|x^{*}-z^{s}_{k}\|^{2}\right]\\
&+\frac{2(1-c_{k}\overline{\beta}_{k})}{\alpha\!-\!1}\left[F(x^{s}_{k-1})-F(x^{*})+F(\widetilde{x}^{s-1})-F(x^{*})\right],
\end{split}
\end{equation}
where the second inequality follows from Lemma~\ref{lemm1} and Corollary~\ref{coro4}. Here, $c_{k}=\alpha_{k}-[{2(1\!-\!c_{k}\overline{\beta}_{k})}]/({\alpha\!-\!1})$, i.e.,
\begin{displaymath}
c_{k}= \frac{\alpha_{k}(\alpha-1)-2}{\alpha-1-2\overline{\beta}_{k}}.
\end{displaymath}
Since $\frac{2}{\alpha-1}<\sigma$ with the suitable choices of $\alpha$ and $\sigma$, we have $c_{k}>\alpha_{k}-\frac{2}{\alpha-1}>1-\sigma$. Thus, (\ref{equ75}) is rewritten as follows:
\begin{equation}\label{equ76}
\begin{split}
&c_{k}\mathbb{E}\!\left[F(x^{s}_{k-1})-F(x^{*})\right]-\frac{c_{k}\overline{\beta}_{k}}{2L(\alpha-1)}\mathbb{E}\!\left[\|p_{i^{s}_{k}}-\!\nabla\! f_{i^{s}_{k}}\!(\widetilde{x}^{s-1})\|^{2}\right]\\
\leq&\,(1-\sigma)\mathbb{E}[F(\widehat{x}^{s}_{k-1})-F(x^{*})]+\frac{L\alpha\sigma^{2}}{2}\mathbb{E}\!\left[\|x^{*}-z^{s}_{k-1}\|^{2}-\|x^{*}-z^{s}_{k}\|^{2}\right]\\
&\,+\frac{2(1-c_{k}\overline{\beta}_{k})}{\alpha-1}\mathbb{E}\!\left[F(\widetilde{x}^{s-1})-F(x^{*})\right].
\end{split}
\end{equation}

Combining the above two inequalities (\ref{equ101}) and (\ref{equ76}), we have
\begin{equation}
\begin{split}
&c_{k}\mathbb{E}\!\left[F(\widehat{x}^{s}_{k})-F(x^{*})\right]\\
\leq\,&(1-\sigma)\mathbb{E}\!\left[F(\widehat{x}^{s}_{k-1})-F(x^{*})\right]+\frac{L\alpha\sigma^{2}}{2}\mathbb{E}\!\left[\|x^{*}-z^{s}_{k-1}\|^{2}-\|x^{*}-z^{s}_{k}\|^{2}\right]\\
&+\frac{2(1-c_{k}\overline{\beta}_{k})}{\alpha-1}\mathbb{E}\!\left[F(\widetilde{x}^{s-1})-F(x^{*})\right].
\end{split}
\end{equation}

Taking the expectation over the random choice of $i^{s}_{1},i^{s}_{2},\ldots,i^{s}_{m}$, summing up the above inequality over $k=1,\ldots,m$, and $\widehat{x}^{s}_{0}=\widetilde{x}^{s-1}$, we have
\begin{equation}\label{equ102}
\begin{split}
&\mathbb{E}\!\left[\sum^{m}_{k=1}\!\left[c_{k}-(1-\sigma)\right][F(\widehat{x}^{s}_{k})-F(x^{*})]\right]\\
\leq\,&(1-\sigma)\mathbb{E}\!\left[F(\widetilde{x}^{s-1})-F(x^{*})\right]+\frac{L\alpha\sigma^{2}}{2}\mathbb{E}\!\left[\|x^{*}-z^{s}_{0}\|^{2}-\|x^{*}-z^{s}_{m}\|^{2}\right]\\
&+\mathbb{E}\!\left[\sum^{m}_{k=1}\frac{2(1-c_{k}\overline{\beta}_{k})}{\alpha-1}[F(\widetilde{x}^{s-1})-F(x^{*})]\right].
\end{split}
\end{equation}
In addition, there exists $\widehat{\beta}^{s}$ for the $s$-th epoch such that
\begin{equation}\label{equ103}
\begin{split}
&\;\mathbb{E}\!\left[\sum^{m}_{k=1}\left[c_{k}-(1-\sigma)\right][F(\widehat{x}^{s}_{k})-F(x^{*})]\right]\\
=&\;\mathbb{E}\!\left[\sum^{m}_{k=1}\left(\sigma-\frac{2}{\alpha-1}+\frac{2c_{k}\overline{\beta}_{k}}{\alpha-1}\right)[F(\widehat{x}^{s}_{k})-F(x^{*})]\right]\\
=&\;\left(\sigma-\frac{2}{\alpha-1}+\widehat{\beta}^{s}\right)\mathbb{E}\!\left[\sum^{m}_{k=1}[F(\widehat{x}^{s}_{k})-F(x^{*})]\right],
\end{split}
\end{equation}
where
\begin{displaymath}
\widehat{\beta}^{s}=\frac{\mathbb{E}\!\left[\sum^{m}_{k=1}\frac{2c_{k}\beta_{k}}{\alpha-1}[F(\widehat{x}^{s}_{k})-F(x^{*})]\right]}{\mathbb{E}\!\left[\sum^{m}_{k=1}[F(\widehat{x}^{s}_{k})-F(x^{*})]\right]}.
\end{displaymath}
Let $\widehat{\beta}=\min_{s=1,\ldots,S}\widehat{\beta}^{s}$. Using
\begin{displaymath}
\widetilde{x}^{s}=\frac{1}{m}\sum^{m}_{k=1}\widehat{x}^{s}_{k},\;\,F(\widetilde{x}^{s})\leq\frac{1}{m}\sum^{m}_{k=1}F(\widehat{x}^{s}_{k}),
\end{displaymath}
(\ref{equ102}) and (\ref{equ103}), we have
\begin{equation*}
\begin{split}
&\left(\sigma-\frac{2}{\alpha-1}+\widehat{\beta}\right)m\mathbb{E}\!\left[F(\widetilde{x}^{s})-F(x^{*})\right]\\
\leq\,&\left(1-\sigma+\frac{2m}{\alpha\!-\!1}\right)\mathbb{E}\!\left[F(\widetilde{x}^{s-1})-F(x^{*})\right]\\
&+\frac{L\alpha\sigma^{2}}{2}\mathbb{E}\!\left[\|x^{*}-z^{s}_{0}\|^{2}-\|x^{*}-z^{s}_{m}\|^{2}\right].
\end{split}
\end{equation*}
Therefore,
\begin{equation*}
\begin{split}
&\mathbb{E}\!\left[F(\widetilde{x}^{s})-F(x^{*})\right]\\
\leq\,&\left(\frac{1-\sigma}{\left(\sigma-\frac{2}{\alpha-1}+\widehat{\beta}\right)m}+\frac{2}{(\alpha\!-\!1)\left(\sigma-\frac{2}{\alpha-1}+\widehat{\beta}\right)}\right)\mathbb{E}\!\left[F(\widetilde{x}^{s-1})-F(x^{*})\right]\\
&+\frac{L\alpha\sigma^{2}}{2m\left(\sigma-\frac{2}{\alpha-1}+\widehat{\beta}\right)}\mathbb{E}\!\left[\|x^{*}-z^{s}_{0}\|^{2}-\|x^{*}-z^{s}_{m}\|^{2}\right].
\end{split}
\end{equation*}
This completes the proof.
\end{proof}
\vspace{3mm}

\section*{Appendix B: Proofs of Corollary 1}
\begin{proof}
For $\mu$-strongly convex problems, and let $x^{s}_{0}=\widehat{x}^{s}_{0}=\widetilde{x}^{s-1}$ and
\begin{equation*}
z^{s}_{0}=\frac{x^{s}_{0}-(1-\sigma)\widehat{x}^{s}_{0}}{\sigma}=\widetilde{x}^{s-1}.
\end{equation*}
Due to the strong convexity of $F(\cdot)$, we have
\begin{equation}
\frac{\mu}{2}\|x^{*}-z^{s}_{0}\|^{2}=\frac{\mu}{2}\|x^{*}-\widetilde{x}^{s-1}\|^{2}\leq F(\widetilde{x}^{s-1})-F(x^{*}).
\end{equation}
Using Theorem 1, we obtain
\begin{equation*}
\begin{split}
&\mathbb{E}\!\left[F(\widetilde{x}^{s})-F(x^{*})\right]\\
\leq&\left(\frac{1-\sigma}{m(\sigma\!-\!\frac{2}{\alpha-1}\!+\!\widehat{\beta})}+\frac{2}{(\alpha\!-\!1)\left(\sigma\!-\!\frac{2}{\alpha-1}\!+\!\widehat{\beta}\right)}+\frac{L\alpha\sigma^{2}}{m\mu\left(\sigma\!-\!\frac{2}{\alpha-1}\!+\!\widehat{\beta}\right)}\right)\mathbb{E}\!\left[F(\widetilde{x}^{s-1})-F(x^{*})\right].
\end{split}
\end{equation*}

Replacing $\alpha$ and $\sigma$ in the above inequality with $19$ and $1/2$, respectively, we have
\begin{equation*}
\begin{split}
&\mathbb{E}\!\left[F(\widetilde{x}^{s})-F(x^{*})\right]\\
\leq&\left(\frac{9}{(7+18\widehat{\beta})m}+\frac{2}{7+18\widehat{\beta}}+\frac{171L}{(14+36\widehat{\beta})m\mu}\right)\mathbb{E}\!\left[F(\widetilde{x}^{s-1})-F(x^{*})\right]\!.
\end{split}
\end{equation*}
This completes the proof.
\end{proof}
\vspace{3mm}

\section*{Appendix C: Proofs of Theorem 2}
\begin{proof}
For non-strongly convex problems, and using Theorem 1 with $\alpha=19$ and $\sigma=1/2$, we have
\begin{equation}\label{equ105}
\begin{split}
\mathbb{E}[F(\widetilde{x}^{s})-F(x^{*})]\leq&\;\frac{171L}{(28+72\widehat{\beta})m}\mathbb{E}\!\left[\left\|x^{*}-z^{s}_{0}\right\|^{2}-\left\|x^{*}-z^{s}_{m}\right\|^{2}\right]\\
&\;+\left(\frac{9}{(7+18\widehat{\beta})m}+\frac{2}{7+18\widehat{\beta}}\right)\left[F(\widetilde{x}^{s-1})-F(x^{*})\right]\!.
\end{split}
\end{equation}

According to the settings of Algorithm 1 for the non-strongly convex case, and let
\begin{equation*}
x^{s}_{0}=\widehat{x}^{s}_{0}=[x^{s-1}_{m}-(1-\sigma)\widehat{x}^{s-1}_{m}]/\sigma,
\end{equation*}
then we have
\begin{equation*}
z^{s}_{0}=\frac{x^{s}_{0}-(1-\sigma)\widehat{x}^{s}_{0}}{\sigma}=\frac{x^{s-1}_{m}-(1-\sigma)\widehat{x}^{s-1}_{m}}{\sigma},
\end{equation*}
and
\begin{equation*}
z^{s-1}_{m}=\frac{x^{s-1}_{m}-(1-\sigma)\widehat{x}^{s-1}_{m}}{\sigma}.
\end{equation*}
Therefore, $z^{s}_{0}=z^{s-1}_{m}$.

Using $z^{0}_{0}=\widetilde{x}^{0}$, and summing up the inequality (\ref{equ105}) over all $s=1,\ldots,S$, then
\begin{equation*}
\begin{split}
\mathbb{E}\!\left[F\!\left(\frac{1}{S}\sum^{S}_{s=1}\widetilde{x}^{s}\right)-F(x^{*})\right]\leq&\;\frac{171L}{(16+40\widehat{\beta})mS}\left\|x^{*}-\widetilde{x}^{0}\right\|^{2}\\
&\;+\left(\frac{9}{(4+8\widehat{\beta})mS}+\frac{1}{(2+4\widehat{\beta})S}\right)\left[F(\widetilde{x}^{0})-F(x^{*})\right]\!.
\end{split}
\end{equation*}

Due to the settings of Algorithm 1 for the non-strongly convex case, we have
\begin{equation*}\label{equ106}
\begin{split}
\mathbb{E}\!\left[F(\overline{x})-F(x^{*})\right]\leq&\;\frac{171L}{(16+40\widehat{\beta})mS}\left\|x^{*}-\widetilde{x}^{0}\right\|^{2}\\
&\;+\left(\frac{9}{(4+8\widehat{\beta})mS}+\frac{1}{(2+4\widehat{\beta})S}\right)\left[F(\widetilde{x}^{0})-F(x^{*})\right]\!.
\end{split}
\end{equation*}

This completes the proof.
 \end{proof}
\vspace{3mm}

\begin{figure}[t]
\centering
\subfigure[Objective gap vs.\ number of passes]{\includegraphics[width=0.432\columnwidth]{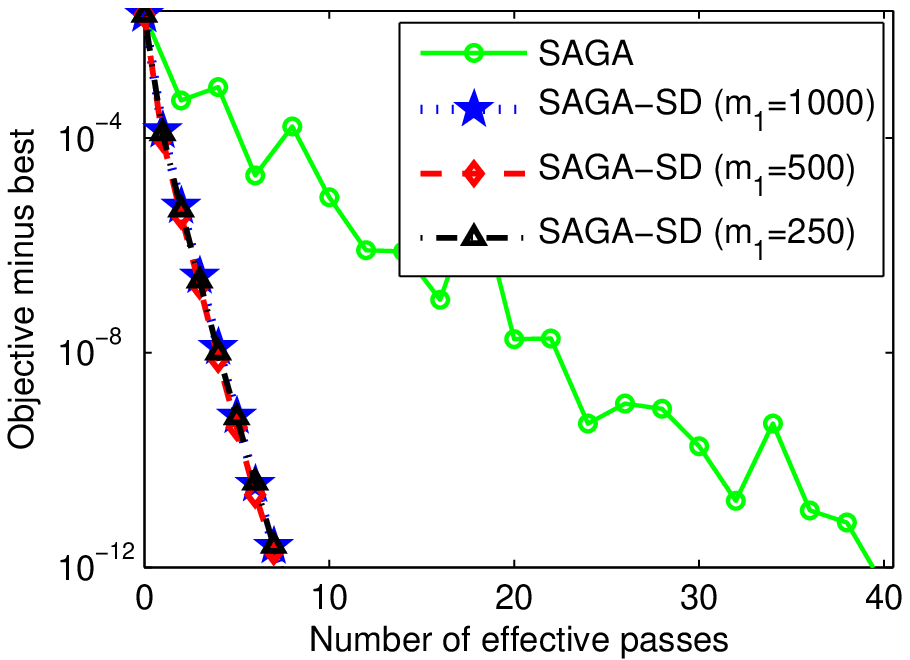}}\,\,\,\;\;
\subfigure[Objective gap vs.\ running time]{\includegraphics[width=0.432\columnwidth]{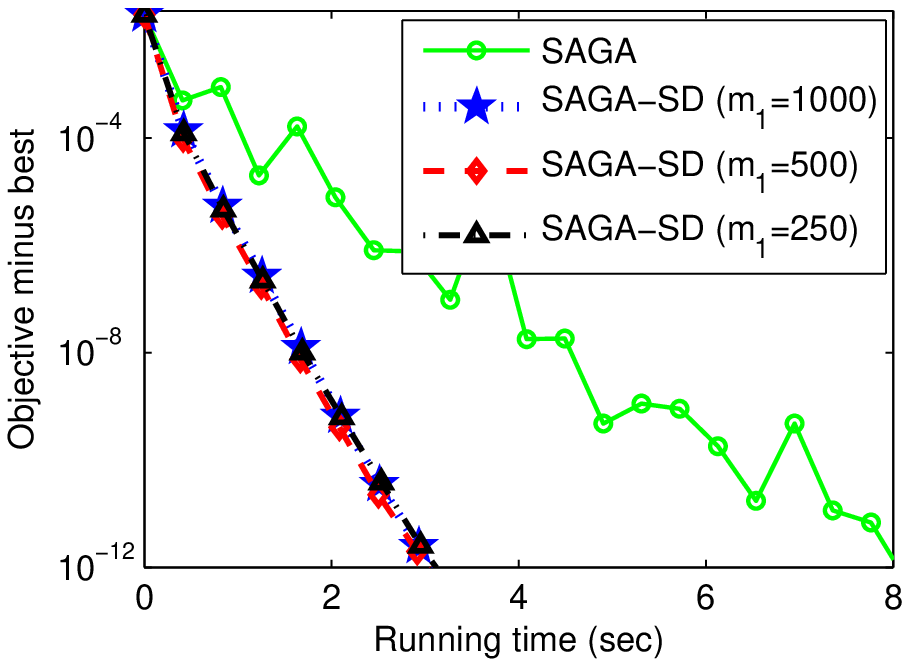}}
\caption{Comparison of SAGA and SAGA-SD with different values of $m_{1}$ for solving ridge regression problems on the Covtype dataset.}
\label{fig11}
\end{figure}

\begin{figure}[t]
\centering
\includegraphics[width=0.326\columnwidth]{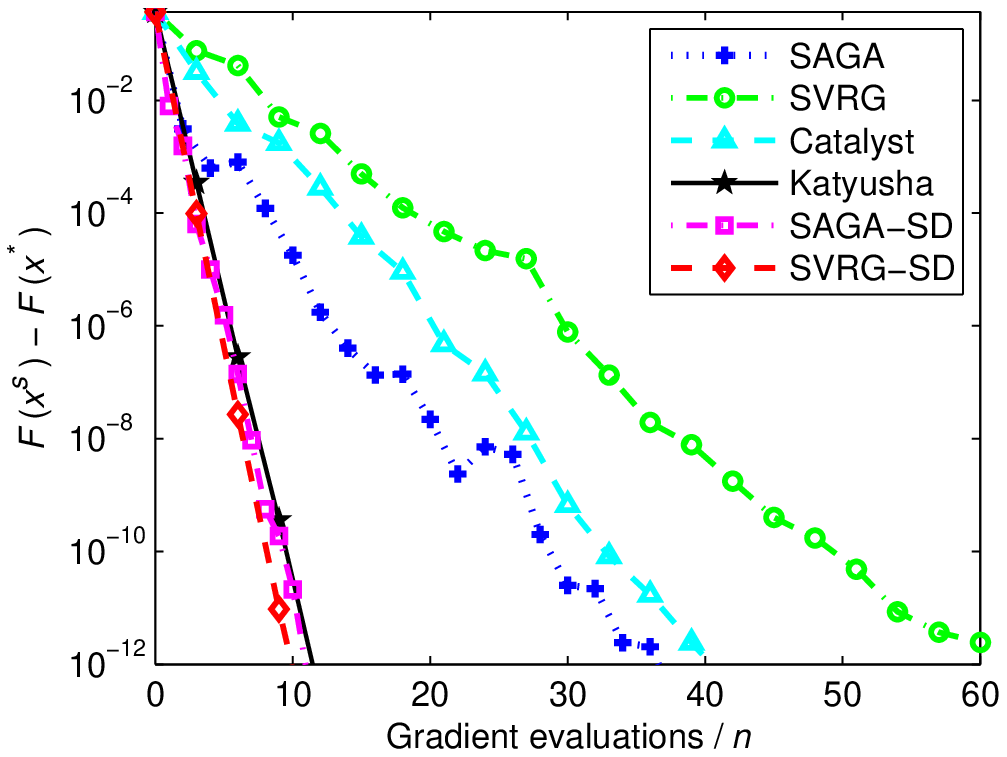}\,
\includegraphics[width=0.326\columnwidth]{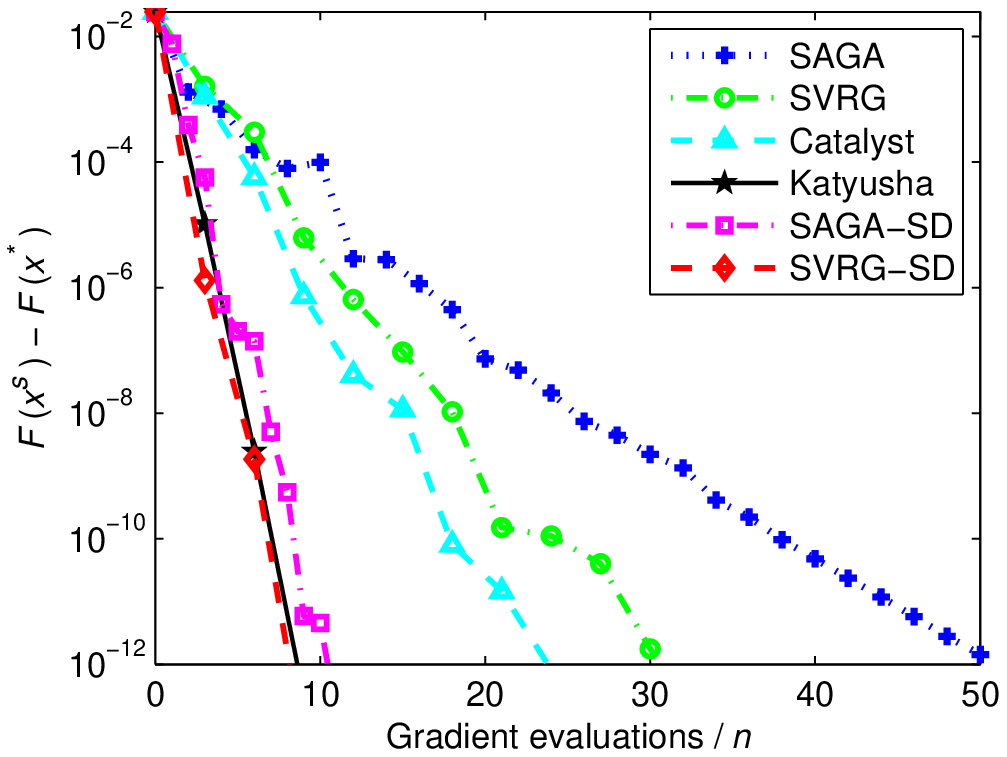}\,
\includegraphics[width=0.326\columnwidth]{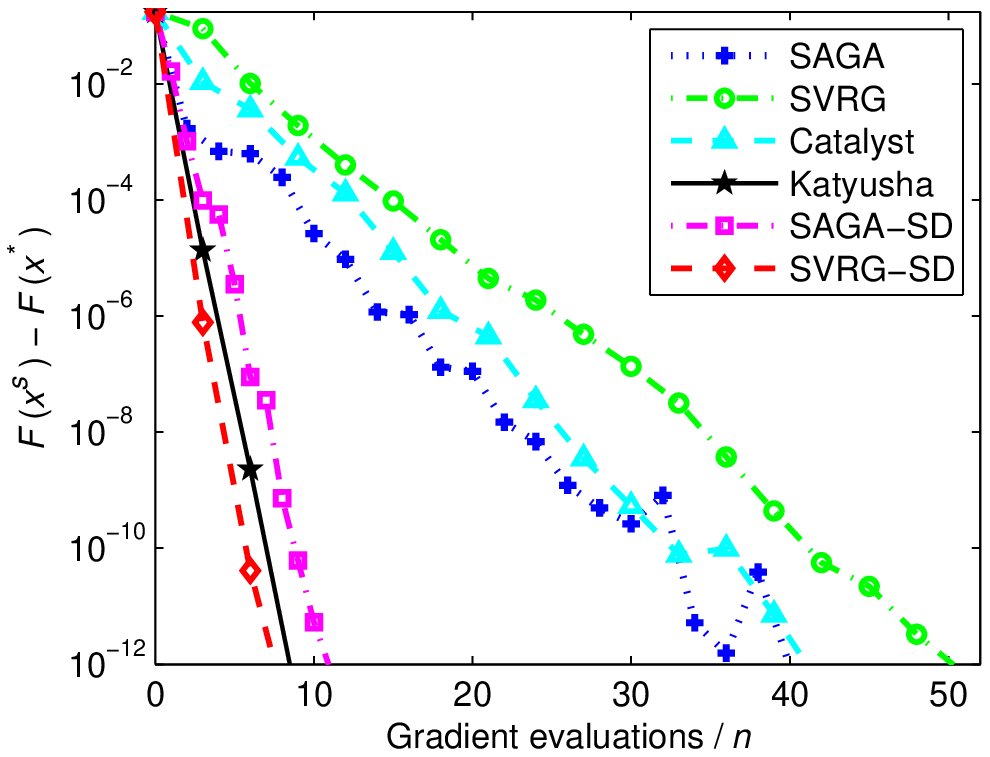}

\subfigure[Ijcnn1, $\lambda_{1}\!=\!10^{-5}$]{\includegraphics[width=0.326\columnwidth]{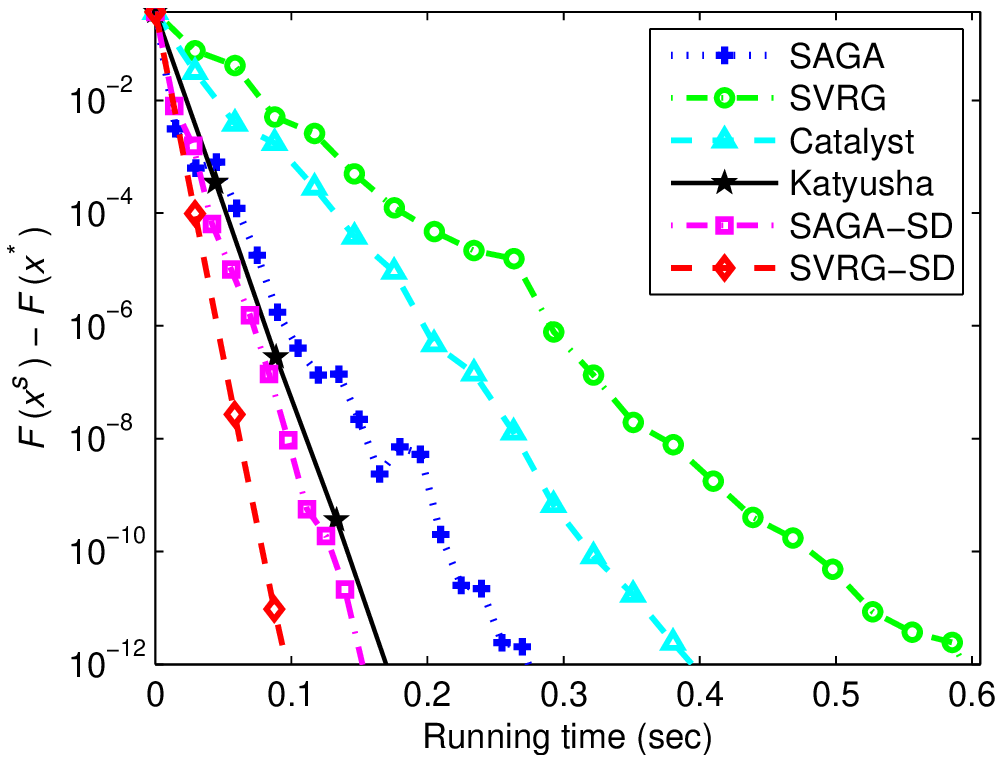}}\,
\subfigure[Covtype, $\lambda_{1}\!=\!10^{-5}$]{\includegraphics[width=0.326\columnwidth]{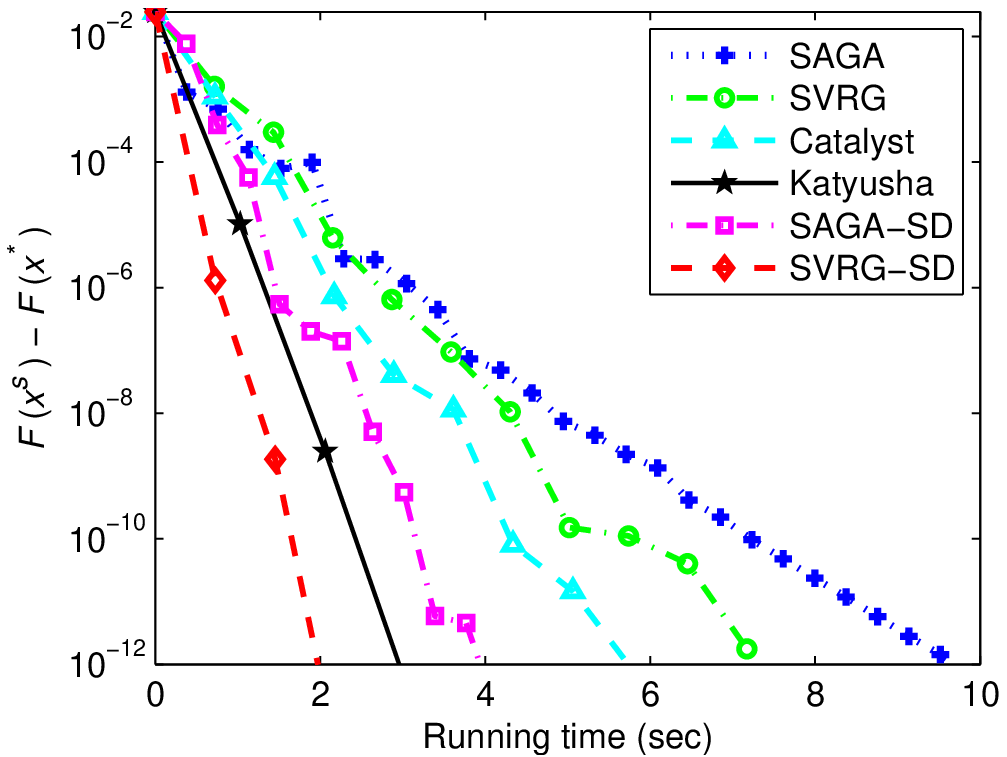}}\,
\subfigure[SUSY, $\lambda_{1}\!=\!10^{-5}$]{\includegraphics[width=0.326\columnwidth]{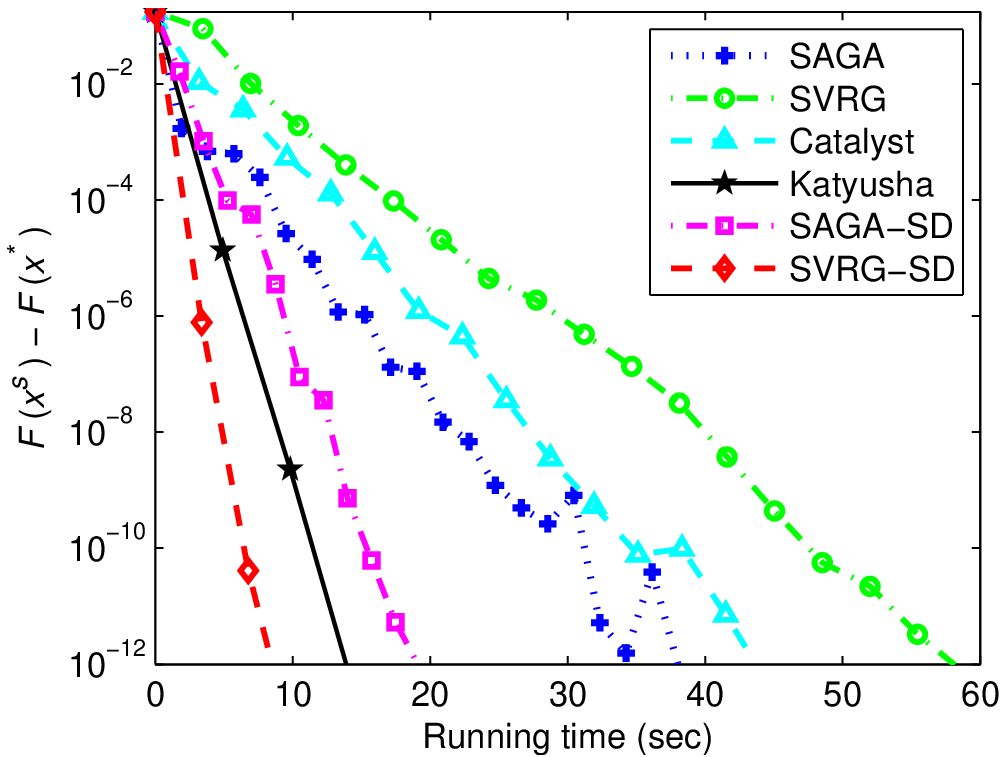}}
\vspace{-3.6mm}

\caption{Comparison of all the stochastic variance reduced gradient methods for solving strongly convex ridge regression problems on the three dense data sets: Ijcnn1, Covtype and SUSY. The vertical axis is the objective value minus the minimum, and the horizontal axis denotes the number of effective passes over the data (top) or the running time (bottom).}
\label{fig_sim1}
\end{figure}

\section*{Appendix D: Experiment Details}
The C++ code of SVRG~\cite{johnson:svrg} was downloaded from~\url{http://riejohnson.com/svrg_download.html}. The code of of SAGA~\cite{defazio:saga} was downloaded from~\url{http://www.aarondefazio.com/software.html}. For fair comparison, we implemented the proposed SVRG-SD and SAGA-SD algorithms, SAGA~\cite{defazio:saga}, Prox-SVRG~\cite{xiao:prox-svrg}, Catalyst~\cite{lin:vrsg} (which is based on SVRG and has the following three important parameters: $\alpha_{k}$, $\kappa$, and the step size, $\eta$), and Katyusha~\cite{zhu:Katyusha} in C++ with a Matlab interface\footnote{The codes of some algorithms can be downloaded by the following anonymous link:\\ \centerline{\url{https://www.dropbox.com/s/pyjeegseht77toh/Code_SVRG_SD.zip?dl=0.}}}, and performed all the experiments on a PC with an Intel i5-2400 CPU and 16GB RAM.

\begin{figure}[t]
\centering
\includegraphics[width=0.326\columnwidth]{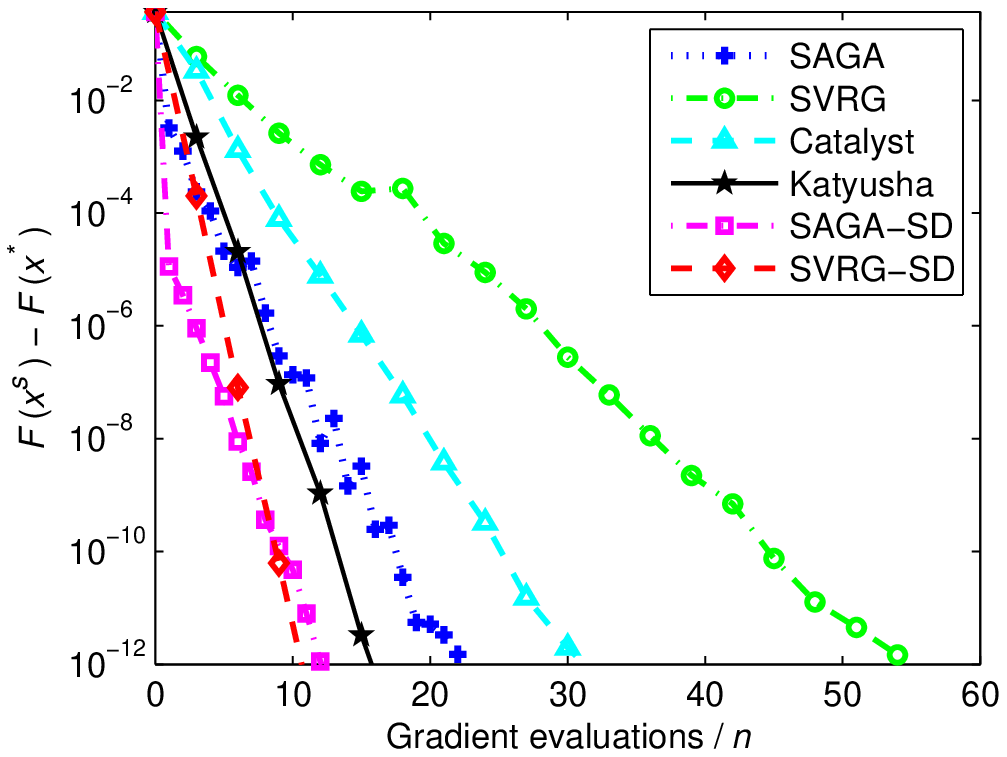}\,
\includegraphics[width=0.326\columnwidth]{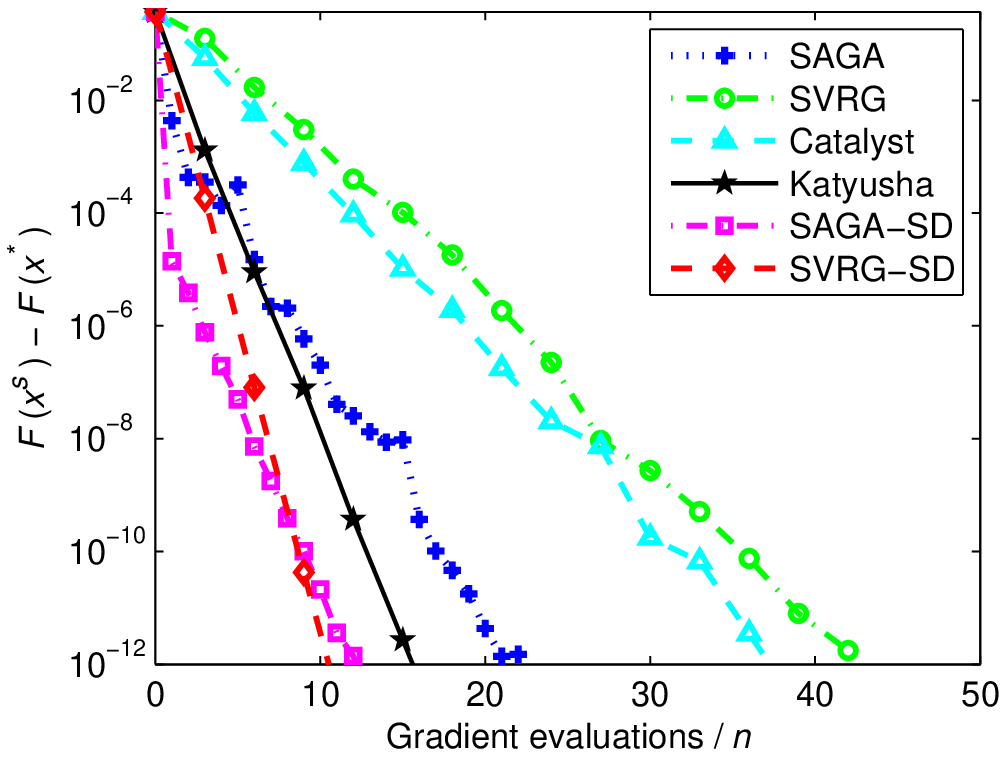}\,
\includegraphics[width=0.326\columnwidth]{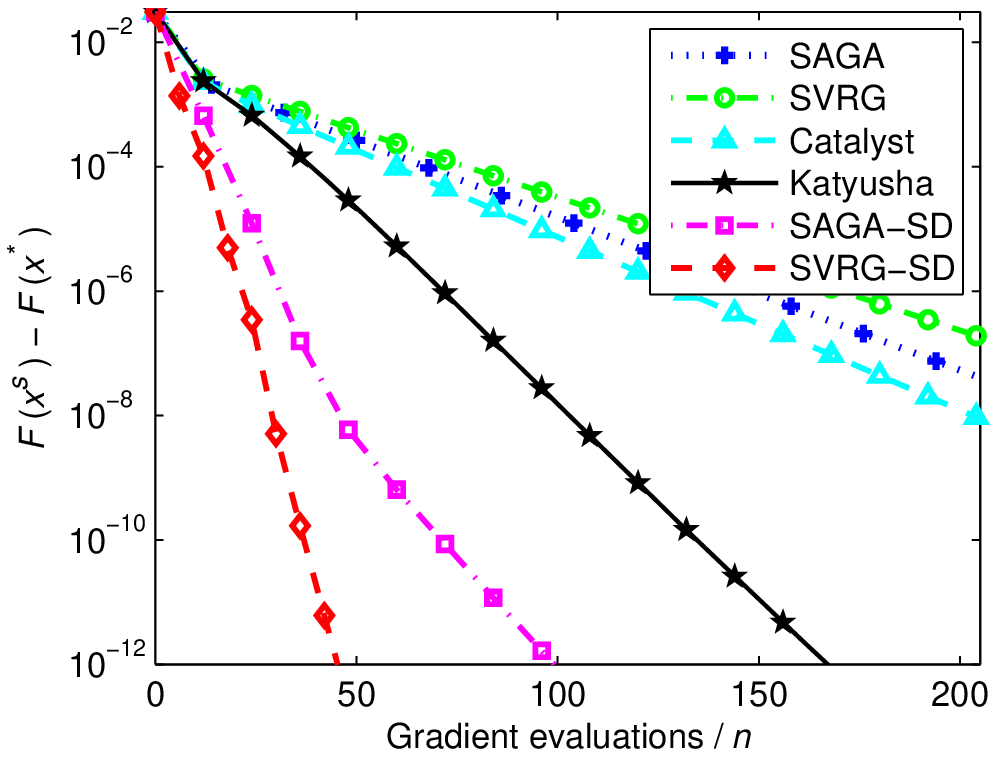}

\subfigure[Ijcnn1, $\lambda_{1}\!=\!10^{-6}$]{\includegraphics[width=0.326\columnwidth]{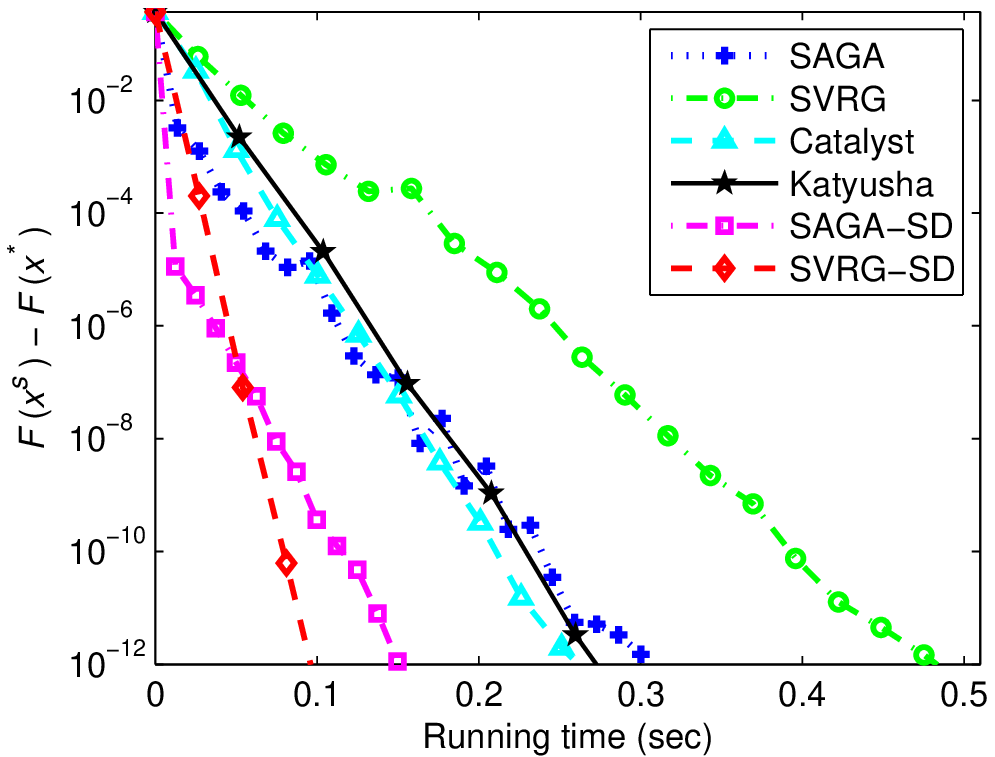}}\,
\subfigure[Ijcnn1, $\lambda_{1}\!=\!10^{-7}$]{\includegraphics[width=0.326\columnwidth]{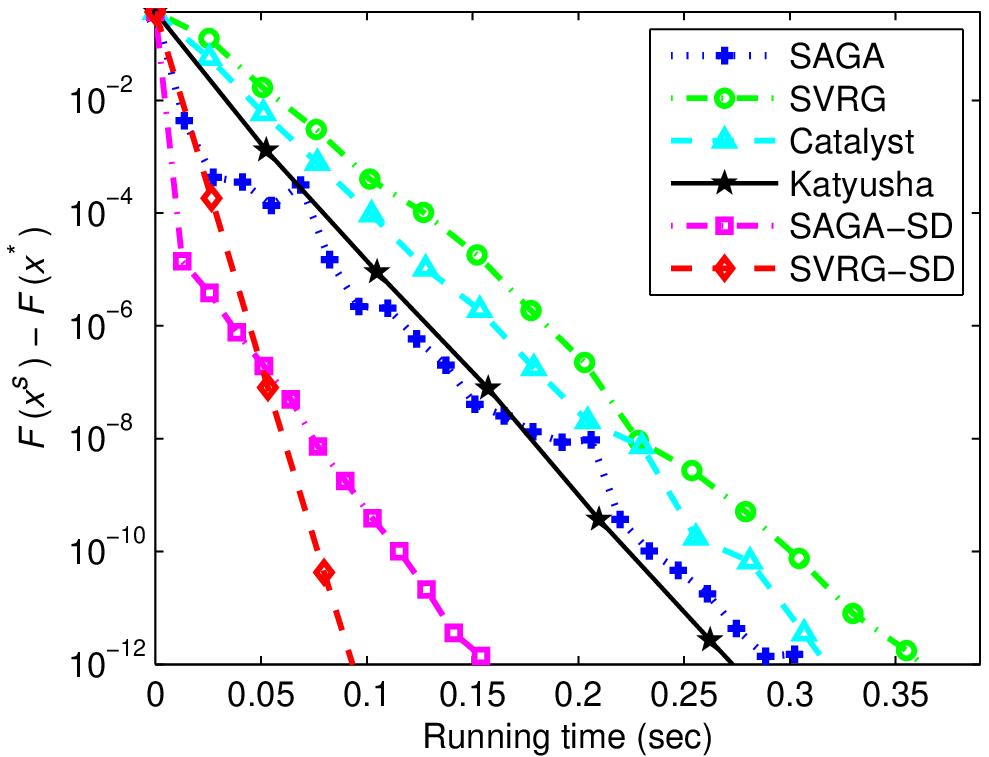}}\,
\subfigure[Covtype, $\lambda_{1}\!=\!10^{-7}$]{\includegraphics[width=0.326\columnwidth]{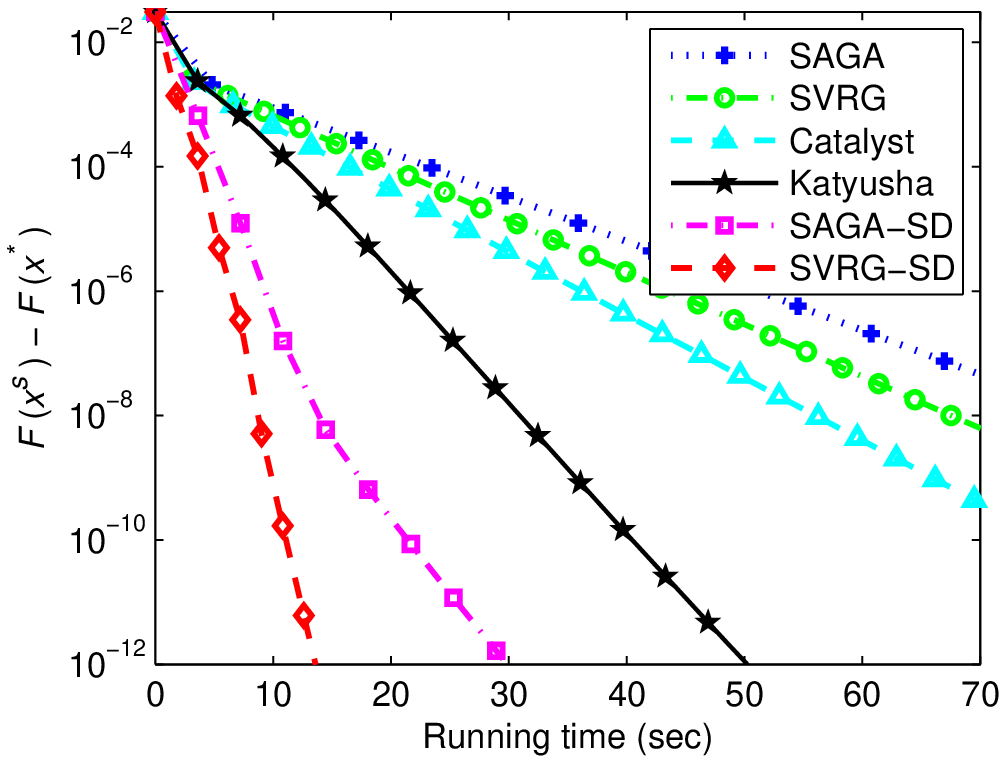}}
\vspace{-3.6mm}

\caption{Comparison of all the stochastic variance reduced gradient methods for solving strongly convex ridge regression problems with relatively small regularization parameters. The vertical axis is the objective value minus the minimum, and the horizontal axis denotes the number of effective passes over the data (top) or the running time (bottom).}
\label{fig_sim3}
\end{figure}

\begin{figure}[th]
\centering
\includegraphics[width=0.326\columnwidth]{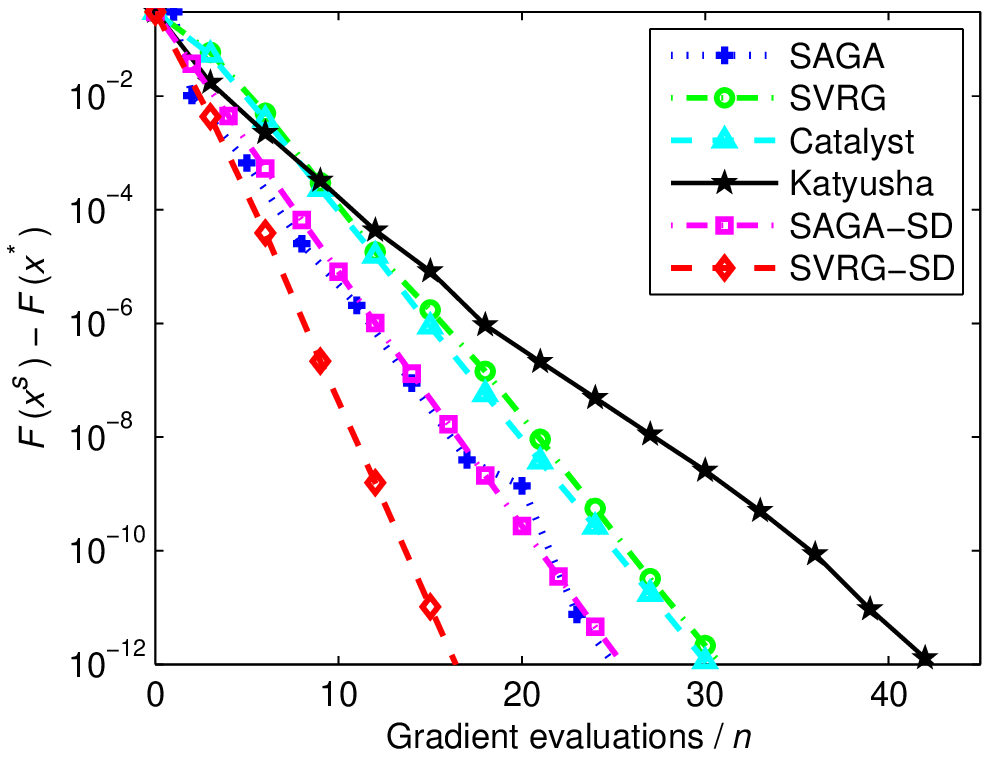}\,
\includegraphics[width=0.326\columnwidth]{Fig703}\,
\includegraphics[width=0.326\columnwidth]{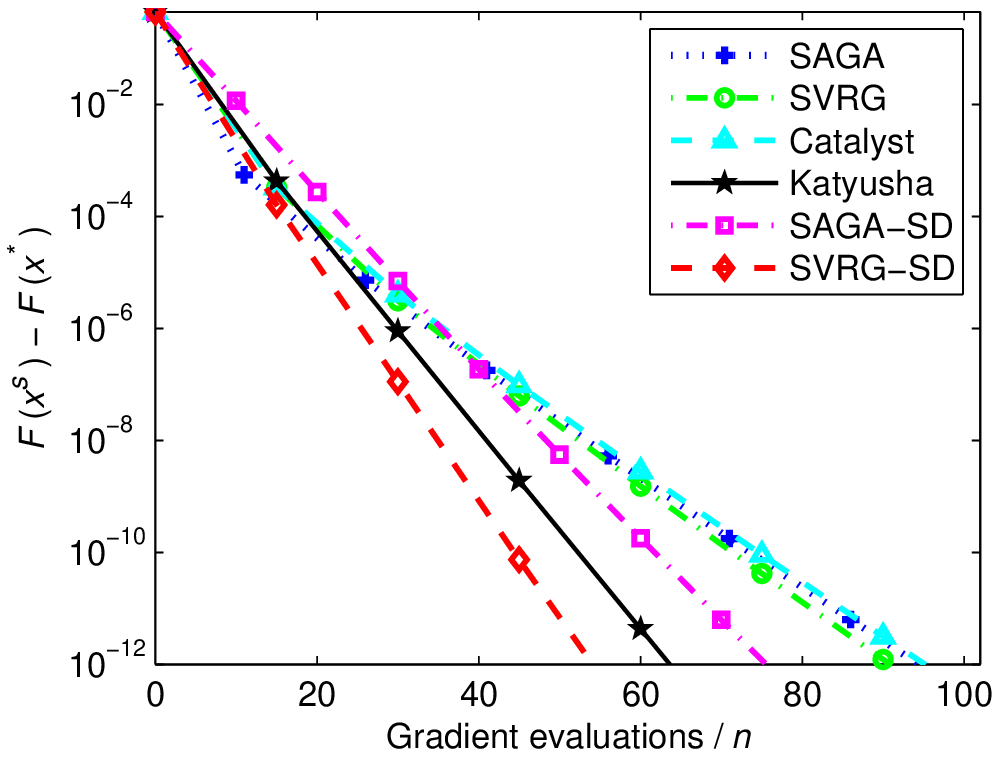}

\subfigure[Rcv1, $\lambda_{1}\!=\!10^{-3}$]{\includegraphics[width=0.326\columnwidth]{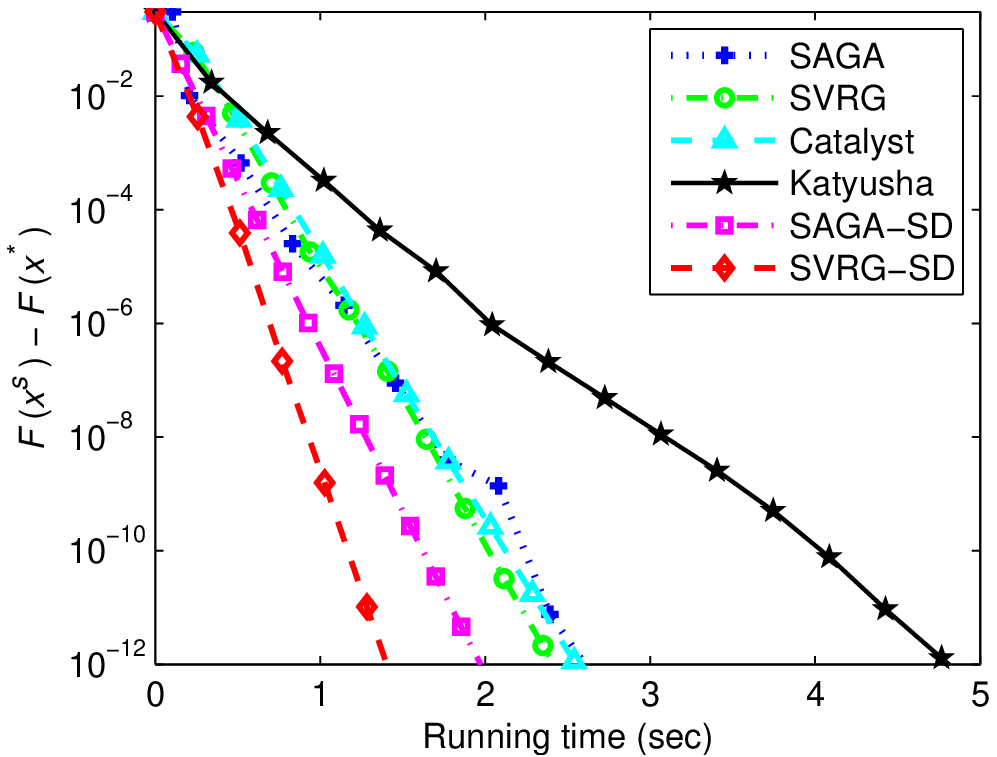}\label{fig2a}}\,
\subfigure[Rcv1, $\lambda_{1}\!=\!10^{-4}$]{\includegraphics[width=0.326\columnwidth]{Fig704}}\,
\subfigure[Rcv1, $\lambda_{1}\!=\!10^{-5}$]{\includegraphics[width=0.326\columnwidth]{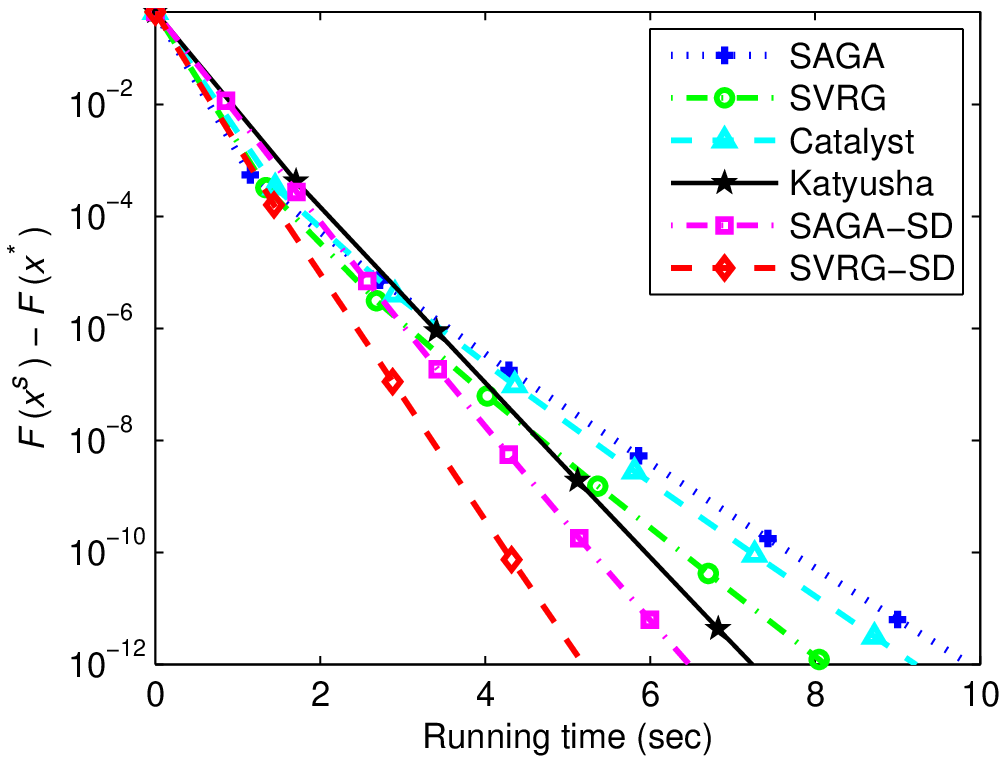}}
\vspace{-3.6mm}

\caption{Comparison of all the stochastic variance reduced gradient methods for solving strongly convex ridge regression problems with different regularization parameters on the sparse data set, Rcv1. The vertical axis represents the objective value minus the minimum, and the horizontal axis denotes the number of effective passes (top) or the running time (bottom).}
\label{fig_sim2}
\end{figure}

\vspace{3mm}
\section*{Appendix E: More Experimental Results}

\subsection*{Robustness}
Figure~\ref{fig11} shows the performance of SAGA~\cite{defazio:saga} and SAGA-SD with different values of $m_{1}$ for solving ridge regression problems on the Covtype data set, where the regularization parameter is $\lambda_{1}\!=\!10^{-4}$. From the result, we can observe that SAGA-SD significantly outperforms SAGA in terms of number of passes and running time. In particular, SAGA-SD, as well as SVRG-SD, has good robustness with respect to the number of iterations with sufficient decrease, which inspires us to use the partial sufficient decrease trick for both SVRG-SD and SAGA-SD.

\vspace{3mm}
\subsection*{Comparison of Results for Ridge Regression}
In this part, we first report the experimental results of SVRG~\cite{johnson:svrg}, SAGA~\cite{defazio:saga}, Catalyst~\cite{lin:vrsg}, Katyusha~\cite{zhu:Katyusha}, SVRG-SD and SAGA-SD for solving strongly convex (SC) ridge regression problems with the regularization parameter $\lambda_{1}\!=\!10^{-5}$ in Figure~\ref{fig_sim1}, where the horizontal axis denotes the number of effective passes over the data set (evaluating $n$ component gradients, or computing a single full gradient is considered as one effective pass) or the running time (seconds). Moreover, we report the performance of all the stochastic variance reduction methods for solving ridge regression problems with relative small regularization parameters (e.g., $\lambda_{1}\!=\!10^{-7}$) in Figure~\ref{fig_sim3}, which shows that SVRG-SD and SAGA-SD, as well as Katyusha, converge significantly faster than SAGA, SVRG, and Catalyst. In particular, SVRG-SD and SAGA-SD usually outperform Katyusha in terms of both number of passes and running time, which further justifies the effectiveness of our sufficient decrease technique for stochastic optimization.

Figure~\ref{fig_sim2} shows the performance of all the methods for solving ridge regression problems with different regularization parameters on the sparse data set, Rcv1. From the results, we can observe that SVRG-SD and SAGA-SD significantly outperform their counterparts: SVRG and SAGA in terms of both number of effective passes and running time. The accelerated method, Catalyst, usually outperforms the non-accelerated methods, SVRG and SAGA. Katyusha converges much faster than SAGA, SVRG, and Catalyst for the cases when the regularization parameter is relatively small (e.g., $\lambda_{1}\!=\!10^{-5}$), whereas it sometime achieves similar or inferior performance when the regularization parameter is relatively large (e.g., $\lambda_{1}\!=\!10^{-3}$), as shown in Figures~\ref{fig2a}. Moreover, SVRG-SD and SAGA-SD achieve at least comparable performance with the accelerated stochastic method, Katyusha~\cite{zhu:Katyusha}, in terms of number of effective passes. Since SVRG-SD and SAGA-SD have much lower per-iteration complexities than Katyusha, they have more obvious advantage over Katyusha in terms of running time.

\begin{figure}[th]
\centering
\includegraphics[width=0.326\columnwidth]{Fig51}\,
\includegraphics[width=0.326\columnwidth]{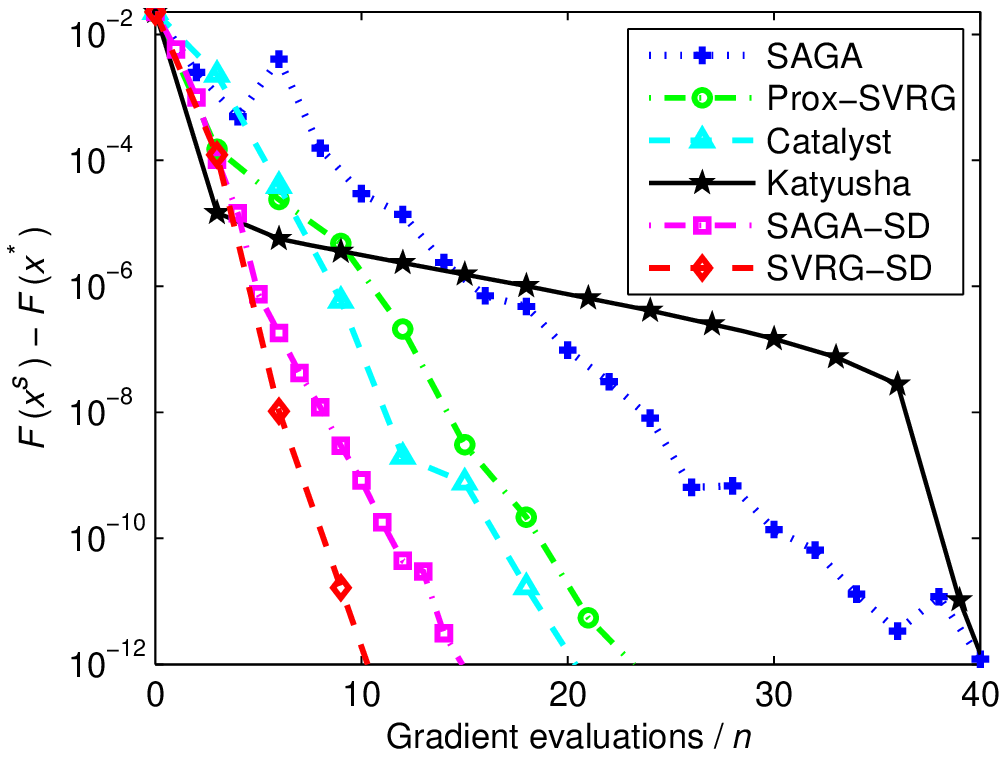}\,
\includegraphics[width=0.326\columnwidth]{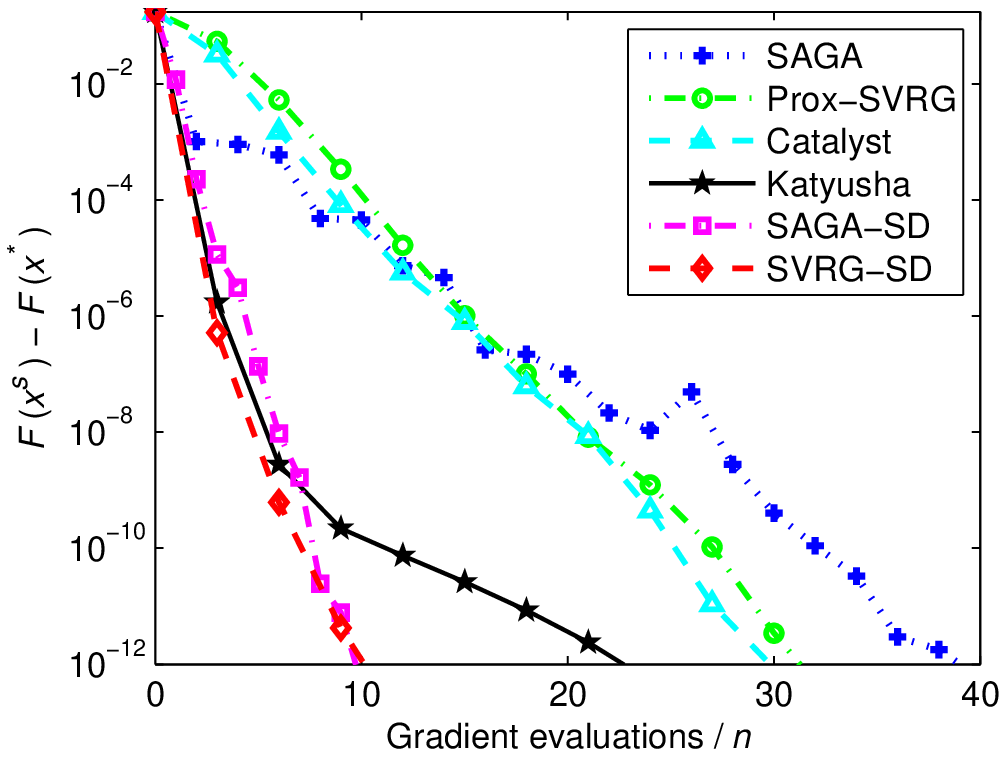}

\subfigure[Ijcnn1, $\lambda_{2}\!=\!10^{-4}$]{\includegraphics[width=0.326\columnwidth]{Fig52}}\,
\subfigure[Covtype, $\lambda_{2}\!=\!10^{-4}$]{\includegraphics[width=0.326\columnwidth]{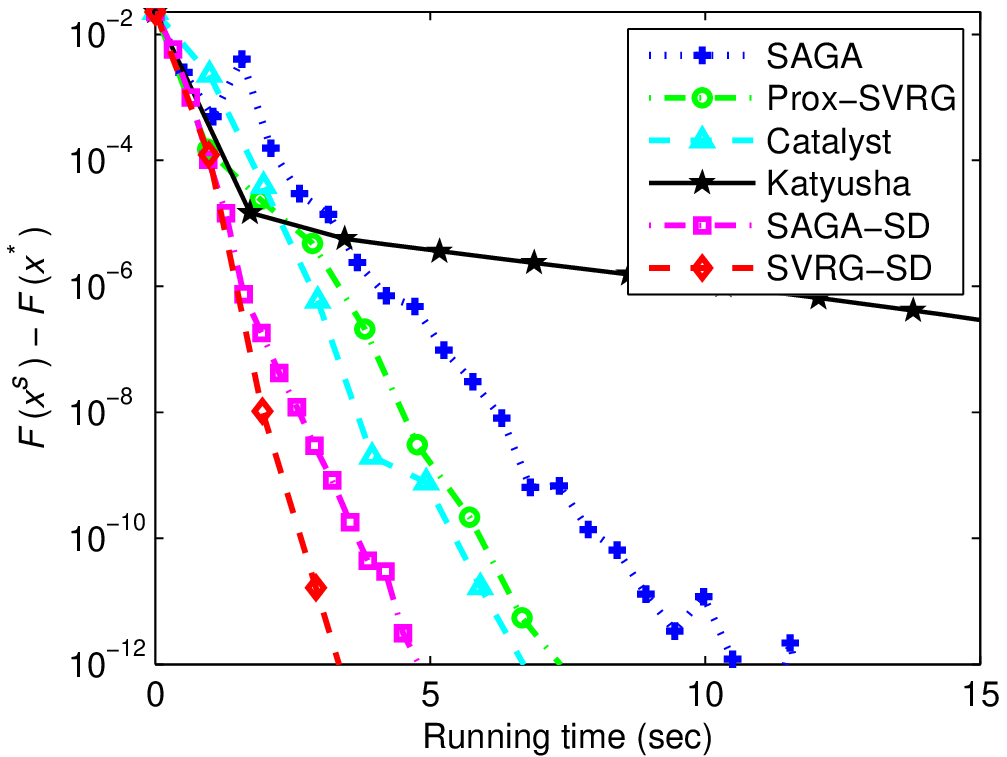}}\,
\subfigure[SUSY, $\lambda_{2}\!=\!10^{-4}$]{\includegraphics[width=0.326\columnwidth]{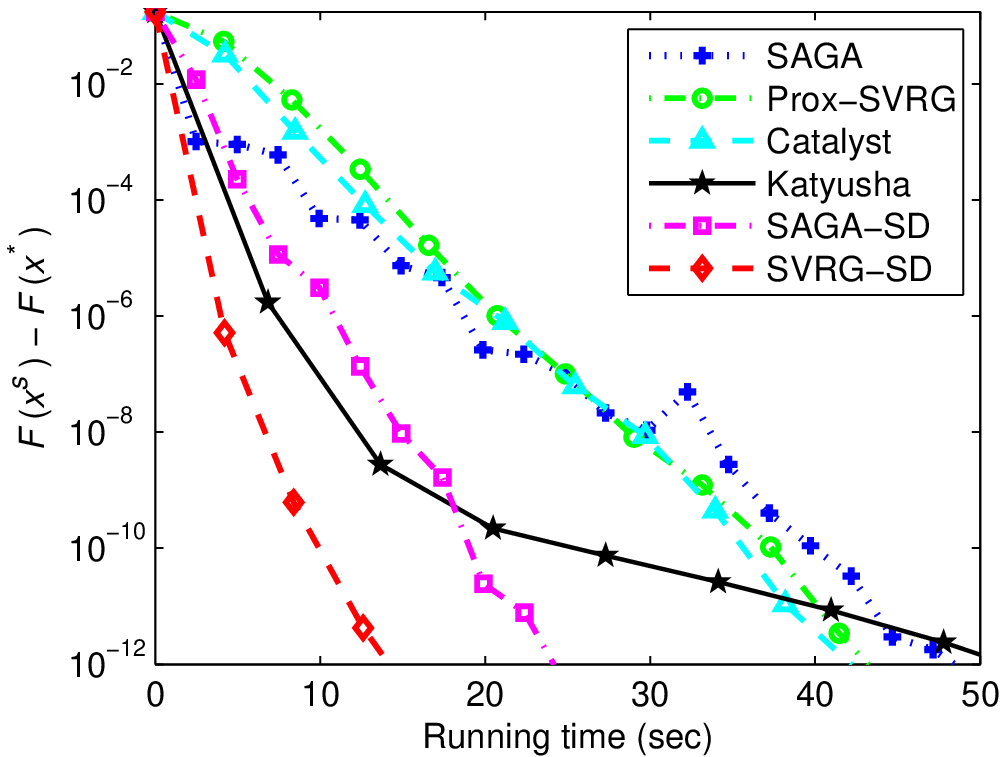}}
\vspace{1.9mm}

\includegraphics[width=0.326\columnwidth]{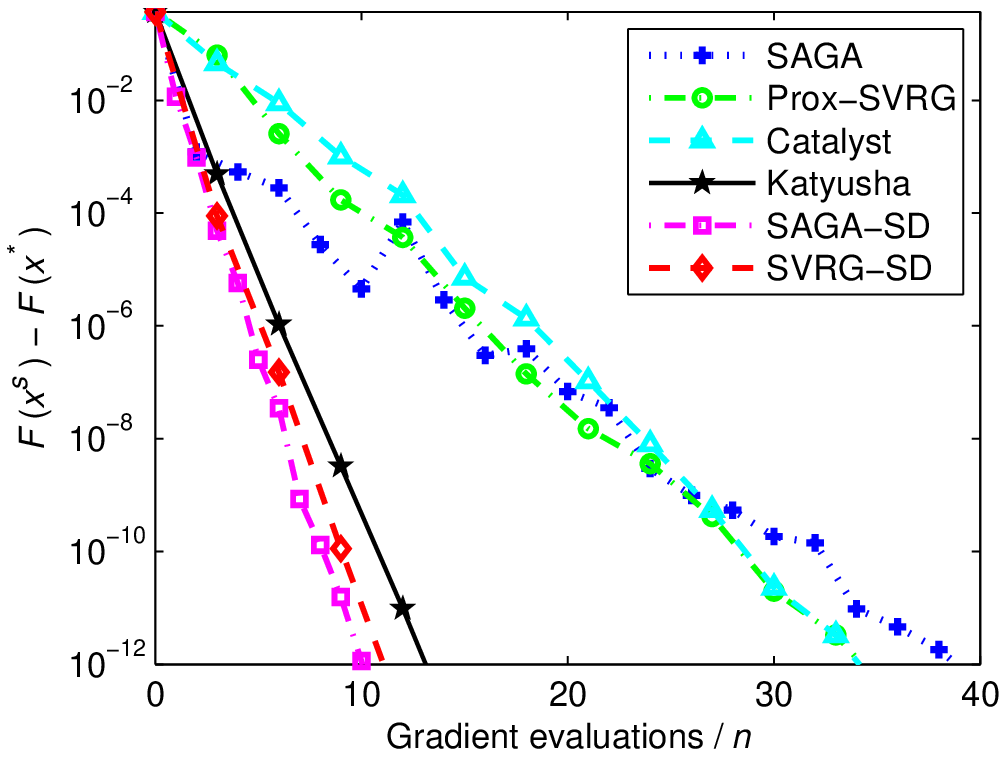}\,
\includegraphics[width=0.326\columnwidth]{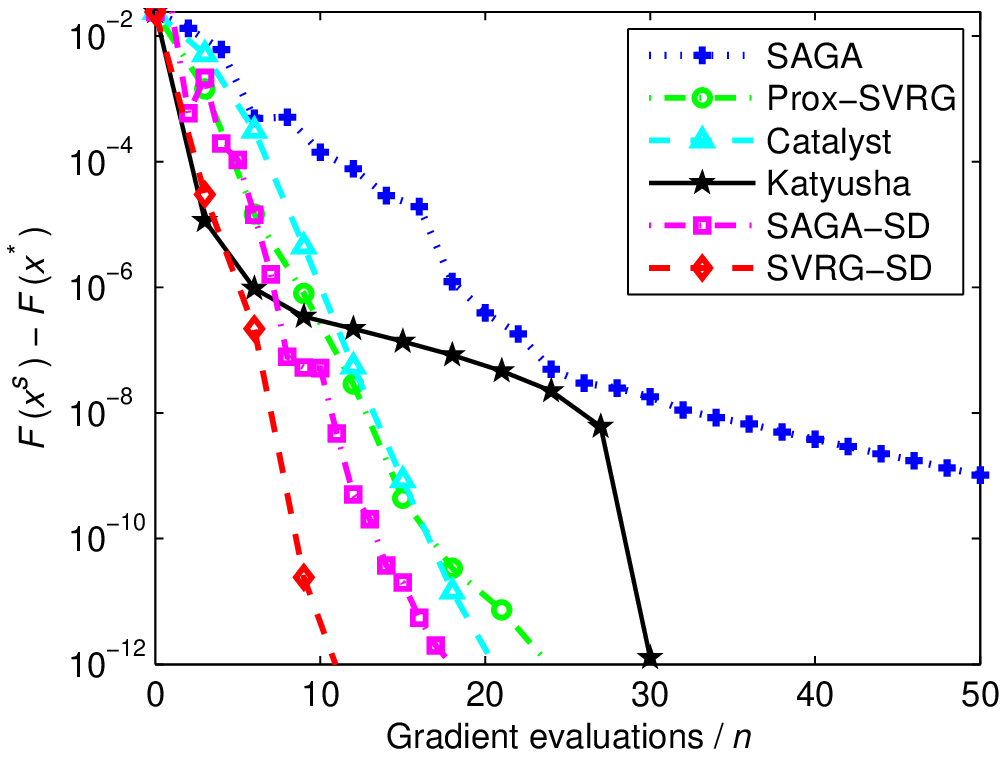}\,
\includegraphics[width=0.326\columnwidth]{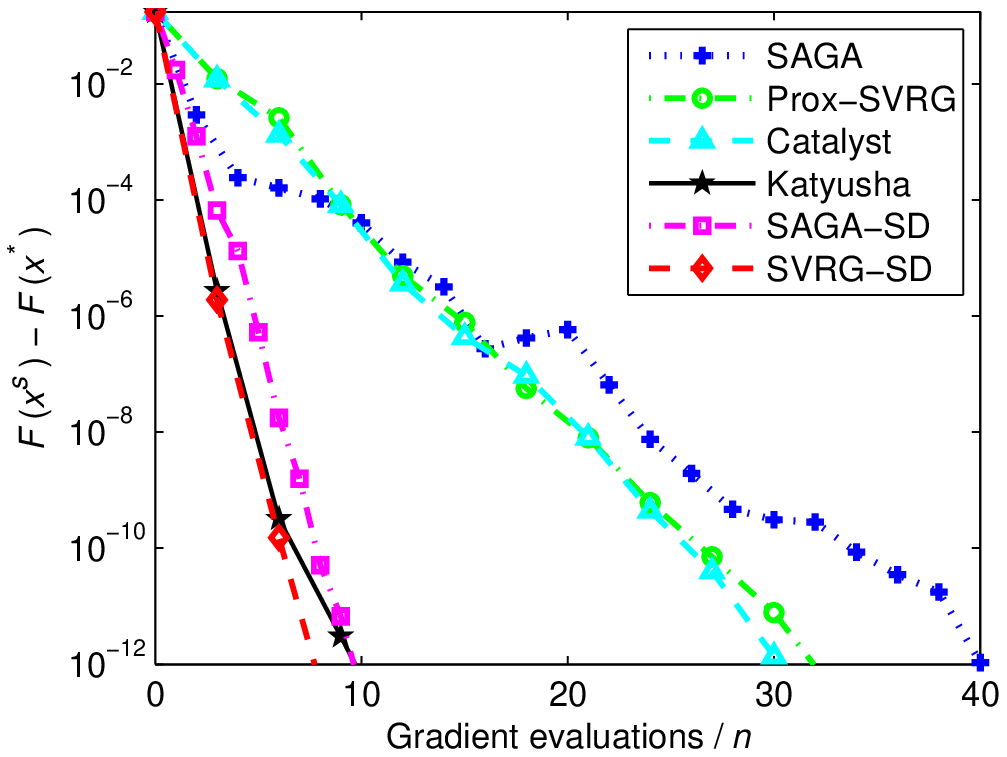}

\subfigure[Ijcnn1, $\lambda_{2}\!=\!10^{-5}$]{\includegraphics[width=0.326\columnwidth]{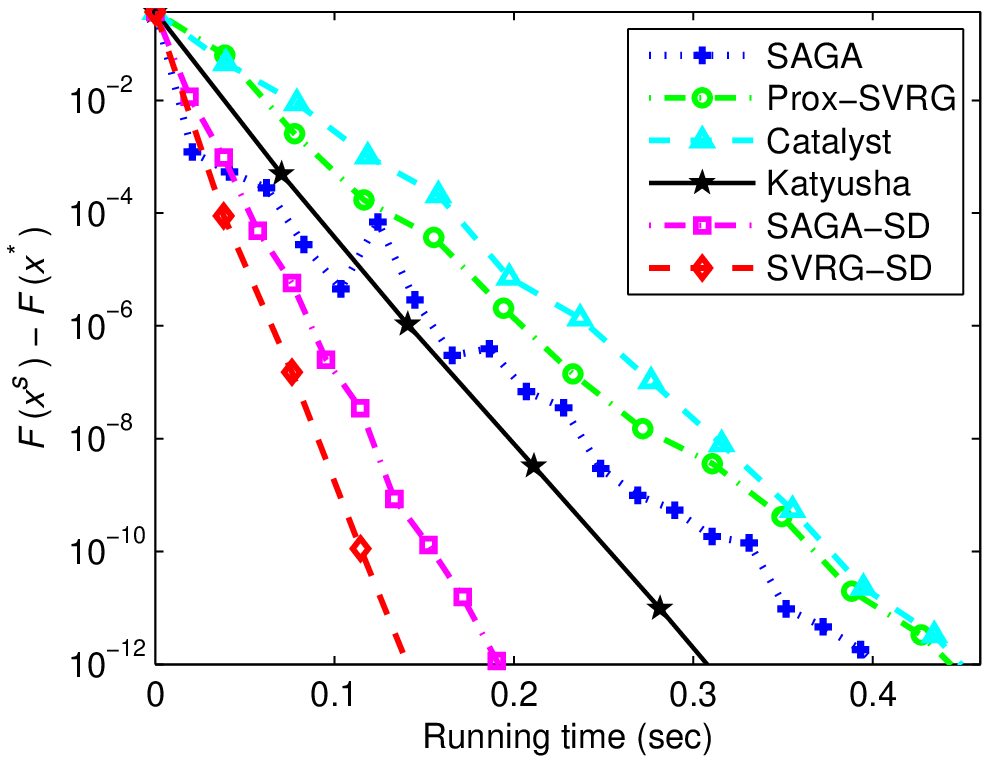}}\,
\subfigure[Covtype, $\lambda_{2}\!=\!10^{-5}$]{\includegraphics[width=0.326\columnwidth]{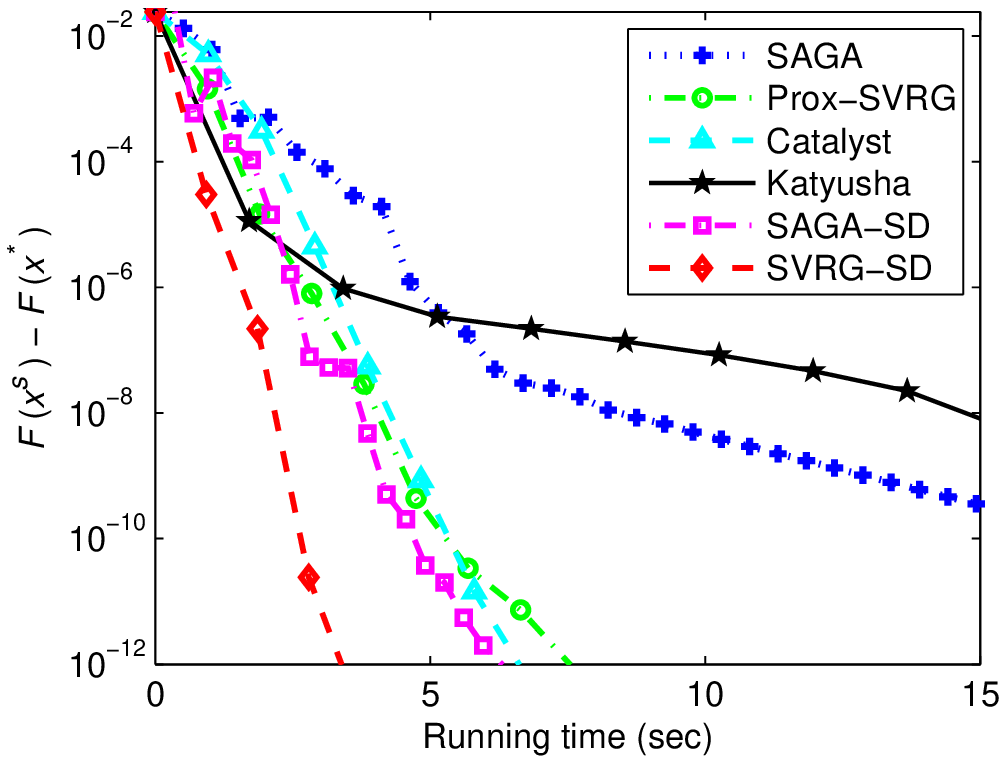}}\,
\subfigure[SUSY, $\lambda_{2}\!=\!10^{-5}$]{\includegraphics[width=0.326\columnwidth]{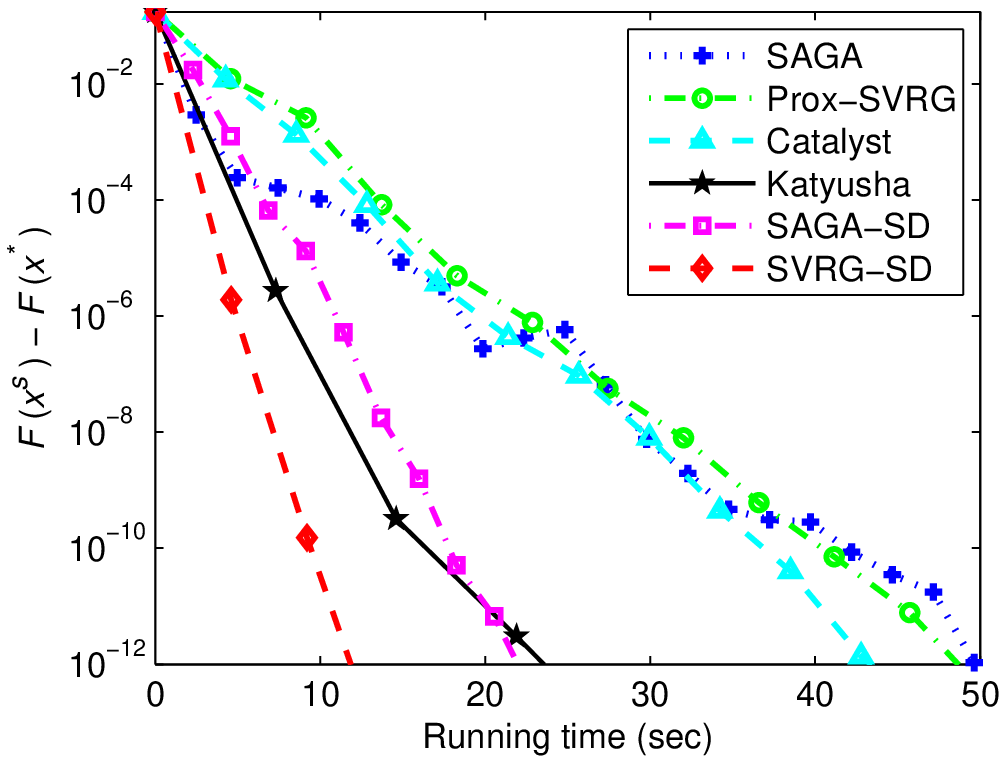}}
\vspace{-3.9mm}

\caption{Comparison of all the stochastic variance reduced gradient methods for solving non-strongly convex Lasso problems on the three data sets. The vertical axis is the objective value minus the minimum, and the horizontal axis denotes the number of effective passes over the data (top) or the running time (seconds, bottom).}
\label{fig_sim4}
\end{figure}

\begin{figure}[th]
\centering
\includegraphics[width=0.326\columnwidth]{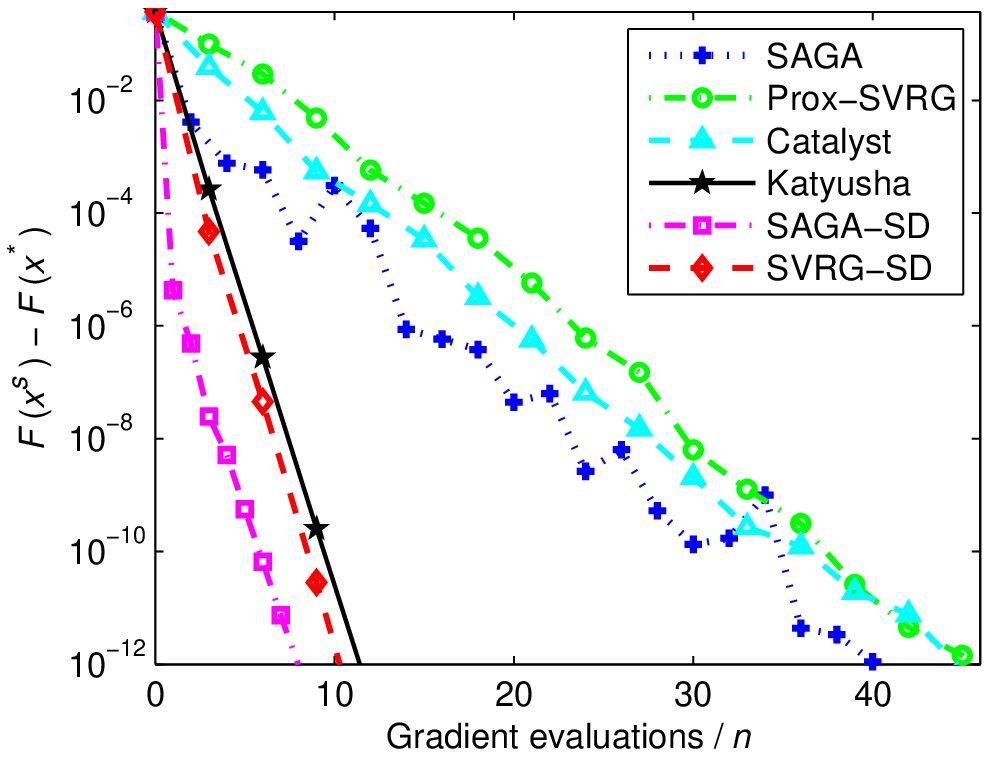}\,
\includegraphics[width=0.326\columnwidth]{Fig76}\,
\includegraphics[width=0.326\columnwidth]{Fig81}

\subfigure[$\lambda_{1}=10^{-5}$ \;and\; $\lambda_{2}=10^{-5}$]{\includegraphics[width=0.326\columnwidth]{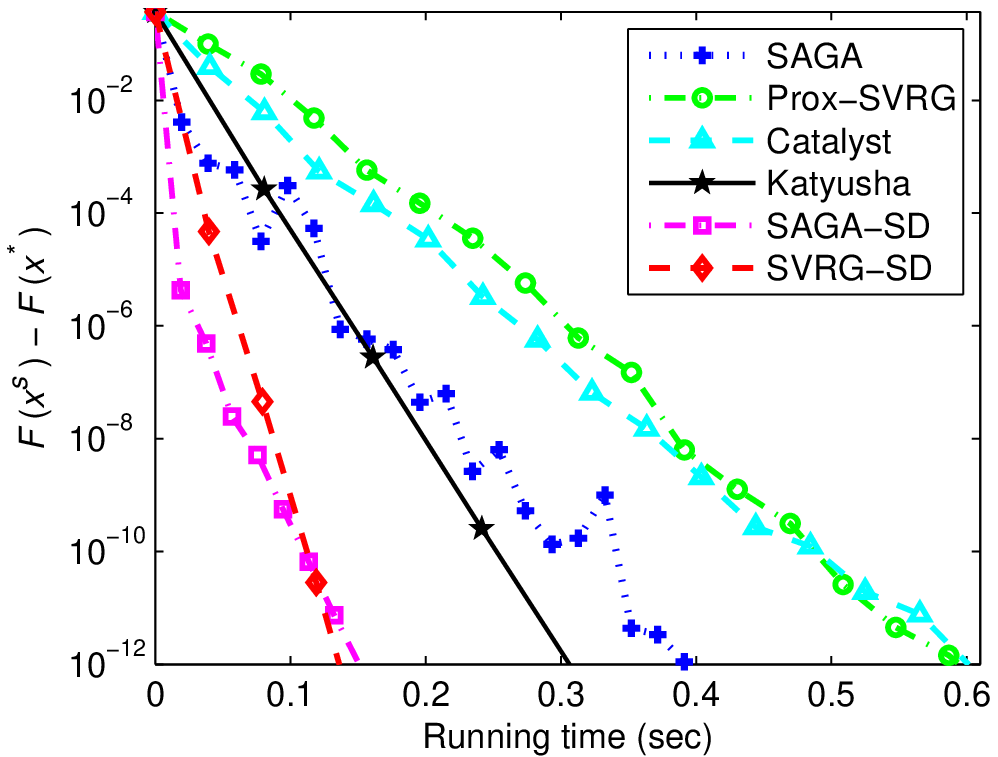}\,\includegraphics[width=0.326\columnwidth]{Fig77}\,\includegraphics[width=0.326\columnwidth]{Fig82}}
\vspace{1mm}

\includegraphics[width=0.326\columnwidth]{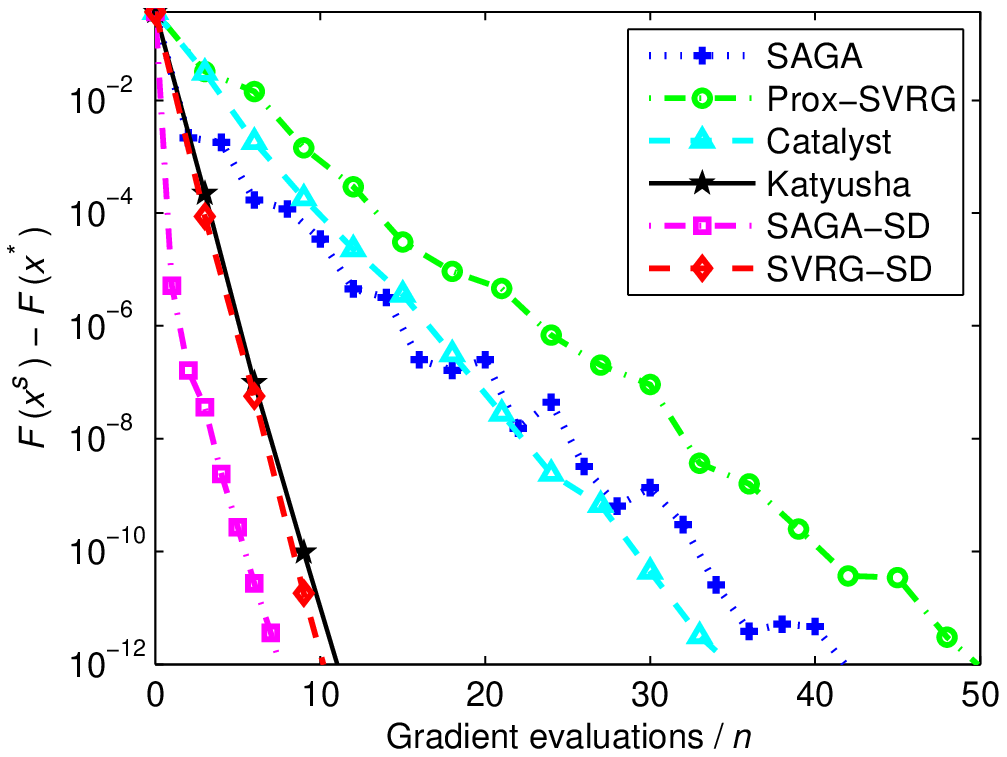}\,
\includegraphics[width=0.326\columnwidth]{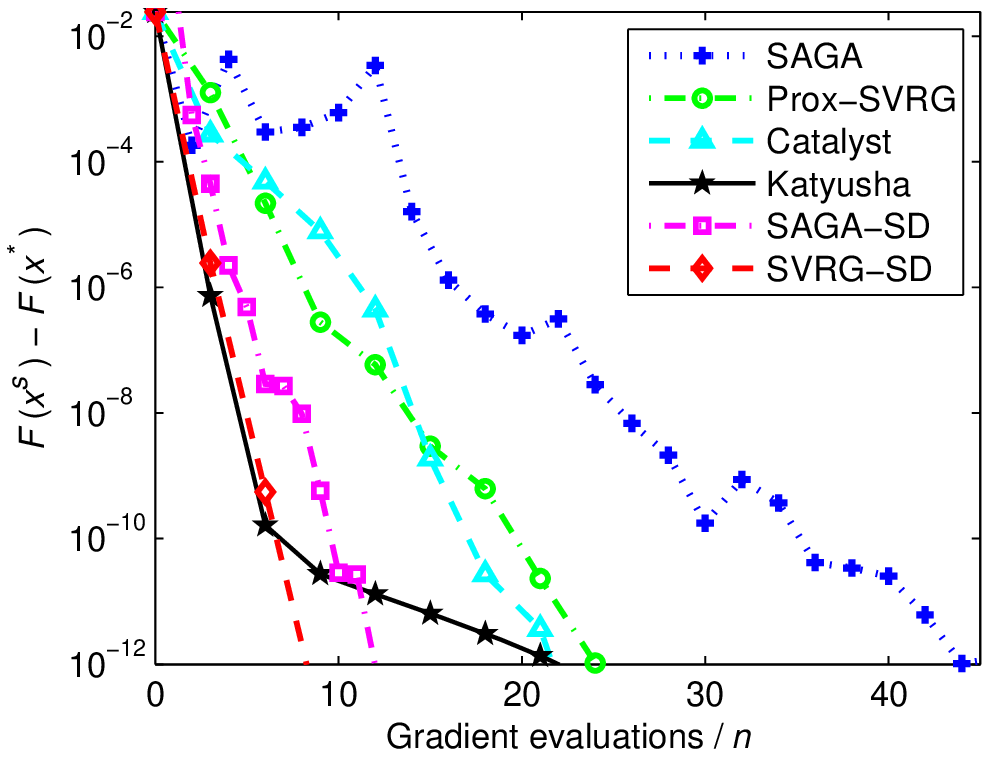}\,
\includegraphics[width=0.326\columnwidth]{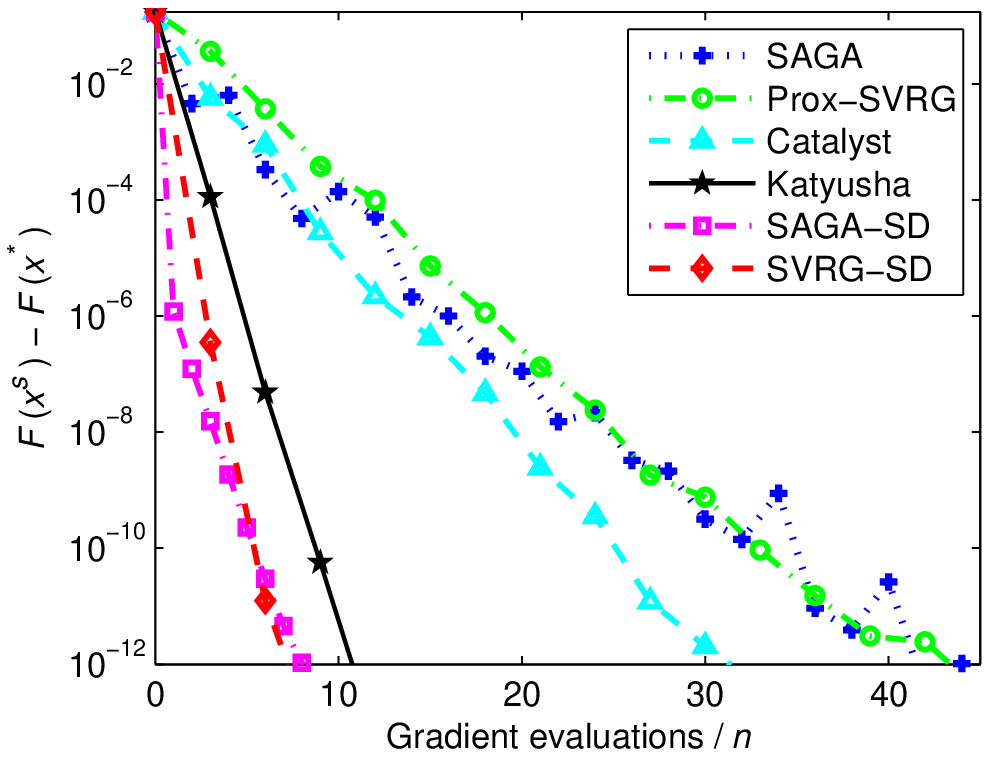}

\subfigure[$\lambda_{1}=10^{-5}$ \;and\; $\lambda_{2}=10^{-6}$]{\includegraphics[width=0.326\columnwidth]{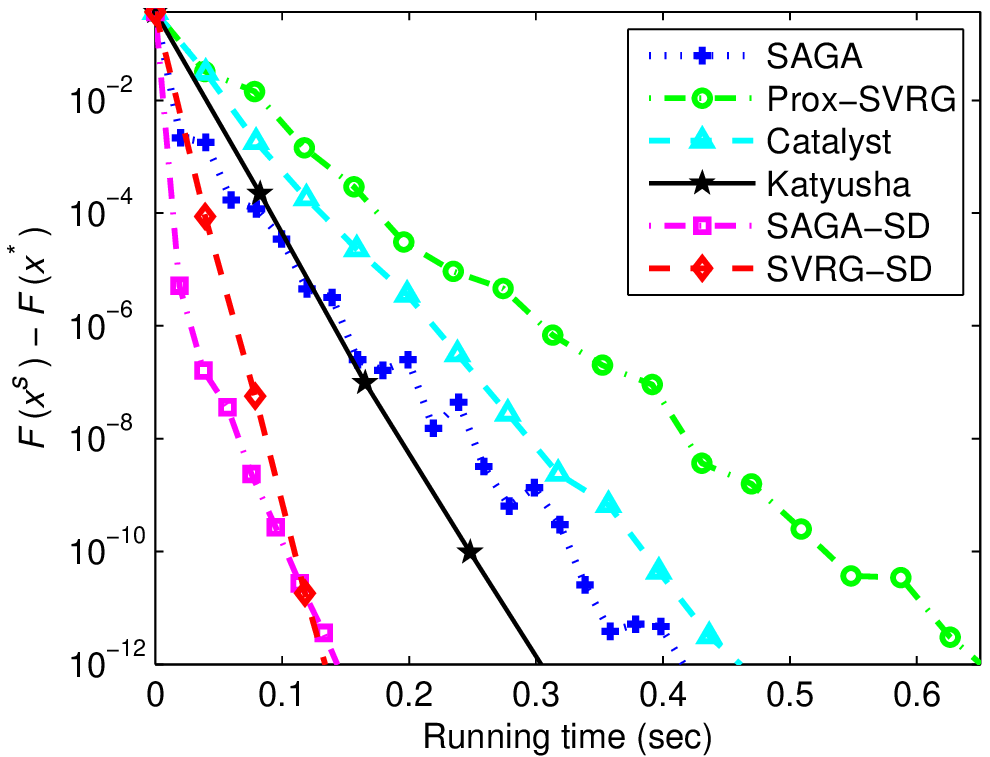}\,\includegraphics[width=0.326\columnwidth]{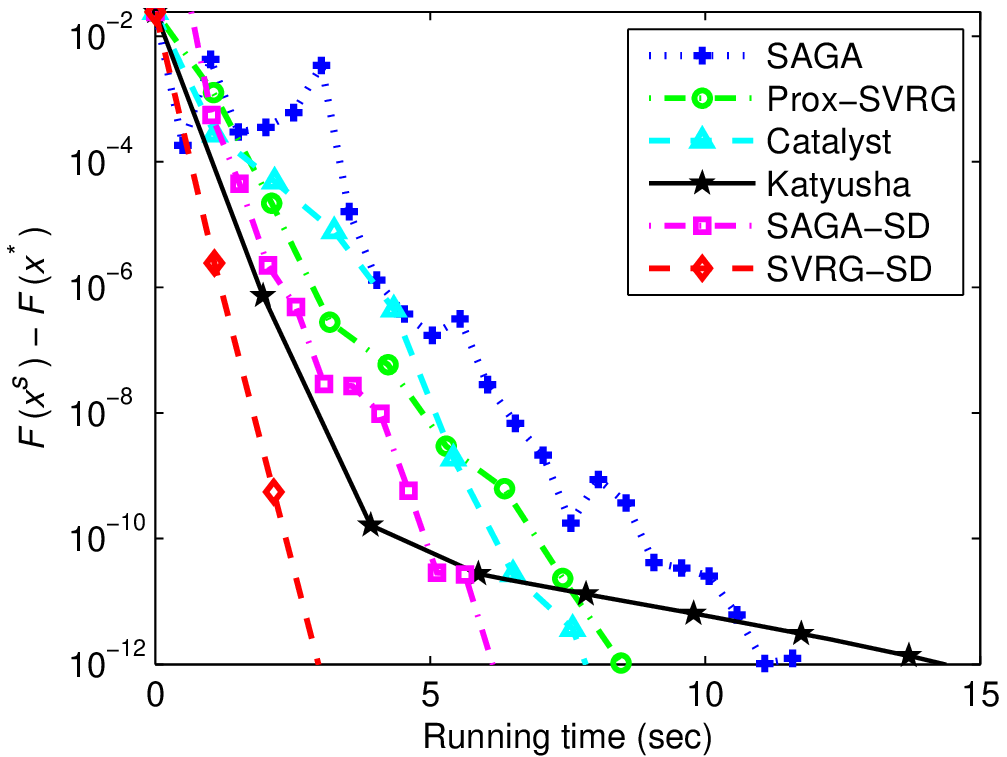}\,\includegraphics[width=0.326\columnwidth]{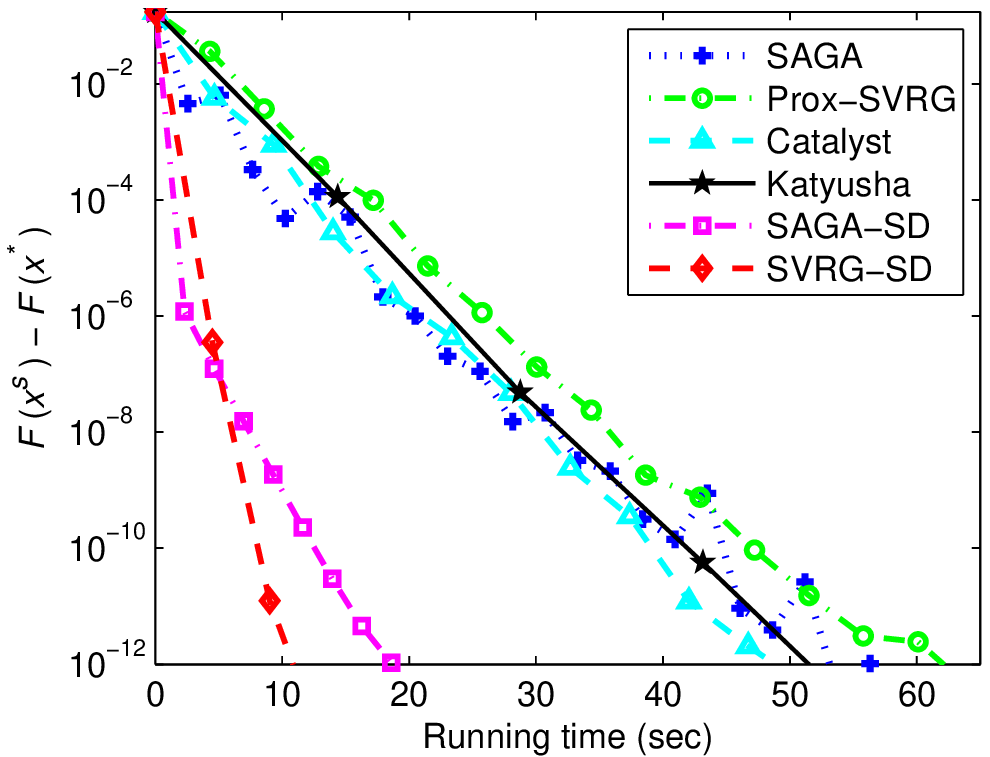}}
\caption{Comparison of all the stochastic methods for solving elastic-net (i.e., $(\lambda_{1}/2)\|\!\cdot\!\|^2\!+\!\lambda_{2}\|\!\cdot\!\|_{1}$) problems on Ijcnn1 (the first column), Covtype (the second column), and SUSY (the last column). The vertical axis is the objective value minus the minimum, and the horizontal axis denotes the number of effective passes over the data (top) or the running time (seconds, bottom).}
\label{fig_sim6}
\end{figure}

\vspace{3mm}

\subsection*{Comparison of Results for Lasso and Elastic-Net}
Finally, we report the performance of Prox-SVRG~\cite{xiao:prox-svrg}, SAGA~\cite{defazio:saga}, Catalyst~\cite{lin:vrsg}, Katyusha~\cite{zhu:Katyusha}, SVRG-SD and SAGA-SD for solving Lasso and elastic-net problems with different regularization parameters in Figures~\ref{fig_sim4} and \ref{fig_sim6}, respectively, from which we can observe that SVRG-SD and SAGA-SD also achieve much faster convergence speed than their counterparts: Prox-SVRG and SAGA, respectively. In particular, they also have comparable or better performance than the accelerated methods, Catalyst and Katyusha for both strongly convex and non-strongly convex problems. For the elastic-net problem, each component function $f_{i}(x)$ is defined as follows:
\begin{equation*}
f_{i}(x)=\frac{1}{2}(a_i^Tx - b_i)^2+\frac{\lambda_{1}}{2}\|x\|^{2}.
\end{equation*}

\bibliographystyle{unsrt}
\bibliography{nips2017}

\begin{thebibliography}{10}

\bibitem{krizhevsky:deep}
A.~Krizhevsky, I.~Sutskever, and G.~E. Hinton.
\newblock Image{N}et classification with deep convolutional neural networks.
\newblock In {\em NIPS}, pages 1097--1105, 2012.

\bibitem{zhang:sgd}
T.~Zhang.
\newblock Solving large scale linear prediction problems using stochastic
  gradient descent algorithms.
\newblock In {\em ICML}, pages 919--926, 2004.

\bibitem{johnson:svrg}
R.~Johnson and T.~Zhang.
\newblock Accelerating stochastic gradient descent using predictive variance
  reduction.
\newblock In {\em NIPS}, pages 315--323, 2013.

\bibitem{zhao:prox-smd}
P.~Zhao and T.~Zhang.
\newblock Stochastic optimization with importance sampling for regularized loss
  minimization.
\newblock In {\em ICML}, pages 1--9, 2015.

\bibitem{rakhlin:sgd}
A.~Rakhlin, O.~Shamir, and K.~Sridharan.
\newblock Making gradient descent optimal for strongly convex stochastic
  optimization.
\newblock In {\em ICML}, pages 449--456, 2012.

\bibitem{shamir:sgd2}
O.~Shamir and T.~Zhang.
\newblock Stochastic gradient descent for non-smooth optimization: Convergence
  results and optimal averaging schemes.
\newblock In {\em ICML}, pages 71--79, 2013.

\bibitem{roux:sag}
N.~Le Roux, M.~Schmidt, and F.~Bach.
\newblock A stochastic gradient method with an exponential convergence rate for
  finite training sets.
\newblock In {\em NIPS}, pages 2672--2680, 2012.

\bibitem{shalev-Shwartz:sdca}
S.~Shalev-Shwartz and T.~Zhang.
\newblock Stochastic dual coordinate ascent methods for regularized loss
  minimization.
\newblock {\em J. Mach. Learn. Res.}, 14:567--599, 2013.

\bibitem{defazio:saga}
A.~Defazio, F.~Bach, and S.~Lacoste-Julien.
\newblock {SAGA}: A fast incremental gradient method with support for
  non-strongly convex composite objectives.
\newblock In {\em NIPS}, pages 1646--1654, 2014.

\bibitem{defazio:Finito}
A.~Defazio, T.~Caetano, and J.~Domke.
\newblock Finito: A faster, permutable incremental gradient method for big data
  problems.
\newblock In {\em ICML}, pages 1125--1133, 2014.

\bibitem{mairal:miso}
J.~Mairal.
\newblock Incremental majorization-minimization optimization with application
  to large-scale machine learning.
\newblock {\em SIAM J. Optim.}, 25(2):829--855, 2015.

\bibitem{schmidt:sag}
M.~Schmidt, N.~Le Roux, and F.~Bach.
\newblock Minimizing finite sums with the stochastic average gradient.
\newblock {\em Math. Program.}, 162:83--112, 2017.

\bibitem{shalev-Shwartz:prox-sdca}
S.~Shalev-Shwartz and T.~Zhang.
\newblock Accelerated proximal stochastic dual coordinate ascent for
  regularized loss minimization.
\newblock {\em Math. Program.}, 155:105--145, 2016.

\bibitem{xiao:prox-svrg}
L.~Xiao and T.~Zhang.
\newblock A proximal stochastic gradient method with progressive variance
  reduction.
\newblock {\em SIAM J. Optim.}, 24(4):2057--2075, 2014.

\bibitem{hofmann:vrsg}
T.~Hofmann, A.~Lucchi, S.~Lacoste-Julien, and B.~McWilliams.
\newblock Variance reduced stochastic gradient descent with neighbors.
\newblock In {\em NIPS}, pages 2296--2304, 2015.

\bibitem{lin:vrsg}
H.~Lin, J.~Mairal, and Z.~Harchaoui.
\newblock A universal catalyst for first-order optimization.
\newblock In {\em NIPS}, pages 3366--3374, 2015.

\bibitem{nitanda:svrg}
A.~Nitanda.
\newblock Stochastic proximal gradient descent with acceleration techniques.
\newblock In {\em NIPS}, pages 1574--1582, 2014.

\bibitem{zhu:Katyusha}
Z.~Allen-Zhu.
\newblock Katyusha: The first direct acceleration of stochastic gradient
  methods.
\newblock In {\em STOC}, pages 1200--1205, 2017.

\bibitem{zhu:univr}
Z.~Allen-Zhu and Y.~Yuan.
\newblock Improved {SVRG} for non-strongly-convex or sum-of-non-convex
  objectives.
\newblock In {\em ICML}, pages 1080--1089, 2016.

\bibitem{babanezhad:vrsg}
R.~Babanezhad, M.~O. Ahmed, A.~Virani, M.~Schmidt, J.~Konecny, and S.~Sallinen.
\newblock Stop wasting my gradients: Practical {SVRG}.
\newblock In {\em NIPS}, pages 2242--2250, 2015.

\bibitem{zhang:svrg}
L.~Zhang, M.~Mahdavi, and R.~Jin.
\newblock Linear convergence with condition number independent access of full
  gradients.
\newblock In {\em NIPS}, pages 980--988, 2013.

\bibitem{shang:fsvrg}
F.~Shang, Y.~Liu, J.~Cheng, and J.~Zhuo.
\newblock Fast stochastic variance reduced gradient method with momentum
  acceleration for machine learning.
\newblock {\em arXiv:1703.07948v2}, 2017.

\bibitem{hazan:svrf}
E.~Hazan and H.~Luo.
\newblock Variance-reduced and projection-free stochastic optimization.
\newblock In {\em ICML}, pages 1263--1271, 2016.

\bibitem{woodworth:bound}
B.~Woodworth and N.~Srebro.
\newblock Tight complexity bounds for optimizing composite objectives.
\newblock In {\em NIPS}, pages 3639--3647, 2016.

\bibitem{li:apg}
H.~Li and Z.~Lin.
\newblock Accelerated proximal gradient methods for nonconvex programming.
\newblock In {\em NIPS}, pages 379--387, 2015.

\bibitem{wolfe:sdg}
P.~Wolfe.
\newblock Convergence conditions for ascent methods.
\newblock {\em SIAM Review}, 11(2):226--235, 1969.

\bibitem{defazio:sagab}
A.~Defazio.
\newblock A simple practical accelerated method for finite sums.
\newblock In {\em NIPS}, pages 676--684, 2016.

\bibitem{more:ls}
J.~More and D.~Thuente.
\newblock Line search algorithms with guaranteed sufficient decrease.
\newblock {\em ACM T. Math. Software}, 20:286--307, 1994.

\bibitem{mahsereci:sgd}
M.~Mahsereci and P.~Hennig.
\newblock Probabilistic line searches for stochastic optimization.
\newblock In {\em NIPS}, pages 181--189, 2015.

\bibitem{liu:sadmm}
Y.~Liu, F.~Shang, and J.~Cheng.
\newblock Accelerated variance reduced stochastic {ADMM}.
\newblock In {\em AAAI}, pages 2287--2293, 2017.

\bibitem{nesterov:fast}
Y.~Nesterov.
\newblock A method of solving a convex programming problem with convergence
  rate ${O}(1/k^2)$.
\newblock {\em Soviet Math. Doklady}, 27:372--376, 1983.

\bibitem{nesterov:co}
Y.~Nesterov.
\newblock {\em Introductory Lectures on Convex Optimization: A Basic Course}.
\newblock Kluwer Academic Publ., Boston, 2004.

\bibitem{beck:fista}
A.~Beck and M.~Teboulle.
\newblock A fast iterative shrinkage-thresholding algorithm for linear inverse
  problems.
\newblock {\em SIAM J. Imaging Sci.}, 2(1):183--202, 2009.

\bibitem{su:nag}
W.~Su, S.~Boyd, and E.~J. Candes.
\newblock A differential equation for modeling {N}esterov's accelerated
  gradient method: Theory and insights.
\newblock {\em J. Mach. Learn. Res.}, 17:5312--5354, 2016.

\bibitem{donoho:st}
D.~L. Donoho.
\newblock De-noising by soft-thresholding.
\newblock {\em IEEE Trans. Inform. Theory}, 41(3):613--627, 1995.

\bibitem{Armijo:line}
L.~Armijo.
\newblock Minimization of functions having {L}ipschitz continuous first partial
  derivatives.
\newblock {\em Pacific J. Math.}, 16(1):1--3, 1966.

\bibitem{bach:sgd}
F.~Bach and E.~Moulines.
\newblock Non-strongly-convex smooth stochastic approximation with convergence
  rate ${O}(1/n)$.
\newblock In {\em NIPS}, pages 773--781, 2013.

\bibitem{shang:svrgsd}
F.~Shang, Y.~Liu, J.~Cheng, K.~Ng, and Y.~Yoshida.
\newblock Guaranteed sufficient decrease for variance reduced stochastic
  gradient descent.
\newblock {\em arXiv: 1703.06807v2}, 2017.

\bibitem{frostig:sgd}
R.~Frostig, R.~Ge, S.~M. Kakade, and A.~Sidford.
\newblock Un-regularizing: approximate proximal point and faster stochastic
  algorithms for empirical risk minimization.
\newblock In {\em ICML}, pages 2540--2548, 2015.

\bibitem{zhu:box}
Z.~Allen-Zhu and E.~Hazan.
\newblock Optimal black-box reductions between optimization objectives.
\newblock In {\em NIPS}, pages 1606--1614, 2016.

\bibitem{nesterov:smooth}
Y.~Nesterov.
\newblock Smooth minimization of non-smooth functions.
\newblock {\em Math. Program.}, 103:127--152, 2005.

\bibitem{koneeny:mini}
Jakub Koneeny, Jie Liu, Peter Richtarik, , and Martin Takae.
\newblock Mini-batch semi-stochastic gradient descent in the proximal setting.
\newblock {\em IEEE J. Sel. Top. Sign. Proces.}, 10(2):242--255, 2016.

\bibitem{kingma:sgd}
D.~P. Kingma and J.~Ba.
\newblock Adam: A method for stochastic optimization.
\newblock In {\em ICLR}, 2015.

\bibitem{shang:svrg}
F.~Shang, K.~Zhou, J.~Cheng, I.~W. Tsang, L.~Zhang, and D.~Tao.
\newblock {VR-SGD}: A simple stochastic variance reduction method for machine
  learning.
\newblock {\em arXiv:1704.04966v2}, 2017.

\bibitem{shamir:sgd}
O.~Shamir.
\newblock Without-replacement sampling for stochastic gradient methods.
\newblock In {\em NIPS}, pages 46--54, 2016.

\bibitem{reddi:sgd}
S.~Reddi, A.~Hefny, S.~Sra, B.~Poczos, and A.~Smola.
\newblock On variance reduction in stochastic gradient descent and its
  asynchronous variants.
\newblock In {\em NIPS}, pages 2629--2637, 2015.

\bibitem{mania:svrg}
H.~Mania, X.~Pan, D.~Papailiopoulos, B.~Recht, K.~Ramchandran, and M.~I.
  Jordan.
\newblock Perturbed iterate analysis for asynchronous stochastic optimization.
\newblock {\em SIAM J. Optim.}, 27(4):2202--2229, 2017.

\bibitem{pedregosa:saga}
F.~Pedregosa, R.~Leblond, and S.~Lacoste-Julien.
\newblock Breaking the nonsmooth barrier: A scalable parallel method for
  composite optimization.
\newblock In {\em NIPS}, pages 55--64, 2017.

\bibitem{lee:dsgd}
J.~D. Lee, Q.~Lin, T.~Ma, and T.~Yang.
\newblock Distributed stochastic variance reduced gradient methods by sampling
  extra data with replacement.
\newblock {\em J. Mach. Learn. Res.}, 18:1--43, 2017.

\bibitem{zhu:univr2}
Z.~Allen-Zhu and Y.~Yuan.
\newblock Improved {SVRG} for non-strongly-convex or sum-of-non-convex
  objectives.
\newblock {\em arXiv: 1506.01972v3}, 2016.

\bibitem{baldassarre:prox}
L.~Baldassarre and M.~Pontil.
\newblock Advanced topics in machine learning part {II}: 5. {P}roximal methods.
\newblock {\em University Lecture}, 2013.

\bibitem{lan:sgd}
G.~Lan.
\newblock An optimal method for stochastic composite optimization.
\newblock {\em Math. Program.}, 133:365--397, 2012.

\end{thebibliography}

\end{document}